\documentclass{article}

    \PassOptionsToPackage{numbers}{natbib}

\usepackage[final]{neurips_2022}

\usepackage[svgnames,dvipsnames,color,table]{xcolor}
\usepackage[utf8]{inputenc} 
\usepackage[T1]{fontenc}    
\usepackage{hyperref}       
\hypersetup{
  colorlinks   = true, 
  urlcolor     = RoyalBlue, 
  linkcolor    = RoyalBlue, 
  citecolor   = RoyalBlue 
}
\usepackage{url}            
\usepackage{booktabs}       
\usepackage{amsfonts}       
\usepackage{nicefrac}       
\usepackage{microtype}      

\usepackage{url}
\usepackage{nicefrac}
\usepackage{bm}
\usepackage{booktabs}
\usepackage{amsfonts}
\usepackage{bbm}
\usepackage{mleftright}
\usepackage{dsfont}
\usepackage{amscd,amssymb,amsmath,amsthm,bbold}
\usepackage{graphicx}
\usepackage{float}
\usepackage{multirow}

\usepackage{pifont}
\usepackage{microtype}
\usepackage{enumitem}
\usepackage{xfrac}
\usepackage{parskip}
\usepackage{numprint}

\usepackage{transparent}

\usepackage{algorithm}
\usepackage{listings}
\usepackage{tikz}
\usepackage{xspace}
\usepackage{subcaption}
\usepackage{etoolbox}
\usepackage{comment}
\usepackage{etoc}
\usepackage{tcolorbox}
\usepackage{placeins}
\usepackage{makecell}

\usepackage{adjustbox}
\usepackage{array}

\usepackage{caption}
\usepackage{subcaption}
\usepackage{sidecap}
\usepackage{placeins}

\usepackage{xpatch}
\makeatletter
\renewcommand\paragraph{\@startsection{paragraph}{4}{\z@}                                     {0.85ex \@plus1ex \@minus.2ex}                                {-.5em}
{\normalfont\normalsize\bfseries}}
\makeatother

\usepackage{stmaryrd}
\usepackage{changepage}

\DeclareMathOperator*{\argmax}{arg\,max}

\usepackage[textsize=scriptsize,textwidth=1.9cm]{todonotes}
\reversemarginpar

\newcolumntype{R}[2]{%
    >{\adjustbox{angle=#1,lap=\width-(#2)}\bgroup}%
    l%
    <{\egroup}%
}
\newcommand*\rot{\multicolumn{1}{R{45}{1em}}}

\newif\ifcomments
\commentstrue

\ifcomments
    \newcommand\gamaga[1]{{\todo[color=teal]{GI: {#1}}}}
    \newcommand\ali[1]{{\todo[color=olive]{AF: {#1}}}}
    \newcommand\hanna[1]{{\todo[color=yellow]{HH: {#1}}}}
    \newcommand\mw[1]{{\todo[color=purple]{MW: {#1}}}}
    \newcommand\ludwig[1]{{\todo[color=orange]{LS: {#1}}}}
    \newcommand\hong[1]{{\todo[color=green]{HN: {#1}}}}
    \newcommand\mike[1]{{\todo[color=blue]{ML: {#1}}}}
\else
    \providecommand{\mw}[1]{}
    \providecommand{\gamaga}[1]{}
    \providecommand{\ali}[1]{}
    \providecommand{\hanna}[1]{}
    \providecommand{\ludwig}[1]{}
    \providecommand{\hong}[1]{}
    \providecommand{\mike}[1]{}
\fi

\newcommand\ourmethod{{\fontfamily{qtm}\selectfont PAINT}\xspace}

\def\newtask{patching task\xspace}
\def\oldtask{supported task\xspace}
\def\newtasks{patching tasks\xspace}
\def\oldtasks{supported tasks\xspace}
\def\bothtasks{patching and supported tasks\xspace}

\title{Patching open-vocabulary models\linebreak by interpolating weights}

\author{%
  Gabriel Ilharco\thanks{Equal contribution. Code available at \url{https://github.com/mlfoundations/patching}.} \\
  University of Washington \\
  {\tt\small gamaga@cs.washington.edu} \\
  \And
  Mitchell Wortsman$^*$ \\
  University of Washington \\
  {\tt\small mitchnw@cs.washington.edu} \\
  \And
  Samir Yitzhak Gadre$^*$ \\
  Columbia University \\
  {\tt\small sy@cs.columbia.edu} \\
  \And
  Shuran Song \\
  Columbia Universiry \\
  {\tt\small shurans@cs.columbia.edu} \\
  \And
  Hannaneh Hajishirzi \\
  University of Washington \\
  {\tt\small hannaneh@cs.washington.edu} \\
  \And
  Simon Kornblith \\
  Google Research, Brain Team \\
  {\tt\small skornblith@google.com}
  \And
  Ali Farhadi \\
  University of Washington \\
  {\tt\small ali@cs.washington.edu} \\
  \And
  Ludwig Schmidt \\
  University of Washington \\
  {\tt\small schmidt@cs.washington.edu}
}

\begin{document}

\maketitle

\vspace{-8pt}
\begin{abstract}
Open-vocabulary models like CLIP achieve high accuracy across many image classification tasks.
However, there are still settings where their zero-shot performance is far from optimal.
We study \emph{model patching}, where the goal is to improve accuracy on specific tasks without degrading accuracy on tasks where performance is already adequate.
Towards this goal, we introduce \ourmethod, a patching method that uses interpolations between the weights of a model before fine-tuning and the weights after fine-tuning on a task to be patched.
On nine tasks where zero-shot CLIP performs poorly, \ourmethod increases accuracy by 15 to 60 percentage points while preserving accuracy on ImageNet within one percentage point of the zero-shot model.
\ourmethod also allows a single model to be patched on multiple tasks and improves with model scale.
Furthermore, we identify cases of \emph{broad transfer}, where patching on one task increases accuracy on other tasks even when the tasks have disjoint classes.
Finally, we investigate applications beyond common benchmarks such as counting or reducing the impact of typographic attacks on CLIP.
Our findings demonstrate that it is possible to expand the set of tasks on which open-vocabulary models achieve high accuracy without re-training them from scratch.

\end{abstract}


\section{Introduction}
\label{sec:intro}

Open-vocabulary models are characterized by their ability to perform any image classification task based on text descriptions of the classes \cite{pham2021combined}.
Thanks to advances in large-scale pre-training, recent examples of open-vocabulary models such as CLIP and BASIC 
have reached parity with or surpassed important task-specific baselines, even 
when the open-vocabulary models are not fine-tuned on task-specific data (i.e., in a zero-shot setting) \cite{radford2021learning, jia2021scaling, pham2021combined,zhai2022lit,alayrac2022flamingo,yu2022coca}.
For instance, the largest CLIP model from \citet{radford2021learning} used in a zero-shot setting matches the ImageNet accuracy of a ResNet-50  trained on 1.2 million ImageNet images \cite{deng2009imagenet,he2016deep}.

\begin{SCfigure}
    \centering
    \begin{minipage}{0.48\linewidth}
    \hspace{-0.52cm}
    \includegraphics[width=\textwidth,scale=1]{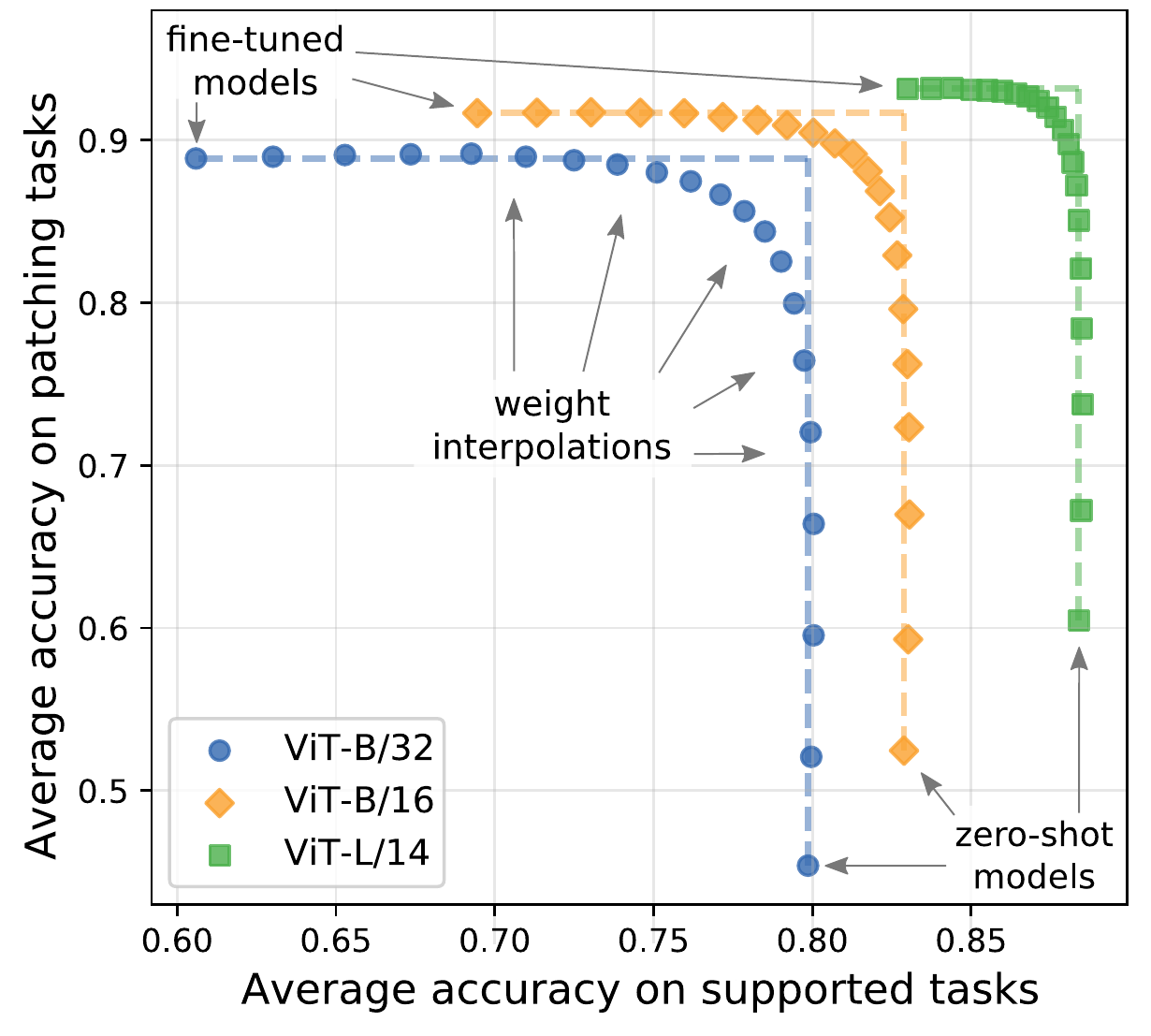}
  \end{minipage}
    \captionsetup{width=1.11\textwidth,font=footnotesize,labelfont=footnotesize}
    \hspace{-0.35cm}
    \caption{\textbf{Patching open-vocabulary models by linearly interpolating weights.}
    We wish to improve accuracy on tasks where a model performs poorly (\textit{patching tasks}), without degrading performance on tasks where accuracy is already adequate (\textit{\oldtasks}).
    When interpolating weights of fine-tuned models and zero-shot (unpatched) models,
    there are intermediate solutions where accuracy improves on the \newtask without reducing accuracy on \oldtasks.
    Results are shown for CLIP models \cite{radford2021learning}, averaged over nine \newtasks (Stanford Cars, DTD, EuroSAT, GTSRB, KITTI distance, MNIST, RESISC45, SUN397 and SVHN \cite{cars,dtd,eurosat,gtsrb,kitti,lecun1998mnist,cheng2017remote,stl10,sun397,svhn}) and five \oldtasks (ImageNet, CIFAR-10, CIFAR-100, STL-10 and Food101 \cite{deng2009imagenet,krizhevsky2009learning,stl10,food101}). We apply \ourmethod separately on each \newtask and average results across experiments.
    The dashed lines illustrate vertical movement from the unpatched models and horizontal movement from the fine-tuned models.
    }
    \label{fig:main_scatter}
\end{SCfigure}

Nevertheless, current open-vocabulary models still face challenges.
The same CLIP model that matches a ResNet-50 on ImageNet has lower MNIST accuracy than simple logistic regression in pixel space \citep{radford2021learning}.
Moreover, even when zero-shot models achieve good performance, they are usually still worse than models trained or fine-tuned on specific downstream tasks.

To address these issues, several authors have proposed methods for adapting zero-shot models to a task of interest using labeled data \cite{wortsman2021robust,zhou2022conditional,gao2021clip,zhang2021tip,kumar2022fine,sung2022vl}.
A common practice is to fine-tune the zero-shot model on the task of interest \cite{wortsman2021robust,pham2021combined}.
However, fine-tuned models can suffer from catastrophic forgetting \cite{mccloskey1989catastrophic,thrun1998lifelong,french1999catastrophic,kirkpatrick2017overcoming}, performing poorly on tasks where the zero-shot model initially performed well \cite{andreassen2021evolution,wortsman2021robust,pham2021combined}.
Additionally, fine-tuning typically produces a task-specific classification head, sacrificing the flexible text-based API that makes open-vocabulary models so appealing.
Whereas an open-vocabulary model can perform any classification task in a zero-shot fashion, a fine-tuned model with a task-specific head can only process the specific task that it was fine-tuned on.
This specialization can prevent knowledge obtained by fine-tuning on one task from transferring to other related tasks with different classes.

Another approach to adapting zero-shot models would be to add data from the downstream task to the pre-training dataset and train a new open-vocabulary model from scratch.
The resulting model could still perform any classification task, and zero-shot performance may improve on related tasks. However, training large image-text models from scratch can require hundreds of thousands of GPU hours \cite{radford2021learning,pham2021combined,yu2022coca}, which makes this approach practically infeasible in most settings.

In this paper, we study \textit{patching} open-vocabulary models, where the goal is to increase accuracy on new target tasks while maintaining the flexibility of the model and its accuracy on other tasks.\footnote{The term \textit{patching} is borrowed from software development terminology, drawing inspiration from recent work which conceptualizes developing machine learning models like open-source software \cite{raffel2021oss,ribeiro2022adaptive,matena2021merging,sung2021training}.}
Patching aims to combine the benefits of fine-tuning and re-training from scratch: improved performance on the task of interest, maintaining the flexibility of an open vocabulary, transfer between tasks, and fast adaptation time.
Motivated by these goals, we extend existing fine-tuning
techniques \cite{wortsman2021robust} to open-vocabulary settings, where the class space is not fixed.
We introduce Patching with Interpolation (\ourmethod), a simple, two-step procedure for patching models: first, fine-tune the model on the patching task without introducing any task-specific parameters; then, linearly interpolate between the weights of the model before and after fine-tuning.
Linearly interpolating neural network weights \cite{nagarajan19uniform, frankle2020linear,neyshabur2020being} has been previously used to improve accuracy on a single task \cite{izmailov2018averaging,wortsman2022model} or robustness to distribution shift \cite{wortsman2021robust}.
Indeed, averaging network weights has been explored in continual learning contexts, although for closed-vocabulary models \cite{lee2017overcoming}.

With \ourmethod, accuracy can improve on new tasks without degrading accuracy on unrelated tasks, as illustrated in Figure~\ref{fig:main_scatter}.
For instance, applying \ourmethod to a CLIP ViT-L/14 \cite{radford2021learning} independently on nine image classification tasks \cite{cars,dtd,eurosat,gtsrb,kitti,lecun1998mnist,cheng2017remote,sun397,svhn} improves accuracy by 15 to 60 percentage points compared to the unpatched model, while accuracy on ImageNet \cite{deng2009imagenet} decreases by less than one percentage point.
We also observe a promising trend: patching becomes more effective with model scale (Section \ref{sec:scale}).

Beyond single tasks, we show that models can be patched on multiple tasks (Section \ref{sec:multi_patches}).
When patching on nine image classification tasks simultaneously, a single CLIP ViT-L/14 model is competitive with using one specialized model for each task---the average accuracy difference is less than 0.5 percentage points.

Moreover, \ourmethod enables \emph{broad transfer} (Section \ref{sec:transfer}): accuracy on related tasks can increase, even when the class space changes. For instance, we partition EuroSAT \cite{eurosat}, a satellite image dataset, into two halves with disjoint labels.
Patching a ViT-L/14 model on the first half improves accuracy on the second half by 7.3 percentage points, even though the classes are unseen during patching.

Finally, we investigate \ourmethod on case studies including typographic attacks \cite{goh2021multimodal}, counting \cite{clevr}, and visual question answering \cite{vqa} (Section \ref{sec:case_studies}).
For instance, applying \ourmethod using synthetic typographic attacks leads to a model that is less susceptible to typographic attacks in the real world, improving its accuracy by 41 percentage points.

In summary:
\begin{itemize}[leftmargin=.28in]
\setlength\itemsep{-0.35em}
    \item Even the best pre-trained models are not perfect. We introduce \ourmethod, a method designed to improve accuracy on new tasks without harming accuracy elsewhere.
    \item \ourmethod incurs no extra computational cost compared to standard fine-tuning, neither during fine-tuning itself nor at inference time.
    \item \ourmethod can also be applied with multiple tasks, providing a single model that is competitive with many specialized models.
    \item Applying \ourmethod with one task can improve accuracy on a related task, even when they do not share the same classes.
    \item \ourmethod improves with model scale, indicating a promising trend for future models.
\end{itemize}

\section{Patching with interpolation (\ourmethod)}
\label{sec:method}

This section details our method for patching models on a single and multiple tasks.

\textbf{Patching on a single task.} Given an open-vocabulary model with weights $\theta_\textrm{zs}$ and a \newtask $\mathcal{D}_\text{patch}$,
our goal is to produce a new model $\theta_\text{patch}$ which achieves high accuracy on $\mathcal{D}_\text{patch}$ without decreasing model performance on tasks where accuracy is already acceptable.
We let $\mathcal{D}_\text{supp}$ denote a representative \oldtask where model performance is adequate, and later show that the method is stable under different choices of
$\mathcal{D}_\text{supp}$ (Section~\ref{sec:ablations}). The two-step procedure we explore for producing $\theta_{\text{patch}}$ is given below.

\begin{tcolorbox}

\textbf{Step 1.} Fine-tune $\theta_\textrm{zs}$ on training data from $\mathcal{D}_\text{patch}$ to produce a model with weights $\theta_\textrm{ft}$.

\textbf{Step 2.} For mixing coefficient $\alpha \in [0, 1]$, linearly interpolate between $\theta_\textrm{zs}$ and $\theta_\textrm{ft}$ to produce $\theta_{\text{patch}} = (1-\alpha) \cdot \theta_\textrm{zs} + \alpha \cdot \theta_\textrm{ft}$. The mixing coefficient is determined via held-out validation sets for $\mathcal{D}_\text{supp}$ and $\mathcal{D}_\text{patch}$. We refer to the resulting model as $\theta_{\text{patch}}$.
\end{tcolorbox}

In our experiments, we do not introduce any additional task-specific parameters when fine-tuning, as discussed in Section \ref{sec:setup} and Appendices~\ref{sec:clip_bg} and~\ref{sec:frozen_head}.

\textbf{Patching on a multiple tasks.} In practice, we often want to improve model accuracy on multiple \newtasks $\mathcal{D}_\text{patch}^{(1)}, ..., \mathcal{D}_\text{patch}^{(k)}$,
which can be accomplished with straightforward modifications to the procedure above.
We explore three alternatives and examine their relative trade-offs in Section~\ref{sec:multi_patches}:
\begin{itemize}[leftmargin=.28in]
\setlength\itemsep{-0.45em}
    \item \emph{Joint patching}, where we merge all the \newtasks $\mathcal{D}_\text{patch}^{(i)}$ into a single task $\mathcal{D}_\text{patch} $ before running the patching procedure;
    \item \emph{Sequential patching}, where we iteratively repeat the patching procedure above on each new task $\mathcal{D}_\text{patch}^{(i)}$ and let $\theta_\textrm{zs} \leftarrow \theta_{\text{patch}}$ after each completed iteration;
    \item \emph{Parallel patching}, where we apply the first step on each task in parallel to produce fine-tuned models with weights $\theta_\textrm{ft}^{(1)},...,\theta_\textrm{ft}^{(k)}$. Then, we search for mixing coefficients $\alpha_i$ to produce $\theta_\text{patch} = (1-\sum_{i=1}^k\alpha_i)\cdot\theta_\textrm{zs} + \sum_{i=1}^k\alpha_i\cdot \theta_\textrm{ft}^{(i)}$.
\end{itemize}

For joint and parallel patching we assume access to held-out validation sets for all tasks, while in sequential patching we only assume access to held-out validation sets from the tasks seen so far.
Unless mentioned otherwise, we pick the mixing coefficient $\alpha$ that optimizes average accuracy on the held-out validation sets from
the supported and patching tasks.


\section{Experimental setup}
\label{sec:setup}

\paragraph{Tasks.} 
We consider a diverse set of image classification tasks from \citet{radford2021learning}.
In most experiments, we use ImageNet \cite{deng2009imagenet} as a representative \oldtask, although we explore other \oldtasks in Section \ref{sec:ablations}.
We categorize tasks into \newtasks or \oldtasks based on the accuracy difference between the zero-shot model and a model specialized to the task.
A large accuracy difference indicates that the task is a relevant target for patching because the zero-shot model is still far from optimal.
Specifically, we consider a subset tasks from \citet{radford2021learning}, categorizing tasks where the linear probes outperform the zero-shot model by over 10 percentage points as \newtasks: Cars \citep{cars}, DTD \citep{dtd}, EuroSAT \citep{eurosat}, GTSRB \citep{gtsrb}, KITTI \citep{kitti}, MNIST \citep{lecun1998mnist}, RESISC45 \citep{cheng2017remote}, SUN397 \citep{sun397}, and SVHN \citep{svhn}.
We use the remaining tasks as \oldtasks: CIFAR10 \citep{krizhevsky2009learning}, CIFAR100 \citep{krizhevsky2009learning}, Food101 \citep{food101}, ImageNet \citep{deng2009imagenet}, and STL10 \citep{stl10}.
We investigate additional \newtasks as case studies in Section \ref{sec:case_studies} and provide further details in Appendix \ref{sec:appendix_datasets}.

\paragraph{Models.} We primarily use CLIP~\cite{radford2021learning} pre-trained vision transformer (ViT) models \cite{dosovitskiy2021an}. Unless otherwise mentioned our experiments are with the ViT-L/14 model, while Section~\ref{sec:ablations} studies ResNets~\cite{he2016deep}.

\paragraph{Fine-tuning on \newtasks.}

Unless otherwise mentioned, we fine-tune with a batch size of 128 for 2000 iterations using learning rate 1e-5 with 200 warm-up steps
with a cosine annealing learning rate schedule and the AdamW optimizer \cite{loshchilov2018decoupled, paszke2019pytorch} (weight decay 0.1).
When fine-tuning, we use the frozen final classification layer output by CLIP's text tower so that we do not introduce additional learnable parameters.
This design decision keeps the model open-vocabulary and does not harm accuracy, as discussed in 
in Appendices~\ref{sec:clip_bg} and~\ref{sec:frozen_head}.

\paragraph{Evaluation.} We use accuracy as the evaluation metric unless otherwise stated.
We refer to the average of the mean accuracy on the \newtasks and the mean accuracy on the \oldtasks as \emph{combined accuracy}.\footnote{In other words, $(\mathbb{E}_{\mathcal{D}_\textrm{supp}}[\mathsf{Acc}(\theta, \mathcal{D}_\textrm{supp})] + \mathbb{E}_{\mathcal{D}_\textrm{patch}} [\mathsf{Acc}(\theta, \mathcal{D}_\textrm{patch})])/2$, where $\mathsf{Acc}(\theta, \mathcal{D}_\textrm{supp})$ and $\mathsf{Acc}(\theta, \mathcal{D}_\textrm{patch})$ are accuracies on \oldtasks and \newtasks, respectively.}  

\section{Patching models on a single new task}
\label{sec:patching}

As shown in Figure \ref{fig:main_scatter}, when patching a model on a single task, we interpolate the weights of the zero-shot and fine-tuned model, producing a model that achieves high accuracy on both the \newtask and the \oldtask.
On the nine tasks, \ourmethod improves the accuracy of ViT-L/14 by 15 to 60 percentage points, while accuracy on ImageNet decreases by less than one percentage point.
\ourmethod also allows practitioners to control the accuracy trade-off on the \bothtasks without re-training a new model, by varying the mixing coefficient $\alpha$.

\subsection{The effect of scale}
\label{sec:scale}

We consistently observe that \ourmethod is more effective for larger models. Our findings are aligned with those of \citet{ramasesh2021effect}, who observed that larger models are less susceptible to catastrophic forgetting.
This section formalizes and provides insights for these observations.

\begin{figure}%
    \centering
    \includegraphics[width=\linewidth]{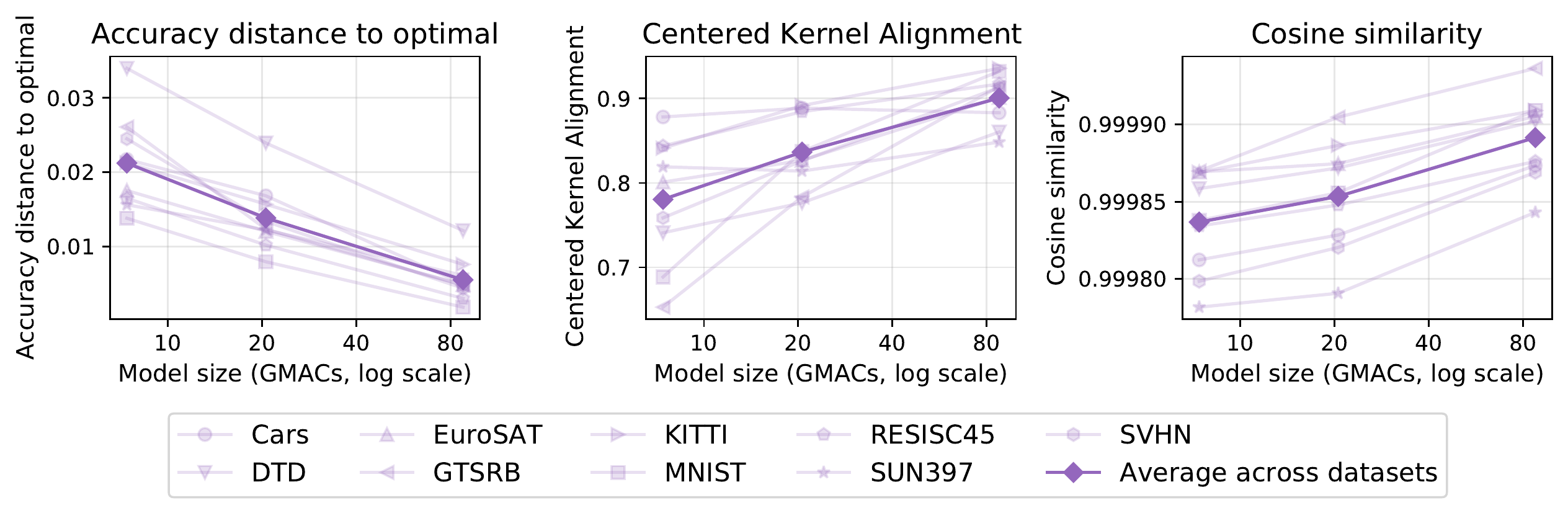}
    \caption{\textbf{Larger models are easier to patch} (left). 
    For larger models, the unpatched and fine-tuned model are more similar with respect to their representations (center) and weights (right). Model scale is measured in Giga Multiply-Accumulate operations (GMACs).}%
    \label{fig:scaling_metrics}%
\end{figure}

\paragraph{Measuring the effectiveness of patching.} We measure the effectiveness of patching via the accuracy difference 
between the single patched model and two specialized models
with the same architecture and initialization.
For both the \oldtask and \newtask, we take specialized models that maximize performance on the task, considering the set of all interpolations between the zero-shot and fine-tuned models.
We refer to this measure as \textit{accuracy distance to optimal}.
Formally, accuracy distance to optimal is given by
\begin{equation}
    \frac{1}{2}\left[ \max_\alpha \mathsf{Acc}(\theta_\alpha, \mathcal{D}_{\textrm{supp}}) + \max_\alpha \mathsf{Acc}(\theta_\alpha, \mathcal{D}_{\textrm{patch}}) \right] - \frac{1}{2}\max_\alpha\left[\mathsf{Acc}(\theta_\alpha, \mathcal{D}_{\textrm{supp}}) + \mathsf{Acc}(\theta_\alpha, \mathcal{D}_{\textrm{patch}}) \right],   
\end{equation}
where $\textrm{Acc}(\theta, \mathcal{D})$ represents the accuracy of model $\theta$ on task $\mathcal{D}$.
In Figure \ref{fig:scaling_metrics} (left), we show that accuracy distance to optimal decreases with scale, indicating that patching becomes more effective for larger models.

\paragraph{Model similarity.} Fine-tuning modifies overparameterized models less~\cite{chizat2019lazy}, which provides insights on why larger models are easier to patch: less movement is required to fit new data.
We demonstrate this by evaluating representational similarity using Centered Kernel Alignment (CKA) \cite{kornblith2019similarity} (see Appendix~\ref{sec:appendix_scaling_details} for details).
As shown in Figure~\ref{fig:scaling_metrics}~(center), the representations of the unpatched and fine-tuned models become more similar as models grow larger, indicated by larger CKA values.
Moreover, Figure~\ref{fig:scaling_metrics}~(right) shows that
the cosine similarity between the weights of the unpatched and fine-tuned models, $\cos(\theta_\textrm{zs},\theta_\textrm{ft}) = \langle\theta_\textrm{zs}, \theta_\textrm{ft}\rangle/(||\theta_\textrm{zs}||\,||\theta_\textrm{ft}||)$, increases with scale.

\begin{figure}%
    \centering
    \includegraphics[width=\linewidth]{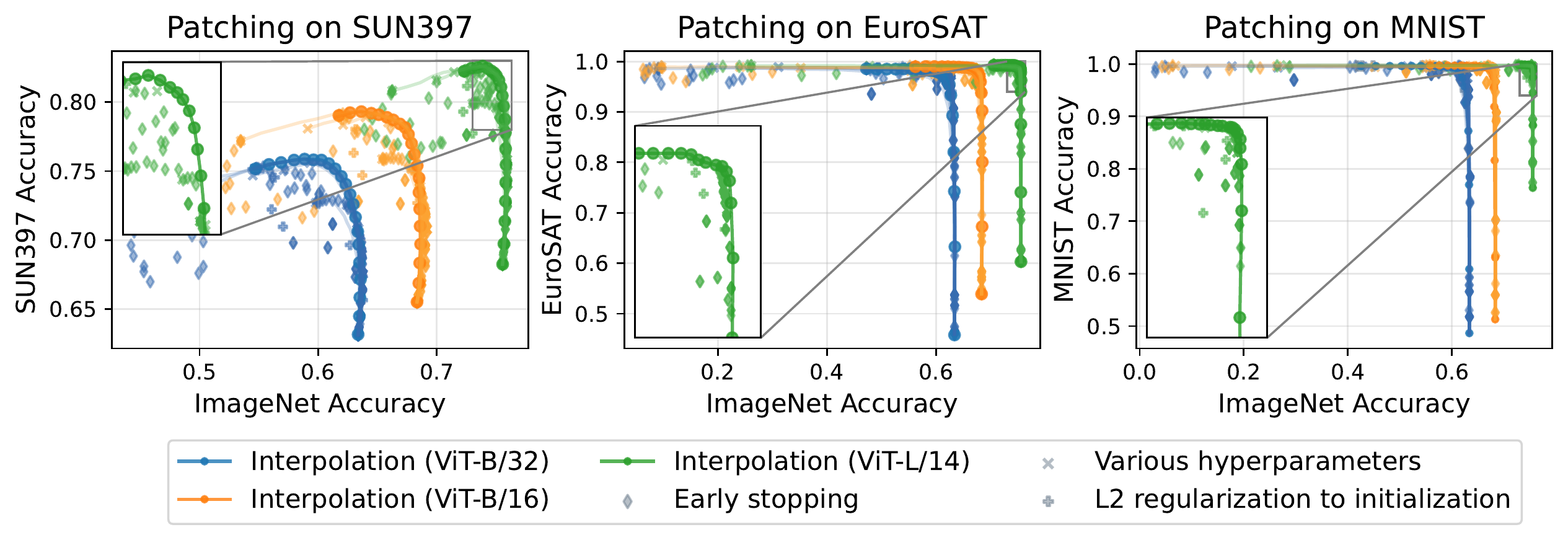}
    \vspace*{-0.3cm}
    \caption{\textbf{The frontier of accuracy trade-offs can be recovered by linearly interpolating weights.} 
    Interpolating the unpatched and fine-tuned models recovers the accuracy trade-off of early stopping, regularization towards the initialization, and changes in hyperparameters.
    Additional details and comparisons can be found in Appendix \ref{sec:appendix_additional_baselines}.}%
    \label{fig:baselines}%
    \vspace*{-0.3cm}
\end{figure}
\subsection{Baselines and ablations}
\label{sec:ablations}

\paragraph{Baselines.} 
There are many alternatives which enable a trade-off between accuracy on the supported and patching tasks.
These methods include early stopping during fine-tuning, applying a regularization term
which penalizes movement from initialization, or training with different hyperparameters including a smaller learning rate.
Unlike interpolation, these methods do not enable navigating the accuracy trade-off without fine-tuning the model again many times.
Moreover, Figure~\ref{fig:baselines} demonstrates that the accuracy trade-off frontier for early stopping, regularization, or varying hyperparameters can be recovered by interpolating weights with different mixing coefficients.
Appendix~\ref{sec:appendix_additional_baselines} provides additional baselines and discussion, including EMA \cite{szegedy2016rethinking}, EWC \cite{kirkpatrick2017overcoming}, LwF \cite{li2017learning}, re-training a model with data from the patching task, and mixing the pre-training and fine-tuning objectives.

\begin{figure*}
    \centering
    \includegraphics[width=\textwidth]{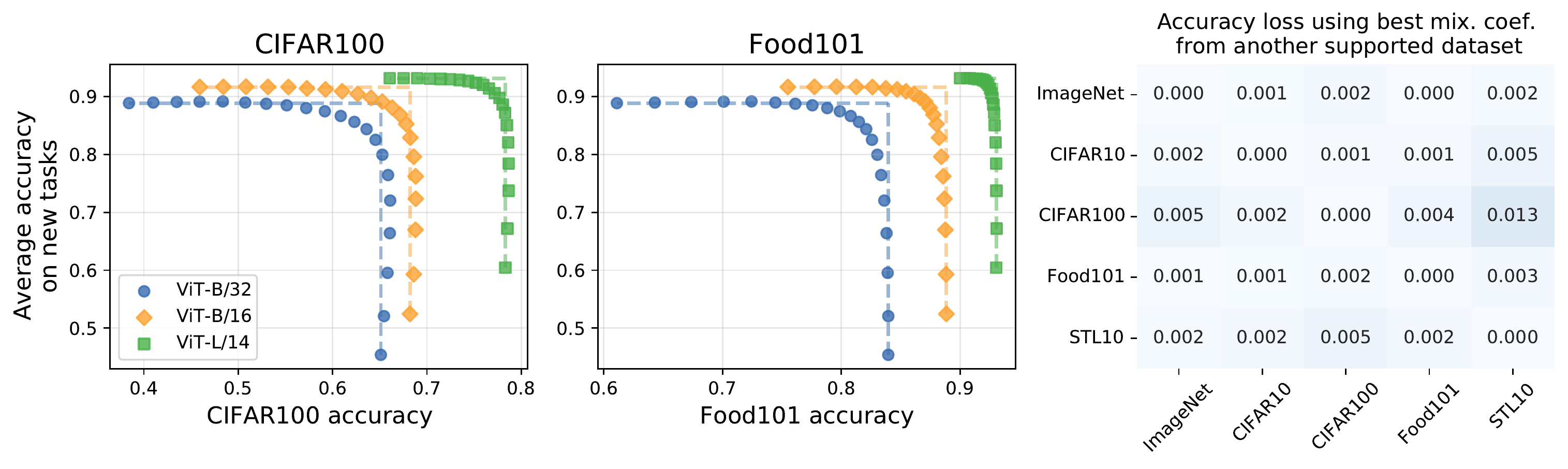}
        \vspace*{-0.5cm}
    \caption{\textbf{Results are consistent across \oldtasks.} For multiple \oldtasks, we observe similar accuracy improvements on \newtasks, without substantially decreasing \oldtask accuracy. Additional results for the supported tasks Food101, STL10 and ImageNet are in Appendix \ref{sec:appendix_additional_sup_tasks}. Moreover, choosing the mixing coefficients using a different \oldtask does not substantially decrease combined accuracy on \bothtasks (right). 
    }
    \label{fig:additional_sup_tasks}
        \vspace*{-0.3cm}
\end{figure*}

\paragraph{Additional \oldtasks.} In Figure~\ref{fig:main_scatter}, we use ImageNet as a representative \oldtask. This section demonstrates that \ourmethod is stable under different choices of the \oldtask.
Instead of ImageNet, we use CIFAR10, CIFAR100, Food101 and STL10.
Figure~\ref{fig:additional_sup_tasks} displays representative  results, where performance is averaged over the nine \newtasks (see Appendix~\ref{sec:appendix_additional_sup_tasks} for additional results).
We observe consistent results across \oldtasks, and that the optimal mixing coefficients are stable across different choices of supported tasks (Figure~\ref{fig:additional_sup_tasks}, right).

\paragraph{Additional models.} In addition to the CLIP ViTs used in the majority of our experiments, we study four ResNet models \cite{he2016deep} from \citet{radford2021learning} in Appendix \ref{sec:appendix_resnets}.
We find that patching is less effective for ResNets compared to ViTs of similar size, which
corroborates the findings of \citet{ramasesh2021effect} that ResNets are generally more susceptible to catastrophic forgetting.
However, similarly to ViTs, we still observe improvements with scale.
Finally, we show that patching is also effective for closed-vocabulary models in Appendix \ref{sec:appendix_closed}.

\section{Patching models on multiple tasks}
\label{sec:multi_patches}

This section details experimental results for patching on multiple datasets.
Recall from Section~\ref{sec:method} that there are various strategies for extending \ourmethod to multiple datasets, which we briefly revisit.
For \emph{joint} patching we merge all the datasets into a single fine-tuning task and apply our patching procedure as before.
For \emph{sequential} patching we iteratively perform our procedure once per task, using the patched model at each step as the initialization for the next step.\footnote{The results are averaged over three random seeds that control the order in which tasks are seen.}
We also explore \emph{parallel} patching, for which we have an unpatched model $\theta_\textrm{zs}$ and independently fine-tune on each of the tasks in parallel.
We then search for mixing coefficients to combine the resulting models. 
For tasks $1,...,k$, let $\theta_\textrm{ft}^{(1)}, ..., \theta_\textrm{ft}^{(k)}$ denote the fine-tuned models for each task.
Since it is impractical to exhaustively search over each $\alpha_i$, we instead search over a one-dimensional scalar $\alpha \in [0,1]$, which interpolates between $\theta_\textrm{zs}$ and the average of all fine-tuned solutions $\frac{1}{k}\sum_{i=1}^{k} \theta_\textrm{ft}^{(i)}$.\footnote{We also explored adaptive black-box optimization algorithms to choose the mixing coefficients $\alpha_i$ \cite{nevergrad}, but observed little improvement (0.3 to 0.4 percentage points on average).}
Appendix \ref{sec:appendix_multi} provides further experimental details.

These methods have various trade-offs and may be applicable for different scenarios.
Joint patching is only possible when data from all tasks you wish to patch is available.
On the other hand, sequential patching is appropriate when the tasks are observed one after another.
Finally, parallel patching can leverage distributed hardware.

\begin{figure*}
    \centering
    \includegraphics[width=\textwidth]{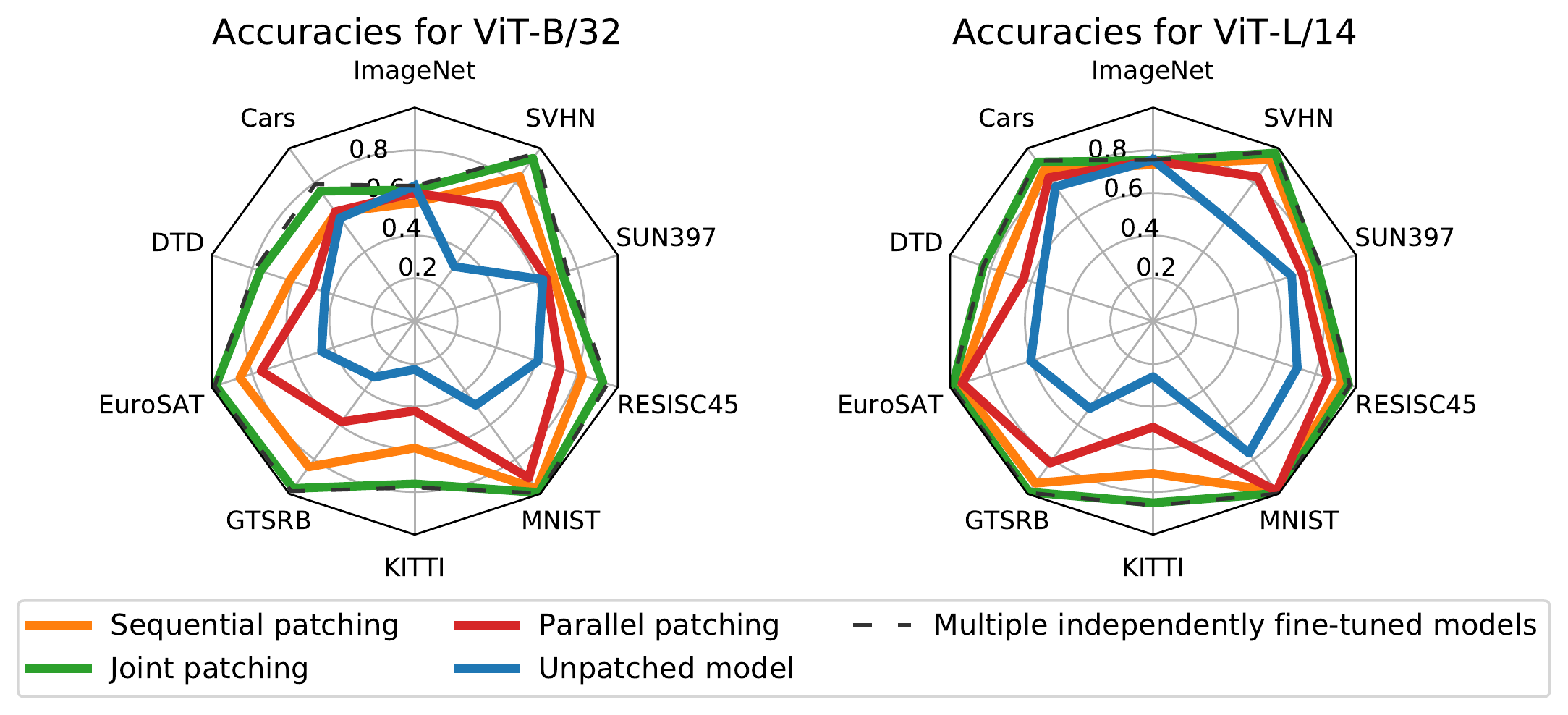}
    \caption{\textbf{Contrasting various strategies for patching on multiple tasks.} On all experiments, ImageNet is used as the supported task while the other nine datasets are used for patching. When data from all patching tasks is available, joint patching yields a single model that is competitive with using ten different specialized models. Weight interpolations greatly mitigate catastrophic forgetting on the sequential case, but do not completely eradicate it. Finally, parallel patching underperforms other patching strategies, but still provides improvements over the unpatched model.}
    \label{fig:multi_task}
        \vspace*{-0.3cm}
\end{figure*}

Figure \ref{fig:multi_task} displays experimental results when patching on all nine tasks from Section~\ref{sec:patching}.
We observe that joint patching is the best-performing method on average.
This is perhaps unsurprising since joint patching has simultaneous access to all patching datasets, unlike other patching strategies.
Nevertheless, it is still interesting that for ViT-L/14, joint patching yields a \emph{single} model with only 0.5 percentage points worse combined accuracy than using \emph{multiple} specialized models.\footnote{Recall from Section~\ref{sec:setup} that combined accuracy weight \bothtasks equally.}
Joint patching also achieves a 15.8 percentage points improvement over the unpatched model.
Moreover, patching a ViT-B/32 model with the joint strategy achieves a combined accuracy 6.1 percentage points higher than a ViT-L/14 unpatched model, which requires 12x more GMACs.

The accuracy of sequential patching approaches that of joint patching, especially for larger models.
Note that, unlike in joint patching, forgetting can compound since the patching procedure is applied multiple times in sequence.
In sequential patching, weight interpolations do not completely eradicate forgetting, but greatly mitigate it.
This is most noticeable for smaller models: sequentially fine-tuning a ViT-B/32 without interpolation \textit{reduces} the combined accuracy by 4.6 percentage points compared to the unpatched model, as shown in Appendix \ref{sec:appendix_multi}. This is compared to a combined accuracy \textit{increase} of 11 percentage points when using sequential patching. Additional results, including experiments on SplitCIFAR \cite{rebuffi2017icarl}, can be found in Appendix~\ref{sec:appendix_multi}.

Finally, parallel patching underperforms other patching strategies.
Like sequential patching, parallel patching is in the challenging setting where data from all patching tasks is not available simultaneously.
Moreover, unlike in joint or sequential patching, no model is optimized on data from all patching tasks.
Using a black box optimization algorithm for finding the mixing coefficients did not yield large improvements over using the same mixing coefficient for all models.
However, it is possible that more sophisticated search methods could yield better results.
In Appendix~\ref{sec:appendix_multi}, we present additional experiments for a subset of the tasks where exhaustively searching the space of mixing coefficients is tractable, finding headroom for improvement in most cases. 

\begin{table*}
\setlength\tabcolsep{3.2pt}
\small
\begin{tabular}{lccccccccc}
{} &  Cars &   DTD &  EuroSAT &  GTSRB & KITTI & MNIST &  RESISC45 &  SUN397 &  SVHN \\
\midrule
Unpatched accuracy  &  86.2 &  64.9 &     79.9 &   51.7 &   43.4 &   82.6 &      73.4 &    76.9 &  72.8  \\\midrule
\multirow{2}{*}{Patched accuracy} &  87.0  &  66.1 &     87.2 &   71.1 &   60.4 &   91.3 &      74.2 &    79.3 &  88.9  \\
&   {\textbf{\footnotesize\textcolor{LimeGreen}{(+0.8)}}} &   {\textbf{\footnotesize\textcolor{LimeGreen}{(+1.2)}}} &      {\textbf{\footnotesize\textcolor{LimeGreen}{(+7.3)}}} &   {\textbf{\footnotesize\textcolor{LimeGreen}{(+19.4)}}} &   {\textbf{\footnotesize\textcolor{LimeGreen}{(+17.0)}}} &   {\textbf{\footnotesize\textcolor{LimeGreen}{(+8.7)}}} &     {\textbf{\footnotesize\textcolor{LimeGreen}{(+0.8)}}} &     {\textbf{\footnotesize\textcolor{LimeGreen}{(+2.4)}}} &  {\textbf{\footnotesize\textcolor{LimeGreen}{(+16.1)}}}  \\\bottomrule
\end{tabular}
\caption{\textbf{\ourmethod can generalize to unseen classes.} We randomly partition each dataset into tasks $A$ and $B$ with disjoint class spaces of roughly equal size.
This table reports how patching on task $A$ affects accuracy on task $B$ for the ViT-L/14 model.
In all cases, accuracy on task $B$ improves when patching on task $A$ even though the classes are \textit{unseen} during patching.}
\vspace{-10pt}
\label{tab:unseen_gen}
\end{table*}

\section{Broad transfer}
\label{sec:transfer}

An alternative to our patching approach is to introduce parameters which are specific to each new task.
By contrast, \ourmethod always maintains a single model.
This section describes an additional advantage of the single model approach:
patching the model on task $A$ can improve accuracy on task $B$, even when task $A$ and $B$ do not share the same classes.
We refer to this phenomenon as \emph{broad transfer}.
Note that we are able to study this phenomenon because the single patched model remains open-vocabulary throughout the patching procedure.
This is a key advantage of \ourmethod compared to maintaining a collection of task-specific models.

We now describe two experiments to measure the effects on a task $B$ when patching the model on a task $A$.
First, we explore broad transfer by randomly partitioning datasets into disjoint sets with no class overlap.
For a dataset $\mathcal{D}$ we partition the class space $\mathcal{Y}$ into two disjoint sets of roughly equal size $\mathcal{Y}_A$ and $\mathcal{Y}_B$.
We build task $A$ with the examples $(x,y) \in \mathcal{D}$ where $y$ belongs to $\mathcal{Y}_A$, and task $B$ with examples $(x,y)$ where $y$ belongs to $\mathcal{Y}_B$.
Table~\ref{tab:unseen_gen} shows how patching a model on task $A$ affects the accuracy on task $B$ for nine datasets $\mathcal{D}$.
The accuracy improvements on task $B$ range from 0.8 to 19.4 percentage points, even though the classes from task $B$ are not seen during patching. 

\begin{table*}
\setlength\tabcolsep{1.0pt}
\begin{center}
\small
\begin{tabular}{lcccccccc}
Task $A$ & MNIST & SVHN & EuroSAT & RESISC45 & MNIST & FashionMNIST & GTSRB & MTSD \\
Task $B$ & SVHN & MNIST & RESISC45 & EuroSAT & FashionMNIST & MNIST & MTSD & GTSRB\\
\midrule
Unpatched accuracy & 58.6 & 76.4 & 71.0 & 60.2 & 67.7 & 76.4& 19.3 & 50.6 \\\midrule
\multirow{2}{*}{Patched accuracy} & 68.9 & 93.2 & 69.7 & 70.4 & 70.8 & 77.5 & 30.8 & 69.8\\
 & {\textbf{\footnotesize\textcolor{LimeGreen}{(+10.3)}}} & {\textbf{\footnotesize\textcolor{LimeGreen}{(+16.8)}}} & {\textbf{\footnotesize\textcolor{FireBrick}{(-1.3)}}} & {\textbf{\footnotesize\textcolor{LimeGreen}{(+10.2)}}} & {\textbf{\footnotesize\textcolor{LimeGreen}{(+3.1)}}} & {\textbf{\footnotesize\textcolor{LimeGreen}{(+1.1)}}} & {\textbf{\footnotesize\textcolor{LimeGreen}{(+11.5)}}} & {\textbf{\footnotesize\textcolor{LimeGreen}{(+19.2)}}}\\
\bottomrule
\end{tabular}
\caption{\textbf{Patching on task $A$ can improve accuracy on a related task $B$.} For a pair of tasks $A$ and $B$, we report accuracy of the ViT-L/14 on task $B$, after patching on task $A$, finding improvements on seven out of eight cases.
}
\vspace{-6pt}
\label{tab:transfer_pairs}
\end{center}
\end{table*}

To further understand transfer, we consider additional task pairs $A$ and $B$, which are now different datasets.
While some pairs $A$, $B$ share classes, there are still instances of broad transfer.
Concretely, Table~\ref{tab:transfer_pairs} examines i) MNIST and SVHN, two digit recognition tasks with shared classes; 
ii) EuroSAT and RESISC45, two satellite imagery recognition tasks where there are unshared classes but some overlap; 
iii) GTSRB and  MTSD \cite{ertler2020mapillary}, two traffic sign recognition datasets where there are unshared classes but some overlap; 
and iv) MNIST and FashionMNIST \cite{fashionmnist}, which do not share any classes but appear visually similar.
In seven out of eight experiments, patching on task $A$ improves accuracy by 1.1 to 19.2 percentage points on task $B$.
The exception is when $A$ is EuroSAT and $B$ is RESISC45, where accuracy decreases by 1.3 percentage points.

In all experiments, when patching on task $A$ we choose the mixing coefficient $\alpha$ by optimizing the held-out validation accuracy on task $A$ and a supported task (in this experiment we use ImageNet).
While it is possible for a method that introduces new parameters for each task to exhibit broad transfer to new data,
this also requires knowing which parameters to apply for the new data.
This is not necessary in the single model approach.

\section{Case studies}
\label{sec:case_studies}

We further examine the performance of \ourmethod in three additional settings, which
highlight weaknesses of the zero-shot CLIP model and showcase broad transfer (Section~\ref{sec:transfer}).

\begin{figure*}
    \centering
    \includegraphics[width=\textwidth]{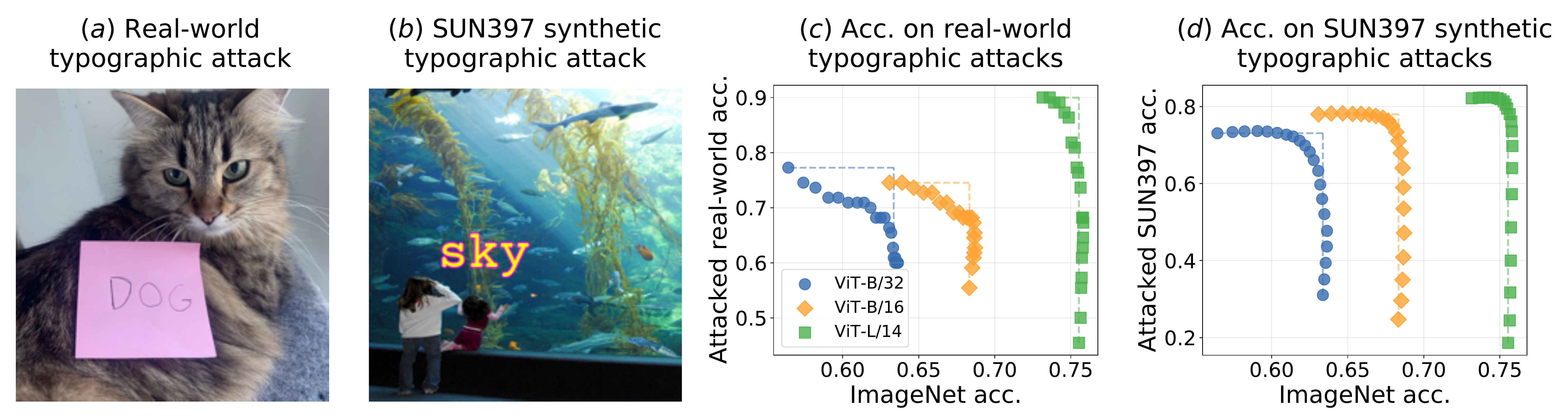}
    \caption{\textbf{Guarding against real-world typographic attacks by patching on synthetic data.}
    \emph{(a)} A sample from our real-world typographic attacks test set. A CLIP ViT-L/14 is ``tricked'' into classifying this image as a dog instead of a cat.
    \emph{(b)} Sample of synthetic typographic attack data.
    \emph{(c)} Performance on real-world data with unseen classes after patching on \emph{only} synthetic typographic attacks (curves produced by interpolating between the unpatched and fine-tuned model).
    \emph{(d)} Analogous curves for the test set of the synthetic data used for patching.
    }
    \vspace{-4pt}
    \label{fig:typo}
\end{figure*}

\paragraph{Typographic attacks.} \citet{goh2021multimodal} find that CLIP models are susceptible to \emph{typographic attacks}, where text superimposed on an image leads to misclassification.
For example, in Figure ~\ref{fig:typo}~\emph{(a)}, the text on the pink note saying ``dog'' leads a CLIP to misclassify the image of a cat as a dog.
To fix this vulnerability,
we procedurally generate typographic attack data by adding text with incorrect class names to SUN397~\cite{sun397}, as seen in Figure~\ref{fig:typo}~\textit{(b)}.
We then collect a test set of 110 real world images by placing notes on objects and taking photos.\footnote{Data available at \url{https://github.com/mlfoundations/patching}.}
After applying \ourmethod using the synthetic data, we evaluate on the real-world images
(Figure~\ref{fig:typo}~\textit{(c)}) and synthetic test set (Figure~\ref{fig:typo}~\textit{(d)}).
We observe that while larger models are more susceptible to typographic attacks, they are also more amenable to patching.
Furthermore, we see an example of broad transfer between the synthetic and real-world data: when patching ViT-L/14 on synthetic data, its accuracy on real-world typographic attacks improves 41 percentage points even though the real-world classes are unseen.
The cost is a reduction of less than 1 percentage point on ImageNet.
We present details on the task and data collection in Appendix \ref{sec:appendix_typo}.

\paragraph{Counting.} \citet{radford2021learning} find that CLIP models struggle to count the number of objects in CLEVR~\cite{clevr}.
Here, the task is to choose an integer between 3 and 10 for each image, corresponding to the number of visible objects.
While a straightforward way to patch such a task is to fine-tune on it directly, we investigate if applying \ourmethod using a subset of the classes allows the patched model to generalize to other numbers.
Specifically, we patch on images with 4, 5, 6, 8, or 9 objects.
To evaluate broad transfer, we test on images with 3, 7, and 10 objects (7 for understanding interpolation and 3 and 10 for extrapolation).
We find that \ourmethod improves accuracy from 59\% to over 99\% on unseen classes with less than half a percentage point decrease in ImageNet accuracy.
For more details see Appendix~\ref{sec:appendix_counting}.

\paragraph{Visual question answering.} As shown by \citet{shen2021much}, zero-shot CLIP models perform poorly on visual question answering~\cite{vqa}. 
Using CLIP for VQA typically involves additional parameters---for instance, \citet{shen2021much} trains a transformer \cite{vaswani2017attention} on CLIP features.
In contrast, our procedure for patching CLIP on VQA does not introduce new parameters.
Following \citet{shen2021much}, we contrast images with a series of text prompts, where each prompt corresponds to an option in multiple-choice VQA, formed by both the question and a candidate answer using the following template: ``Question: [question text] Answer: [answer text]''.
We evaluate on multiple-choice VQA v1 \cite{vqa}, where each question is associated with 18 candidate answers.
Our results, further detailed in Appendix \ref{sec:appendix_vqa}, show that patching is effective for visual question answering: \ourmethod improves the accuracy of a ViT-L/14 model by 18 percentage points, while accuracy drops by less than one percentage point on ImageNet.

\section{Related work}
\label{sec:related}

\paragraph{Continual learning and catastrophic forgetting.}
Learning tasks sequentially remains a challenge for neural networks. 
When a neural network learns a new task, the accuracy on other tasks often decreases, a phenomenon known as \emph{catastrophic forgetting}~\cite{mccloskey1989catastrophic,thrun1998lifelong,french1999catastrophic,kirkpatrick2017overcoming}.
While forgetting in neural networks may actually aid learning~\cite{zhou2022fortuitous}, researchers have proposed various approaches for alleviating catastrophic forgetting, including: i) Regularization-based approaches such as elastic weight consolidation (EWC) \cite{kirkpatrick2017overcoming} and synaptic intelligence (SI)~\cite{zenke2017continual} which
penalize the movement of parameters and are related to weight-interpolation by \citet{lubana2021quadratic};
ii) Replay methods~\cite{rebuffi2017icarl,shin2017continual,lopez2017gradient,chaudhry2018efficient,rolnick2019experience,mirzadeh2020linear}, which
incorporate data or gradient information from previous tasks when learning a new task; and iii) Introducing task-specific parameters
\cite{rusu2016progressive,yoon2017lifelong,mallya2018piggyback,cheung2019superposition,Oswald2020Continual,wortsman2020supermasks}.

In contrast to these approaches, \ourmethod requires no modification to the standard fine-tuning process besides the later weight interpolation step.
Moreover, unlike regularization or replay based methods, \ourmethod requires no extra computational cost during training.
In contrast to methods with task specific parameters, we maintain a single model.
Having a single model is beneficial when there is new data which is similar to one of the tasks which have already been patched.
Even without explicitly knowing which task the new data is similar to, we can observe accuracy improvements (see  Section~\ref{sec:transfer}).

Similar to our work is that of ~\citet{mirzadeh2020linear}, who observe high accuracy on task A on the linear path between a model which achieves high accuracy on task A and a model which is fine-tuned jointly on task A and B. Moreover, they observe high accuracy on task B on the linear path between a model fine-tuned on task B, and the jointly fine-tuned model.
Therefore, there exists a path between a model which achieves good performance on task A and a model fine-tuned on task B along which accuracy is high on both tasks.
However, in \citet{mirzadeh2020linear} this combined path can be non-linear, leading them to propose a regularization and replay based method.
In our work, we find that examining models on a linear path between the unpatched model (which has high accuracy on task A) and the model fine-tuned on task B is often sufficient for obtaining a model which achieves high accuracy on both tasks (Figure \ref{fig:main_scatter}).
We speculate that this is due to scale and model architecture:
in contrast to~\citet{mirzadeh2020linear}, we initialize with a model pre-trained on a large dataset consisting of 400 million images \cite{radford2021learning}, and primarily use vision transformers~\cite{dosovitskiy2021an}.
As shown in Section \ref{sec:ablations}, our method performs substantially worse with ResNets~\cite{he2016deep}, which are used by
\citet{mirzadeh2020linear}. 

Finally, \citet{ramasesh2021effect} and \citet{mehta2021empirical} also observed that catastrophic forgetting is less problematic for large and pre-trained models.
In addition, \citet{ramasesh2021effect} found---similar to our results---that vision transformers are less susceptible to forgetting than ResNets of the same size.

\paragraph{Linear mode connectivity and robust fine-tuning.}

Linearly interpolating neural network weights is a key step in PAINT.
Because of the many nonlinear activations in a neural network, it is not clear a priori that linearly interpolating
between two sets of weights can result in a high accuracy solution.
However, researchers have observed that interpolating neural network weights can
achieve high accuracy when training on MNIST from a common initialization~\cite{nagarajan19uniform} or when part of the optimization trajectory is shared~\cite{frankle2020linear,izmailov2018averaging,neyshabur2020being,fort2020deep,wortsman2021robust,matena2021merging,entezari2021role,wortsman2022model,choshen2022fusing}.
The term linear mode connectivity was coined by \citet{frankle2020linear}: two networks exhibit linearly mode connectivity if the accuracy does not
decrease when using weights on the linear path between them \cite{nagarajan19uniform, frankle2020linear}. Weight averaging for continual learning has also been studied by \citet{lee2017overcoming} for closed-vocabulary models.

While~\citet{nagarajan19uniform} and \citet{frankle2020linear} focused on accuracy on a single task, \citet{wortsman2021robust} use linear mode connectivity to fine-tune models while preserving their robustness to natural distribution shifts.
By interpolating the weights of a zero-shot and fine-tuned model, they find a solution which performs well both on the fine-tuning task and under distribution shift.
In contrast to \citet{wortsman2021robust}, we do not modify any task-specific parameters when fine-tuning, preserving the open-vocabulary nature of the models we patch.
Unlike \citet{wortsman2021robust}, we examine accuracy trade-offs across different tasks with little or no class overlap and adapt a model to multiple tasks.

In addition, closely related to our work is that of \citet{matena2021merging}, who use Fisher-weighted averaging of language models before and after fine-tuning on downstream tasks.
Unlike Fisher-weighted averaging of \citet{matena2021merging}, we do not use different mixing coefficients for each parameter, and thus require no extra compute when patching.
Moreover, we explore new strategies for patching on multiple tasks (see Section \ref{sec:multi_patches}), and focus on open-vocabulary image classifiers.

\paragraph{Interventions to change the behavior of a trained model.}
Several authors have studied the problem of updating a model to locally alter its behavior on certain inputs without external disruptions on other inputs \cite{sinitsin2020editable,de2021editing,mitchell2021fast,shibani2021editing,ribeiro2020beyond,ribeiro2022adaptive}.
Previous literature uses various terms to refer to this process, including model editing, patching or debugging.
A popular use case is to update trained language models to reflect changes in the world (for instance, facts like who is the current president of Brazil) \cite{jang2022temporalwiki,luu2021time,lazaridou2021mind,jang2021towards}.
Moreover, inspired by software engineering practice, previous work explored ``debugging'' language models through user interaction \cite{ribeiro2020beyond,ribeiro2022adaptive}, including providing corrective feedback to the models via natural language \cite{anonymous2022fixing}.
\citet{mitchell2021fast,de2021editing} propose training auxiliary networks to perform local edits on pre-trained models.
\citet{shibani2021editing} introduce a method for rewriting the prediction rules of a classifier, focusing on specific failure modes such as reliance on spurious correlations.
In contrast with previous literature, our work explores patching models at the \textit{task} level, aiming to systemically improve accuracy on a dataset---for instance, enabling a model to recognize dozens of satellite imagery classes with a single patch.

\section{Limitations and conclusion}
\label{sec:conclusion}
\textbf{Limitations.} When applying \ourmethod, accuracy on \oldtasks can still decrease, especially for smaller models.
This limitation is perhaps best reflected in the case of sequential patching:
patched models underperform using multiple specialized models when many tasks are added sequentially.
Using larger models and weight interpolations can alleviate this issue, but do not completely resolve it.
Finally, better understanding on which datasets patching is more effective is an exciting direction for future research.

\textbf{Conclusion.}
In this work, we explore several techniques for patching open-vocabulary models with the goal of improving accuracy on new tasks without decreasing accuracy elsewhere.
\ourmethod is effective in several scenarios, ranging from classifying digits to defending against typographic attacks.
\ourmethod becomes more effective with scale, and can be applied on multiple tasks sequentially or simultaneously.
Our findings demonstrate that in many circumstances it is possible to expand the set of tasks on which models achieve high accuracy, without introducing new parameters, without re-training them from scratch, and without catastrophic forgetting.

\section*{Acknowledgments}

We thank Akari Asai, Alex Fang, David Fleet, Huy Ha, Ari Holtzman, Pieter-Jan Kindermans, Marco Tulio Ribeiro, Ofir Press, Sarah Pratt, Sewon Min, Thao Nguyen and Tim Dettmers for helpful discussions and feedback, and Hyak at UW for computing support.
This work is in part supported by the NSF AI Institute for Foundations of Machine Learning (IFML), Open Philanthropy, NSF IIS 1652052, NSF IIS 17303166, NSF IIS 2044660, NSF IIS 2132519, ONR N00014-18-1-2826, DARPA N66001-19-2-4031, DARPA W911NF-15-1-0543, the Sloan Fellowship and gifts from Allen Institute for AI.

{
\bibliographystyle{plainnat}
\bibliography{main}
}



\clearpage
\appendix

\section{Dataset details}
\label{sec:appendix_datasets}

In Table \ref{tab:datasets}, we present the number of classes and the size of the training, validation and test sets we use for the each patching and supported tasks: Stanford Cars \cite{cars}, Describable Textures (DTD) \cite{dtd}, EuroSAT \cite{eurosat}, German Traffic Sign Recognition Benchmark (GTSRB) \cite{gtsrb}, KITTI distance \cite{kitti}, MNIST \cite{lecun1998mnist}, RESISC45 \cite{cheng2017remote}, SUN397 \cite{sun397}, SVHN \cite{svhn}
ImageNet \cite{deng2009imagenet}, FashionMNIST \cite{fashionmnist}, MTSD \cite{ertler2020mapillary}, CIFAR10 \cite{krizhevsky2009learning}, CIFAR100 \cite{krizhevsky2009learning}, Food101 \cite{food101}, STL10 \cite{stl10} and ImageNet \cite{deng2009imagenet}.
For datasets that did not have publicly available, labeled test sets, we use the validation set as the test set, and split the training set into training and validation sets in a stratified fashion.
For all these datasets, we use accuracy as our evaluation metric.

\begin{table*}
\setlength\tabcolsep{5.2pt}
\centering
\small
\begin{tabular}{lrrrr}
\toprule
& \multicolumn{3}{c}{Size of the set used for} & \\
Dataset & Training & Validation & Testing & Number of classes\\
\midrule
Cars \cite{cars} & 7,330 & 814 & 8041 & 196\\
DTD \cite{dtd} & 3,384 & 376 & 1,880 & 47\\
EuroSAT \cite{eurosat} & 21,600 & 2,700 & 2,700 & 10\\
GTSRB \cite{gtsrb} & 23,976 & 2,664 & 12,630 & 43\\
KITTI \cite{kitti} & 6,347 & 423 & 711 & 4\\
MNIST \cite{lecun1998mnist} & 55,000 & 5,000 & 10,000 & 10\\
RESISC45 \cite{cheng2017remote} & 17,010 & 1,890 & 6,300 & 45\\
SUN397 \cite{sun397} & 17,865 & 1,985 & 19,850 & 397\\
SVHN \cite{svhn} & 68,257 & 5,000 & 26,032 & 10\\
FashionMNIST \cite{fashionmnist} & 55,000 & 5,000 & 10,000 & 10\\
MTSD \cite{ertler2020mapillary} & 55,078 & 5,000 & 8,737 & 235\\
\midrule
CIFAR10 \cite{krizhevsky2009learning} & 45,000 & 5,000 & 10,000 & 10\\
CIFAR100 \cite{krizhevsky2009learning} & 45,000 & 5,000 & 10,000 & 100\\
Food101 \cite{food101} & 70,750 & 5,000 & 25,250 & 101\\
STL10 \cite{stl10} & 4,500 & 500 & 8,000 & 10\\
ImageNet \cite{deng2009imagenet} & 1,255,167 & 26,000 & 50,000 & 1,000\\
\bottomrule
\end{tabular}
\caption{\textbf{Dataset statistics} for patching and supported tasks.
When a set for validation is not available we use held-out data from the official training set for validation purposes.
In cases like ImageNet we use the official validation set for testing.
}
\label{tab:datasets}
\end{table*}

\section{Background on open-vocabulary models}
\label{sec:clip_bg}

In contrast to typical image classifiers, open-vocabulary models are not constrained to a fixed classification space.
Instead, they are able to perform any image classification task, by using textual descriptions of the class names.
Recently, many open-vocabulary models have been proposed \cite{radford2021learning,jia2021scaling,pham2021combined,zhai2022lit,yu2022coca,alayrac2022flamingo}.

A popular class of open-vocabulary models are contrastive image-text models like CLIP, BASIC and ALIGN.
Following \citet{radford2021learning} we use the term CLIP to refer to any contrastive image-text model.
This class of models are the focus of this work, although \ourmethod can naturally be extended to other open vocabulary models like Flamingo \cite{alayrac2022flamingo}.
CLIP models are trained to contrast images and textual descriptions, and are usually pre-trained on large, heterogeneous data collected from the web, ranging from hundreds of millions to billions of pairs of images and captions.
This class of models consists of a vision encoder $f$ which processes images and a text encoder $g$ which processes text.
Given a set of image-caption pairs $\{(x_1, y_1), ..., (x_k, y_k)\}$, the model is optimized to maximize the similarity of aligned pairs $\langle f(x_i), g(y_i)\rangle$ relative to unaligned pairs.

When performing a classification task, a set of captions $\{y_1, ..., y_k\}$ are procedurally generated based on text descriptions of the classes and some template.
For instance, for distinguishing between cats and dogs, the set of candidate captions could be
$\{$``an image of a dog'', ``an image of a cat''$\}$.
Then, given an image $x$, the chosen class is selected based on the caption that maximizes feature similarity with the image ($\argmax_i \langle f(x), g(y_i)\rangle$).
A common practice is to generate multiple candidate captions for each class (for instance, ``an image of a dog'' and ``a photo of a dog'') and average their representations before computing the similarities.
This practice, introduced by \citet{radford2021learning}, is known as prompt ensembling.

Adapting open-vocabulary models to downstream tasks is typically done by introducing task-specific parameters, mapping visual representations to a fixed class space determined by the downstream task.
The model can then be fine-tuned end-to-end, where both the visual encoder and the new classification head are updated, or adapted via training linear probe on top of the visual representations.
By contrast, we do not introduce or modify any task-specific parameters.
To do so, we adapt only the weights of the visual encoder, maintaining the text encoder without updating its weights.
Freezing the text encoder allows for faster fine-tuning, since the text features can be pre-computed once per task.
Moreover, in Appendix \ref{sec:frozen_head}, we show that the decision of freezing the text encoder does not substantially harm performance.
Importantly, this fine-tuning procedure allows us to adapt the model without loosing its ability to perform any image classification task.

\section{Frozen CLIP heads for fine-tuning}
\label{sec:frozen_head}

\begin{figure}
    \centering
    \includegraphics[width=\textwidth]{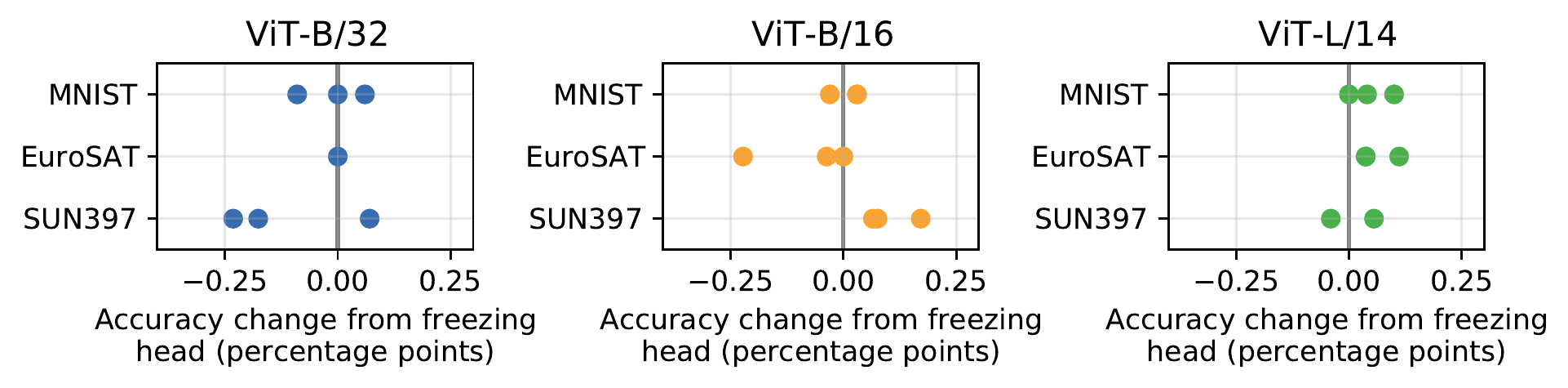}
    \caption{\textbf{Freezing the classification layer output by CLIP's text encoder has minimal effect on downstream accuracy.} Each experiment is repeated for three random seeds, shown as different points.}
    \label{fig:nofreeze}
\end{figure}

Instead of introducing a learnable classification layer, we use the frozen output of CLIP's text encoder to map image features to the space of classes.
Figure~\ref{fig:nofreeze} shows that this modification has a negligible effect on downstream accuracy for MNIST, EuroSAT, and SUN397.

\section{The effect of scale on patching}
\label{sec:appendix_scaling_details}

\begin{figure}%
    \centering
    \includegraphics[width=\linewidth]{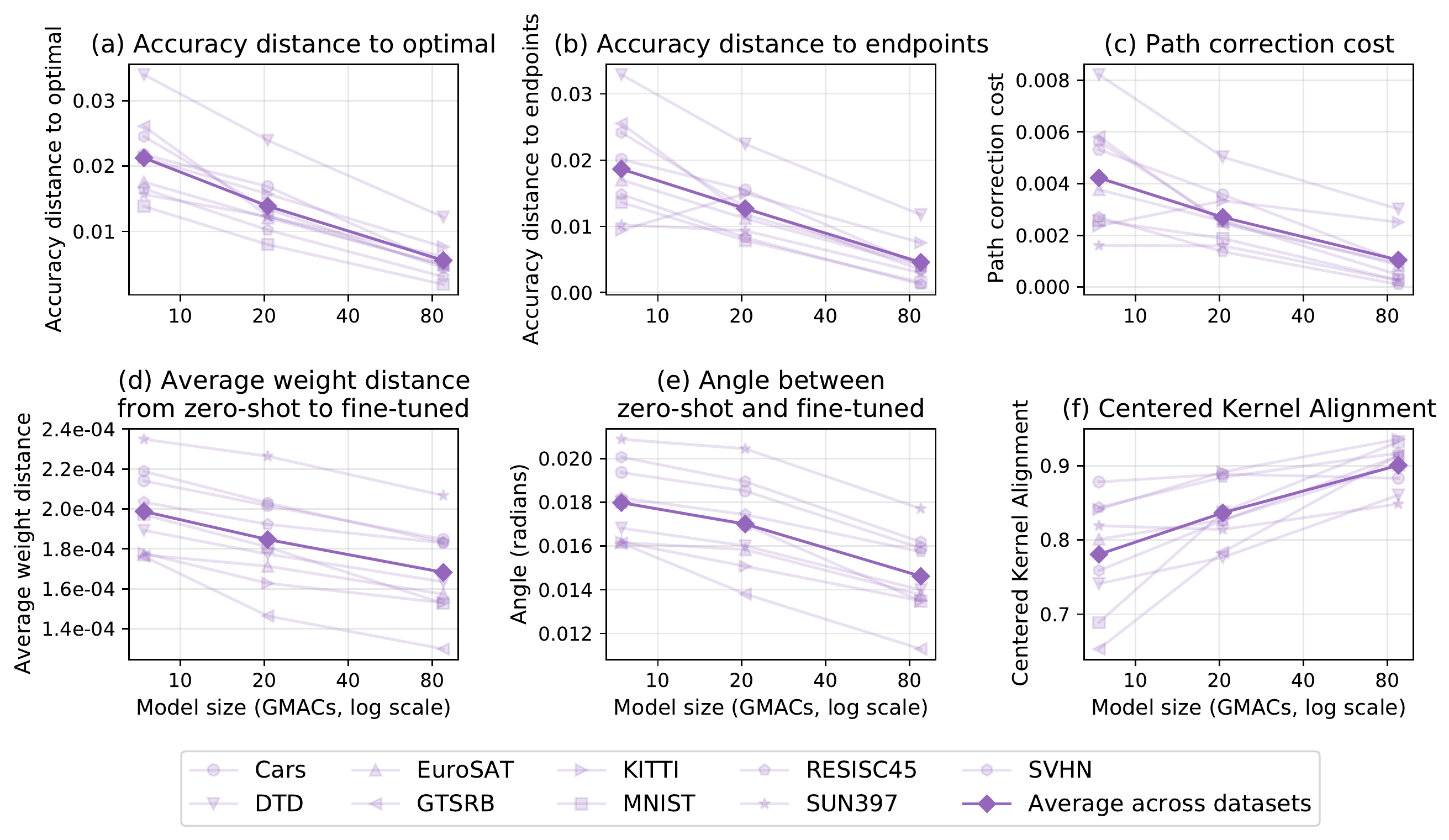}
    \caption{\textbf{The effect of scale on model patching}. (a-c) Patching is more effective for larger models. (d, e) Unpatched and fine-tuned models have more similar weights at scale. (f)
    For larger models, the unpatched and fine-tuned model are more similar with respect to their representations.}%
    \label{fig:app_scale}%
\end{figure}

This section introduces additional metrics for the effectiveness of patching and how similar the unpatched and fine-tuned models are.
Results are shown in Figure \ref{fig:app_scale}.

\subsection{Measuring the effectiveness of patching}

In addition to \textit{accuracy distance to optimal} presented in Section \ref{sec:scale}, we present two other metrics to measure the effectiveness of patching, \textit{accuracy distance to endpoints} and \textit{path correction cost}, which are summarized in Table \ref{tab:metrics}.
For simplicity, let $x_\alpha$ denote the accuracy of model $\theta_\alpha$ on the \oldtask and $y_\alpha$ on the \newtask.
For all metrics, lower values indicate that patching is more effective.

\textit{Accuracy distance to endpoints} contrasts the accuracy of a single model with using two specialized models, the fine-tuned model for the patching task and the unpatched model for the supported task.
Recall that $x_\alpha$ denotes the accuracy of model $\theta_\alpha$ on the \oldtask and $y_\alpha$ on the \newtask.
Accuracy distance to endpoints is given by $(x_0 + y_1)/2 - \max_\alpha(x_\alpha + y_\alpha)/2$.

\textit{Path correction cost} measures how the curve traced by varying the mixing coefficient compares to a curve without accuracy trade-offs. Its value is the average cost needed to move points in the traced curve to the ideal curve $\mathcal{I}$, where $\mathcal{I} = \{x, y \in [0,1]\times [0,1]\mid x = x_\textrm{zs} \vee y = y_\textrm{ft}\}$ is the set of points on the scatter plots where either accuracy on the \oldtask is equal to that of the unpatched model or accuracy on the \newtask is equal to that of the fine-tuned model.
More precisely, patch correction cost is given by, $\mathbb{E}_\alpha[\delta\left((x_\alpha, y_\alpha), \mathcal{I}\right) \mathbb{1}[x_\alpha{<}x_0 \wedge y_\alpha{<}y_1]]$, where $\delta((x,y), \mathcal{I}) = \min_{x',y' \in \mathcal{I}}||(x-x',y-y')||_2$ is the distance between $(x,y)$ and $\mathcal{I}$, and $\mathbb{1}[\,]$ is the indicator function.
The indicator term $\mathbb{1}[\,]$ ensures costs are only computed for points where both accuracy on the \oldtask is smaller than that of the unpatched model and accuracy on the \newtask is smaller than that of the fine-tuned model.

\begin{table}
\setlength\tabcolsep{3.6pt}
\begin{center}
\small
\begin{tabular}{c|c|c}
 Acc. distance to endpoints & Acc. distance to optimal & Path correction cost\\
 $\frac{x_0 + y_1}{2} - \max_\alpha\left(\frac{x_\alpha + y_\alpha}{2}\right)$ & $\frac{\max_\alpha x_\alpha + \max_\alpha y_\alpha}{2} - \max_\alpha\left(\frac{x_\alpha + y_\alpha}{2}
     \right)$ & $\mathbb{E}_\alpha[\delta\left((x_\alpha, y_\alpha), \mathcal{I}\right) \mathbb{1}[x_\alpha{<}x_0 \wedge y_\alpha{<}y_1]]$\\

\end{tabular}
\vspace{5pt}
\caption{\textbf{Measures of patching effectiveness.} For all metrics, lower values indicate that patching is more effective. $x_\alpha$ denotes the accuracy of model $\theta_\alpha$ on the \oldtask and $y_\alpha$ on the \newtask.}
\label{tab:metrics}
\end{center}
\end{table}

\subsection{Model similarity}

In addition to the angle between the unpatched and fine-tuned models, we also measure the average absolute difference $||\theta_\textrm{ft}-\theta_\textrm{zs}||_1/n$ between the weights $\theta_\textrm{zs}, \theta_\textrm{ft} \in \mathbb{R}^n$.

Representational similarity is measured through CKA \cite{kornblith2019similarity}, which is shown in Equation \ref{eq:cka}.
In short, CKA can be used to compare the similarity between two models by inspecting their representations on a given set of inputs.
Formally, let $\Theta_\textrm{zs}, \Theta_\textrm{ft} \in \mathbb{R}^{n\times d}$ denote the $d$-dimensional features of the unpatched and fine-tuned model on a dataset with $n$ samples, pre-processed to center the columns.
CKA is then given by:
\begin{equation}
    CKA = \frac{||\Theta_\textrm{ft}^\top\Theta_\textrm{zs}||^2_F}{||\Theta_\textrm{zs}^\top\Theta_\textrm{zs}||_F||\Theta_\textrm{ft}^\top\Theta_\textrm{ft}||_F},
    \label{eq:cka}
\end{equation}
where $||A||_F$ indicates the Frobenious norm of a matrix $A$.
Larger values of CKA indicate more similar representations.
When reporting CKA values, we compute the last layer features using samples from the \oldtask.

\section{Additional baselines and comparisons}
\label{sec:appendix_additional_baselines}

This section details additional baselines for our patching procedure in the context of patching a single task.
We discuss exponential moving averages (Section~\ref{sec:ema}), elastic weight consolidation (Section~\ref{sec:ewc}), learning without forgetting (Section \ref{sec:lwf}, re-training (Section \ref{sec:retrain}), and finally mixing the pre-training and fine-tuning objective (Section~\ref{sec:objmix}).

\subsection{Exponential moving averages (EMA)}
\label{sec:ema}

\begin{figure}
    \centering
    \includegraphics[width=\textwidth]{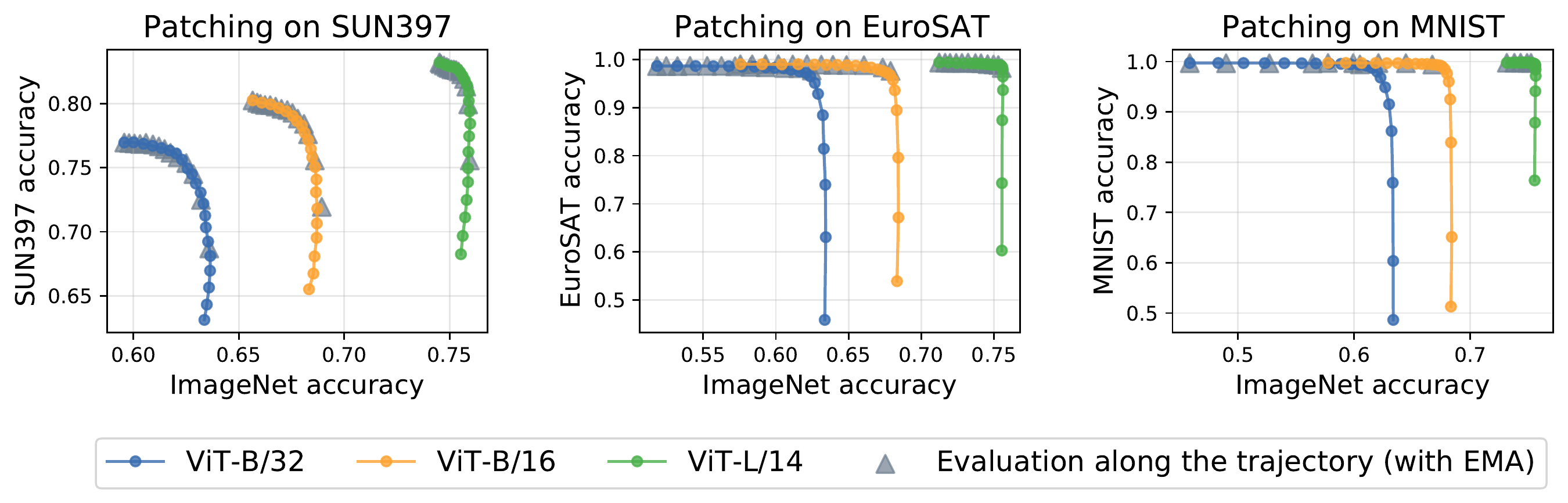}
    \caption{\textbf{Comparisons with EMA.} When fine-tuning with a constant learning rate and using EMA~\cite{szegedy2016rethinking}, interpolating the unpatched and fine-tuned model recovers a similar accuracy trade-off as terminating training early.}
    \label{fig:ema}
\end{figure}

In Figure~\ref{fig:baselines} we show that interpolating the weights of the unpatched and fine-tuned model can recover the ``forgetting frontier''~\cite{ramasesh2021effect}. The forgetting frontier is formed by
fine-tuning with various hyperparameters.
In particular, we show that interpolating the unpatched and fine-tuned models can recover a solution with similar or better accuracies than early termination. 
However, these early termination solutions can potentially be suboptimal because learning rate has not yet decayed to zero.
As such, we recreate this comparison in Figure~\ref{fig:ema} where we fine-tune with a constant learning rate and EMA~\cite{szegedy2016rethinking}. 

\subsection{Elastic weight consolidation (EWC)}
\label{sec:ewc}

\begin{figure}
    \centering
    \includegraphics[width=\textwidth]{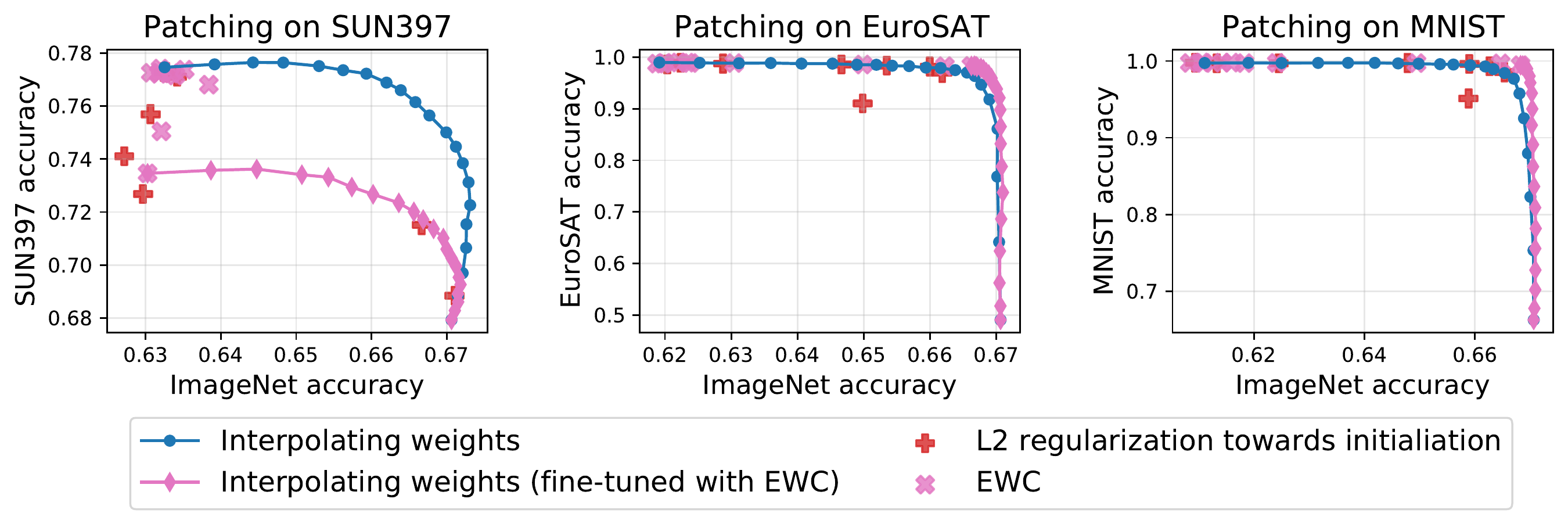}
    \caption{\textbf{Comparisons with EWC and regularization towards initialization.} When data is available from the pre-training set it is possible to augment standard fine-tuning with EWC~\cite{kirkpatrick2017overcoming}.
    EWC provides a solution with a good accuracy trade-off when patching on MNIST and EuroSAT, but not SUN397.
    We also show interpolations between the weights of the unpatched model and a model fine-tuned with EWC.
    As pre-training data is required to compute the fisher information matrix for EWC, these experiments use a ViT-B/16 model from a CLIP reproduction OpenCLIP~\cite{ilharco_gabriel_2021_5143773} pre-trained on LAION 400M~\cite{laion}.
    }
    \label{fig:ewc}
\end{figure}

Elastic weight consolidation~\cite{kirkpatrick2017overcoming} (EWC) is a method which
penalizes the movement of parameters which are deemed important for solving previous tasks.
However, access to the pre-training data is required to investigate which parameters are important.
Therefore, we could not use this method for Figure~\ref{fig:baselines} since the pre-training data for the CLIP models of \citet{radford2021learning} is private.
To examine the performance of EWC, we use a reproduction of CLIP from the OpenCLIP repository~\cite{ilharco_gabriel_2021_5143773}, which is pre-trained on the open source LAION 400M dataset~\cite{laion}.
We use 2,000 iterations of pre-training to compute the fisher information matrix required for EWC.
The results are illustrated in Figure~\ref{fig:ewc} which show EWC solutions corresponding to
different coefficients from the EWC loss ($\{0.001, 0.01, 0.1, 1, 10, 100, 1000, 10000, 100000, 1000000 \}$).
We also interpolate the weights of the unpatched model and the EWC solution fine-tuned with coefficient 1000000.
On MNIST and EuroSAT, an EWC solution exhibits slightly better accuracy trade-offs than any solution on the interpolation between the unpatched model and model fine-tuned without EWC.
However, on SUN397, interpolating the unpatched and fine-tuned model provides a better trade-off than EWC.

\subsection{Learning without forgetting (LwF)}
\label{sec:lwf}

\begin{figure}
    \centering
    \includegraphics[width=\textwidth]{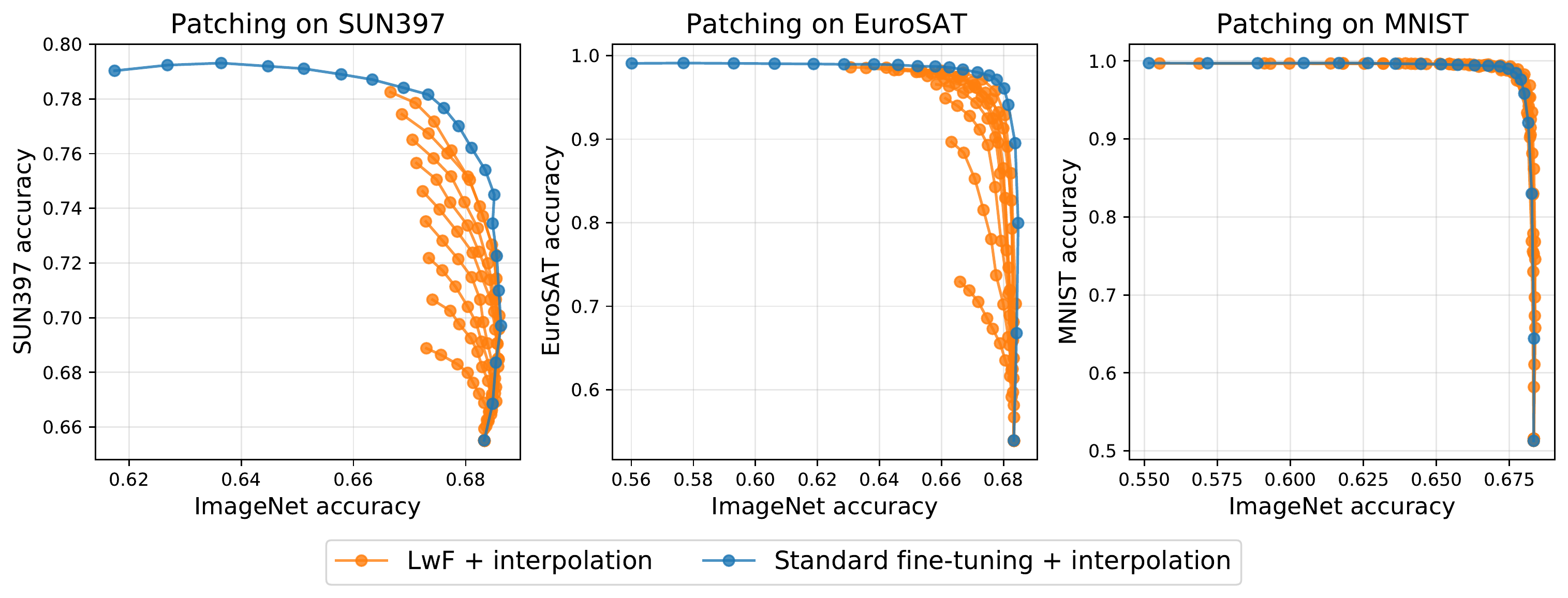}
    \caption{\textbf{Comparisons with LwF.}
    LwF provides a solution with a good accuracy trade-off when patching on MNIST and EuroSAT, but not SUN397.
    We also show interpolations between the weights of the unpatched model and a model fine-tuned with LwF, showing that \ourmethod can be complementary with LwF.
    }
    \label{fig:lwf}
\end{figure}

Learning without forgetting~\cite{li2017learning}(LwF) adds an additional regularization term when fine-tuning based on knowledge distillation \cite{hinton2015distilling}.
We contrast \ourmethod with LwF in Figure \ref{fig:lwf}.
When fine-tuning with LwF, we use multiple loss balance weights $(0.1, 0.2, ..., 0.9)$, and leave the remaining hyper-parameters unchanged. As in Section \ref{sec:ewc}, we use SUN397, MNIST and EuroSAT as our patching tasks, ImageNet as our supported task, and a patch CLIP ViT-B/16 model. As shown in Figure \ref{fig:lwf}, \ourmethod is competitive or better than LwF on all tasks.
Moreover, weight interpolations can further improve on LwF, showing that \ourmethod and LwF are complementary, rather than mutually exclusive alternatives.

\subsection{Re-training with data from the patching task}
\label{sec:retrain}

We further contrast patching with re-training a model from scratch, adding data from the patching task to the pre-training dataset. For such, we use a ViT-B/32 model, training for 32 epochs with cosine learning rate schedule with lienar warmup of 5000 steps and learning rate of 0.001, AdamW optimizer with weight decay of 0.1 and global batch size of 1024, using the open-source library  OpenCLIP~\cite{ilharco_gabriel_2021_5143773}.
We train both on data from YFCC-15M \cite{radford2021learning} alone, and with both YFCC-15M and MNIST data, without upsampling.

\begin{figure}
    \centering
    \includegraphics[width=.6\textwidth]{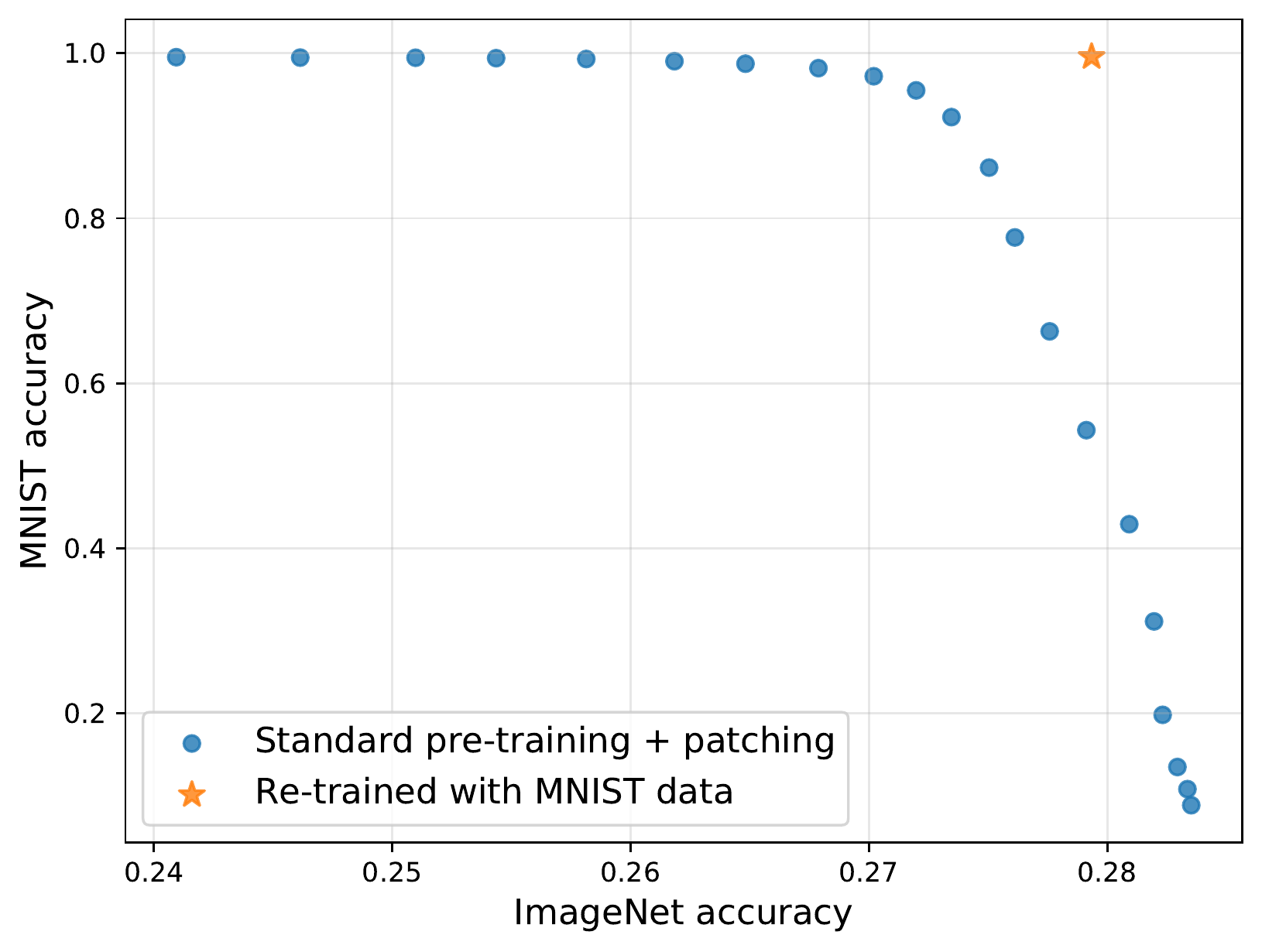}
    \caption{\textbf{Comparisons with re-training.}
    Re-training with data from the patching task is effective at improving accuracy on that task. However, this approach is orders of magnitude more expensive than patching.}
    \label{fig:retrain}
\end{figure}

Results are shown in Figure \ref{fig:retrain}. We find that re-training with data from the patching task is highly effective at improving accuracy on that task, and only slightly decreases zero-shot accuracy on ImageNet. However, we note that pre-training is substantially more expensive than patching. For instance, considering the ViT-L/14 models trained by \citet{radford2021learning}, pre-training takes around 10,000 times more compute than our patching procedure, which makes re-training impractical in most scenarios.

\subsection{Objective mixing}
\label{sec:objmix}

When data is available from the pre-training dataset it becomes possible mix the pre-training and fine-tuning objectives.
This baseline, which we refer to as \emph{objective mixing}, is similar to replay methods in continual learning \cite{rebuffi2017icarl,shin2017continual,lopez2017gradient,chaudhry2018efficient,rolnick2019experience,mirzadeh2020linear}.
However, objective mixing is only possible when the pre-training data is available, which is not the case for the official CLIP models of~\citet{radford2021learning}.
Therefore, we use models from a CLIP reproduction, OpenCLIP~\cite{ilharco_gabriel_2021_5143773}.

\begin{figure}
    \centering
    \includegraphics[width=0.95\textwidth]{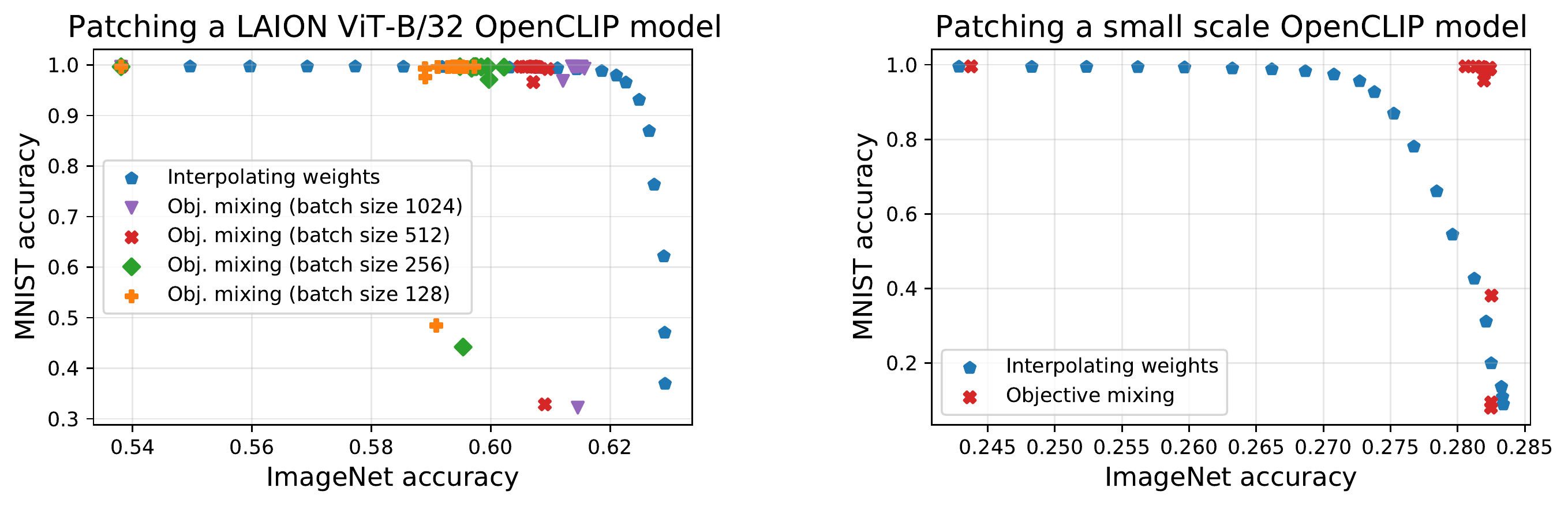}
    \caption{\textbf{Comparisons with objective mixing.} Objective mixing mixes together the pre-training objective and fine-tuning objective.
    Since the pre-training objective requires access to pre-training data, we experiment with a ViT-B/32 model which is pre-trained on LAION 400M~\cite{laion} from the CLIP reproduction OpenCLIP~\cite{ilharco_gabriel_2021_5143773} (left) and a ViT-B/32 model pre-trained by us with a small batch (512) on YFCC-15m~\cite{thomee2016yfcc100m,radford2021learning}.
    The left plot shows that objective mixing is difficult because it requires a large batch size.
    Pre-training on LAION 400M uses batch size 32k, and zero-shot accuracy drops when we continue pre-training with a smaller batch size.
    In this setting, the patching procedure we propose (denoted by ``interpolating weights'') offers the better accuracy trade-off.
    The right plot shows that, while objective mixing can be successfully applied to models pre-trained with small batch sizes, these models perform worse overall.
    Details are in Section~\ref{sec:objmix}.
    }
    \label{fig:objmix}
\end{figure}

First, we use an OpenCLIP ViT-B/32 model which is trained on the publicly available LAION 400M dataset~\cite{laion} and showcase results in Figure~\ref{fig:objmix} (left).
Like CLIP models from~\citet{radford2021learning}, this model is trained with a batch size of approximately 32,000.
We use a single machine and try batch sizes 128, 256, 512, and 1024 for the pre-training objective.
Overall the loss is given by $(1-\beta) \cdot \ell_{\text{pre-training}}  + \beta \cdot \ell_{\text{fine-tuning}}$ for $\beta \in \{0, 0.001, 0.01, 0.1, 0.2, ..., 0.9, 1.0\}$ where
$\ell_{\text{pre-training}}$ and $\ell_{\text{fine-tuning}}$ are the pre-training and fine-tuning loss, respectively.
We notice and interesting phenomenon: even when $\beta = 0$, the zero-shot accuracy on ImageNet still drops
from the pre-trained model (Figure~\ref{fig:objmix}).
We hypothesize that the drop in zero-shot accuracy is likely due to a reduction in batch size and note that the importance of large batches for CLIP objectives has been studied by~\citet{pham2021combined,radford2021learning}.
Because of this drop in accuracy, the patching procedure we study matches or provides better accuracy trade-offs in this setting.

Next, we pre-train our own ViT-B/32 model on a single machine using OpenCLIP~\cite{ilharco_gabriel_2021_5143773}.
using a batch size of 512 for the pre-training objective.
We also use the smaller pre-training dataset YFCC-15M~\cite{thomee2016yfcc100m, radford2021learning}.
In Figure~\ref{fig:objmix} (right) we perform objective mixing with coefficients $\beta \in \{0, 0.001, 0.01, 0.1, 0.2, ..., 0.9, 1.0\}$ and batch size 512 for the pre-training objective.
In this setting, zero-shot accuracy does not drop for $\beta=0$ and objective mixing performs better than the patching procedure we propose.
We believe this is because scale helps our method (see Section~\ref{sec:scale}), and this is the smallest scale model we examine in terms of pre-training set size.
In terms of absolute accuracy, applying \ourmethod to the LAION pre-trained ViT-B/32 model is the best option.

In conclusion, the findings of these experiments are that:
\begin{enumerate}
    \item Objective mixing is difficult because it requires a large batch size.
    \item Objective mixing with a small batch size performs well when applied to CLIP models trained with small batch sizes, but these models are worse overall.
\end{enumerate}

\section{Additional plots for patching on a single task}
\label{sec:appendix_additional_sup_tasks}

\begin{figure*}
    \centering
    \includegraphics[width=\textwidth]{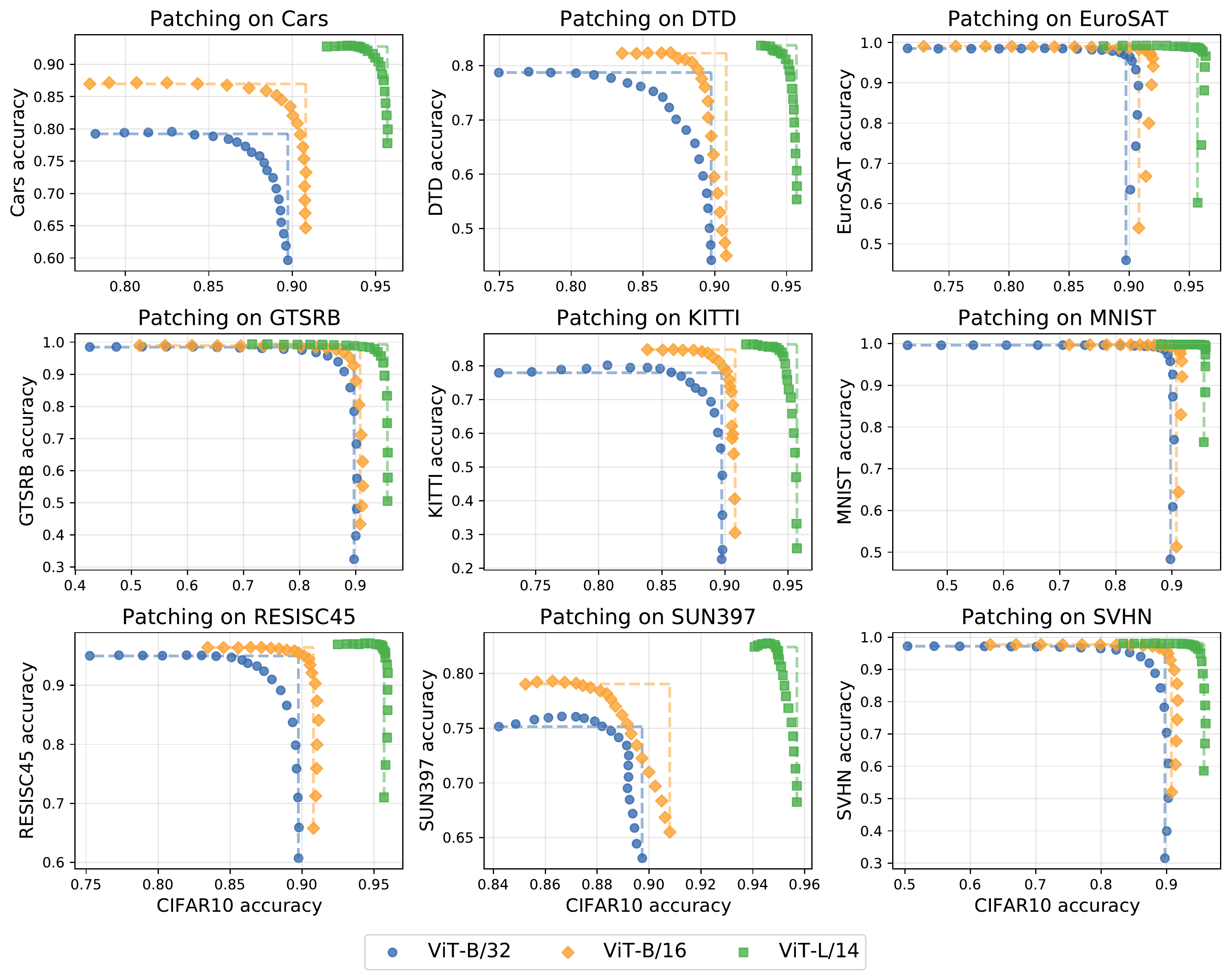}
    \caption{\textbf{Patching results for CIFAR10 as the \oldtask.} Results are shown for nine \newtasks.}
    \label{fig:app_sup_cifar10}
\end{figure*}

\begin{figure*}
    \centering
    \includegraphics[width=\textwidth]{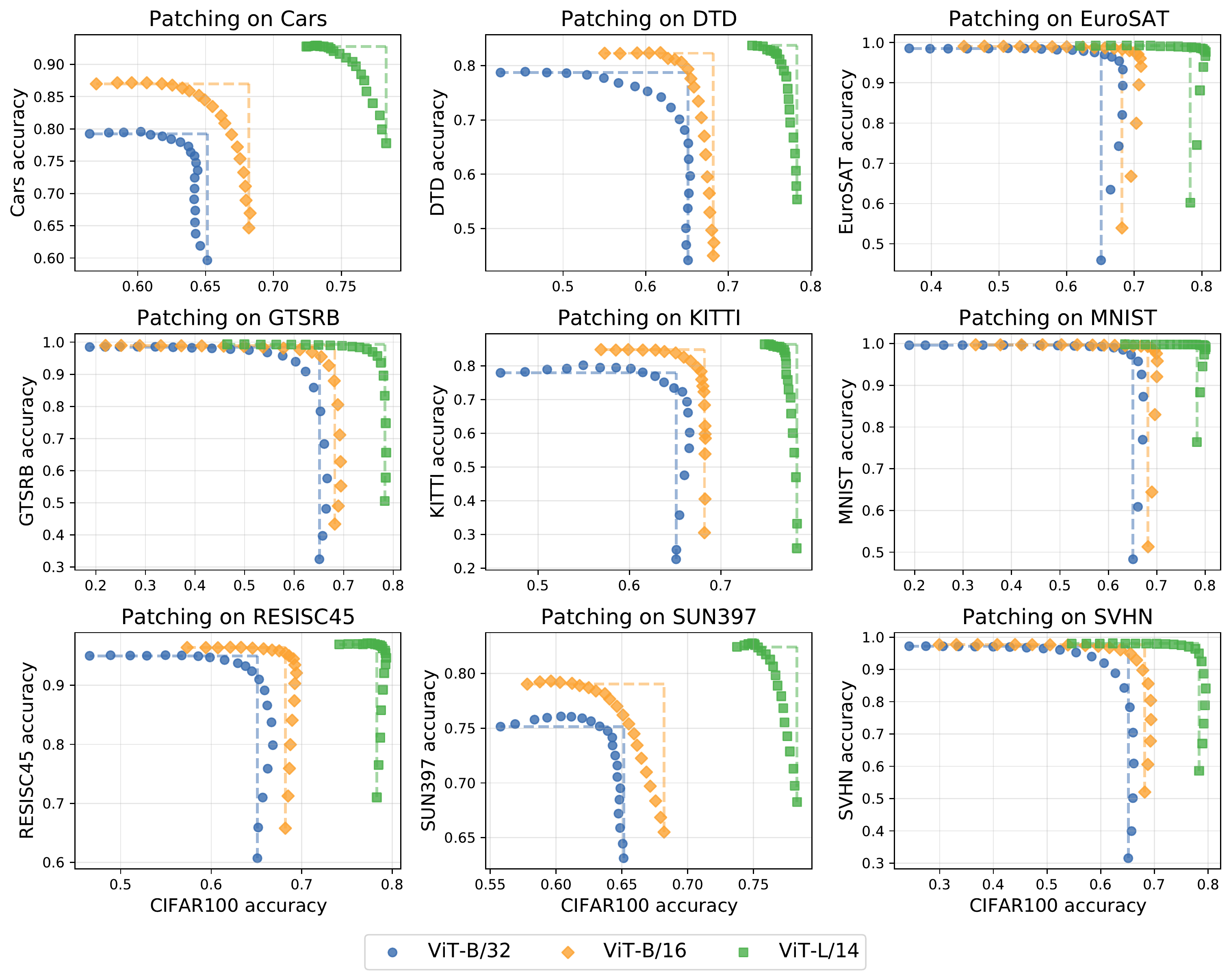}
    \caption{\textbf{Patching results for CIFAR100 as the \oldtask.} Results are shown for nine \newtasks.}
    \label{fig:app_sup_cifar100}
\end{figure*}

\begin{figure*}
    \centering
    \includegraphics[width=\textwidth]{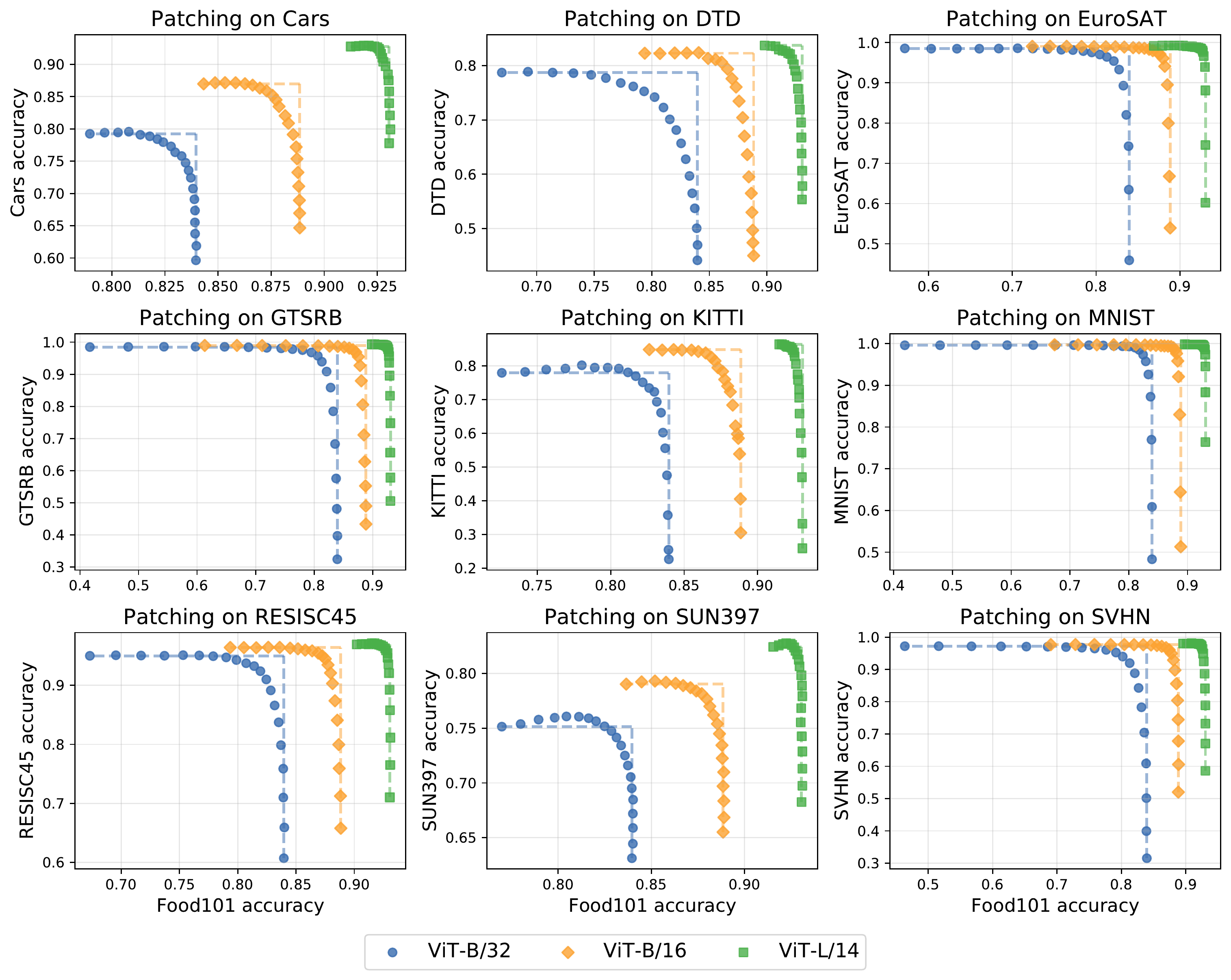}
    \caption{\textbf{Patching results for Food101 as the \oldtask.} Results are shown for nine \newtasks.}
    \label{fig:app_sup_food101}
\end{figure*}

\begin{figure*}
    \centering
    \includegraphics[width=\textwidth]{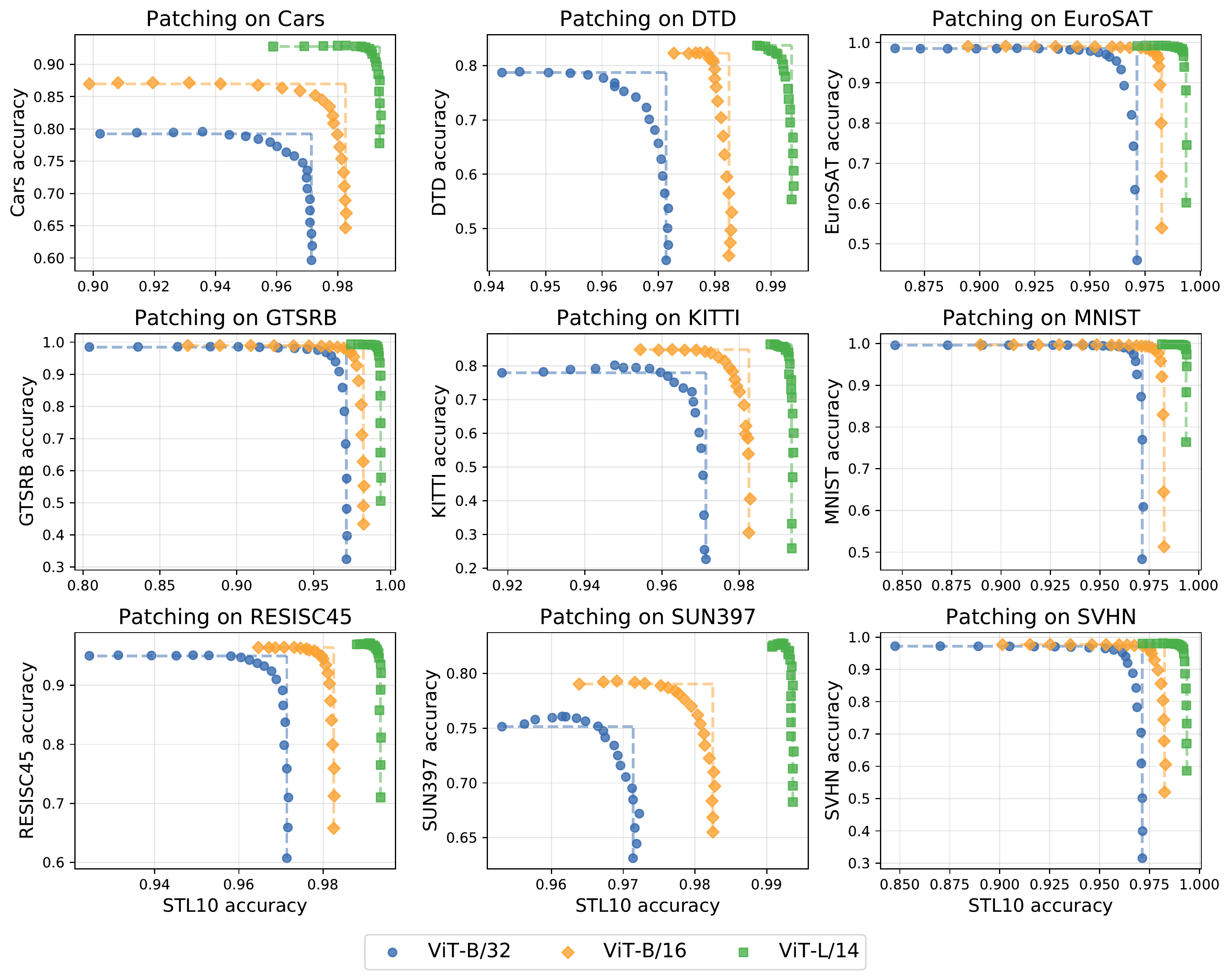}
    \caption{\textbf{Patching results for STL10 as the \oldtask.} Results are shown for nine \newtasks.}
    \label{fig:app_sup_stl10}
\end{figure*}

Breakdowns for each \oldtask and \newtask are shown in Figures \ref{fig:app_sup_cifar10} to \ref{fig:app_sup_stl10}.

\begin{figure*}
    \centering
    \includegraphics[width=\textwidth]{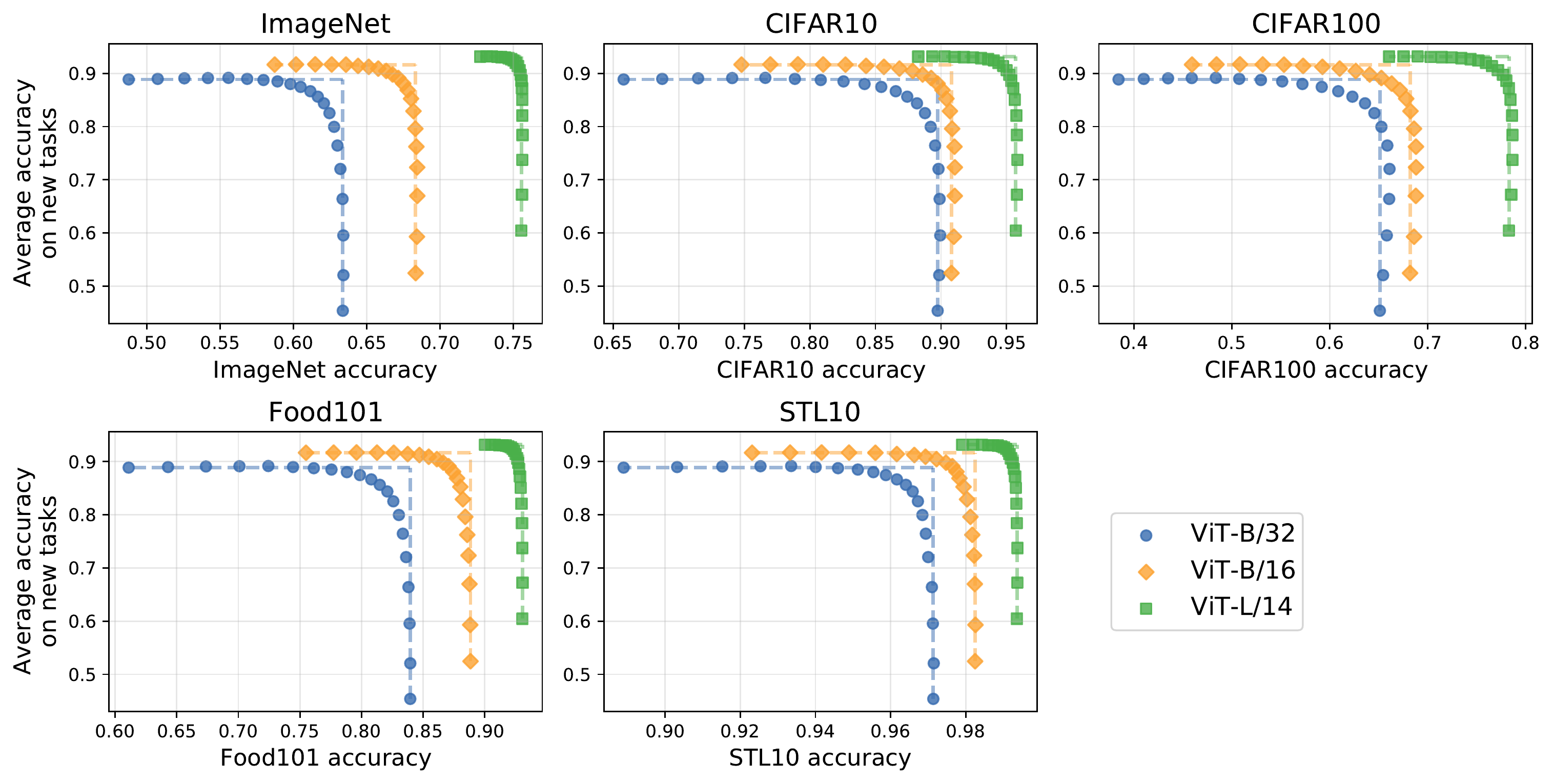}
    \caption{\textbf{Average patching results for various \oldtasks} For multiple \oldtasks, we observe similar accuracy improvements on \newtasks, without substantially decreasing \oldtask accuracy.}
    \label{fig:app_additional_sup_tasks}
\end{figure*}

\begin{figure*}
    \centering
        \vspace*{-0.3cm}
    \includegraphics[width=\textwidth]{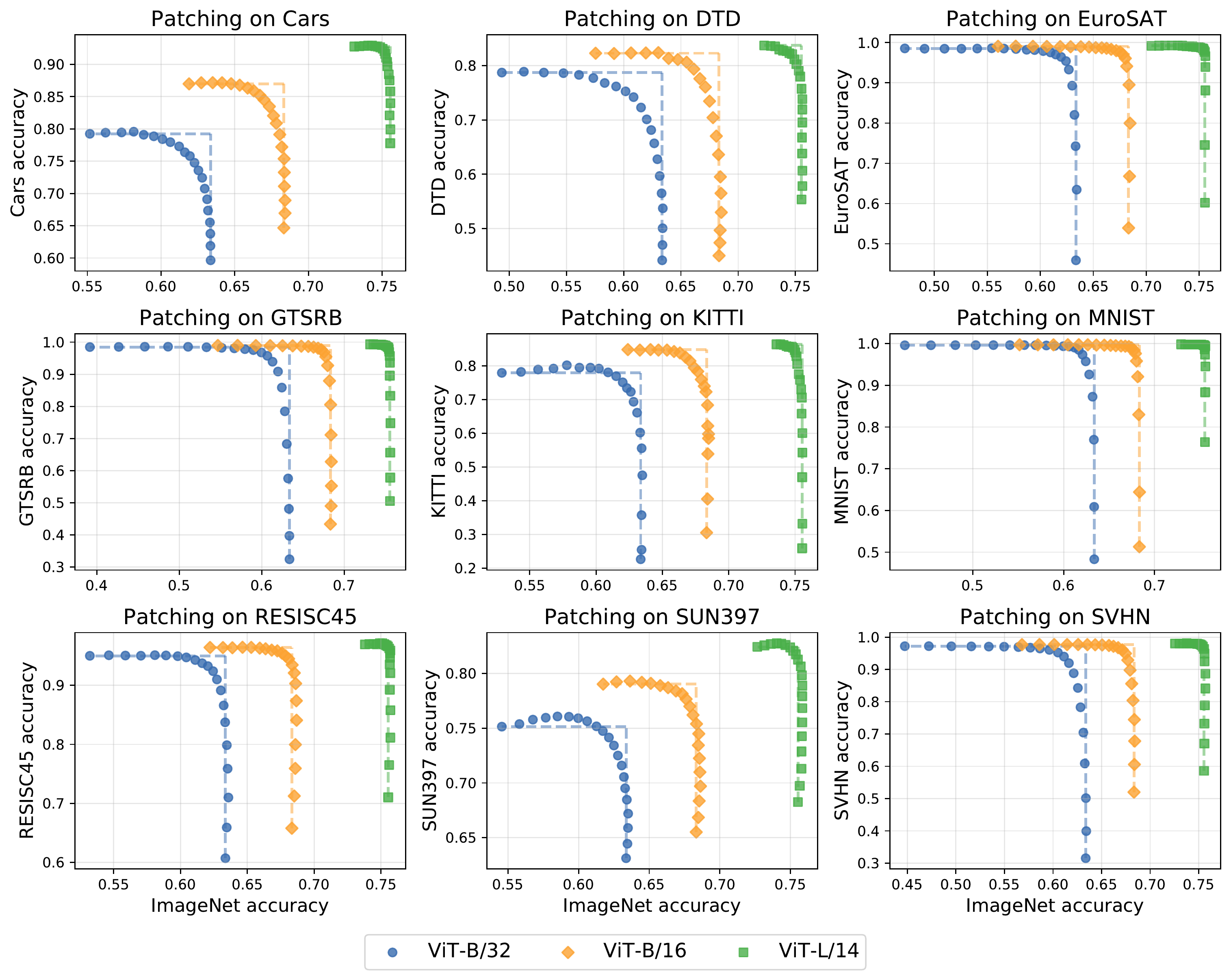}
    \caption{{Patching results for ImageNet as the \oldtask.} Results are shown for nine \newtasks.}
    \label{fig:app_sup_imagenet}
    \vspace*{-0.3cm}
\end{figure*}

\FloatBarrier

\begin{figure}%
    \centering
    \includegraphics[width=\linewidth]{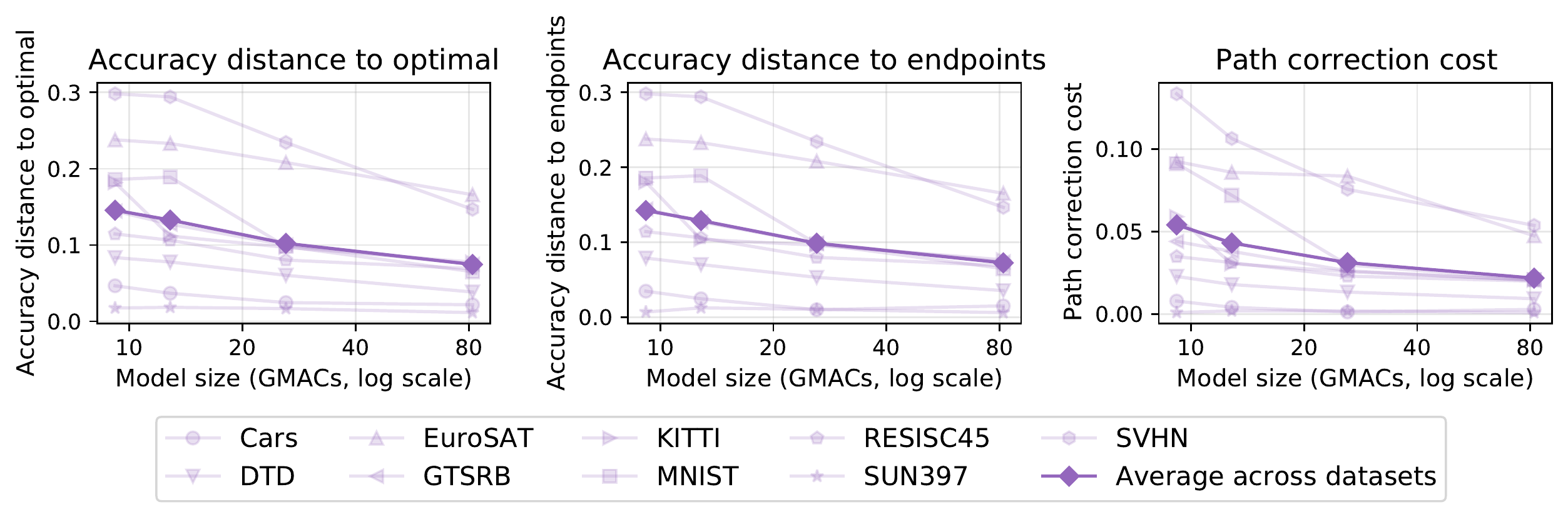}
    \caption{\textbf{The effect of model scale in patching ResNets.} Compared to Vision Transformers (ViTs), patching is less effective for ResNets, corroborating findings of \citet{ramasesh2021effect}.
    Similarly to ViTs, patching is more effective for larger models. }%
    \label{fig:model_ablation}%
\end{figure}

\section{Additional models}
\label{sec:appendix_resnets}

In addition to ViT models, we measure the effectiveness of patching for four ResNet models \cite{he2016deep}.
Specifically, we examine ResNet-50, ResNet-101, and two wider networks, ResNet-50x4, ResNet-50x16 \cite{he2016deep, radford2021learning}. Results are shown in Figure in \ref{fig:model_ablation}.

\section{Patching closed-vocabulary models}
\label{sec:appendix_closed}

Beyond open-vocabulary models, we show that \ourmethod is also effective for closed-vocabulary image classifiers. Our experimental setting is as follows: we start with a (closed-vocabulary) model trained on ImageNet from scratch, from the Pytorch ImageNet Models library \cite{rw2019timm}. Our goal is to expand the set of categories known by the model to improve its performance on MNIST, without hurting accuracy on ImageNet. In other words, we wish to build a model that is competent at classifying an image both amongst the 1000 categories from ImageNet, and amongst 10 digit categories from MNIST. For such, we expand the classification head from the original model by adding 10 new classes, and initialize the corresponding weights and biases to zero. We then fine-tune the model on MNIST without any frozen weights, and interpolate with the model before fine-tuning.

\begin{figure}%
    \centering
    \includegraphics[width=.5\linewidth]{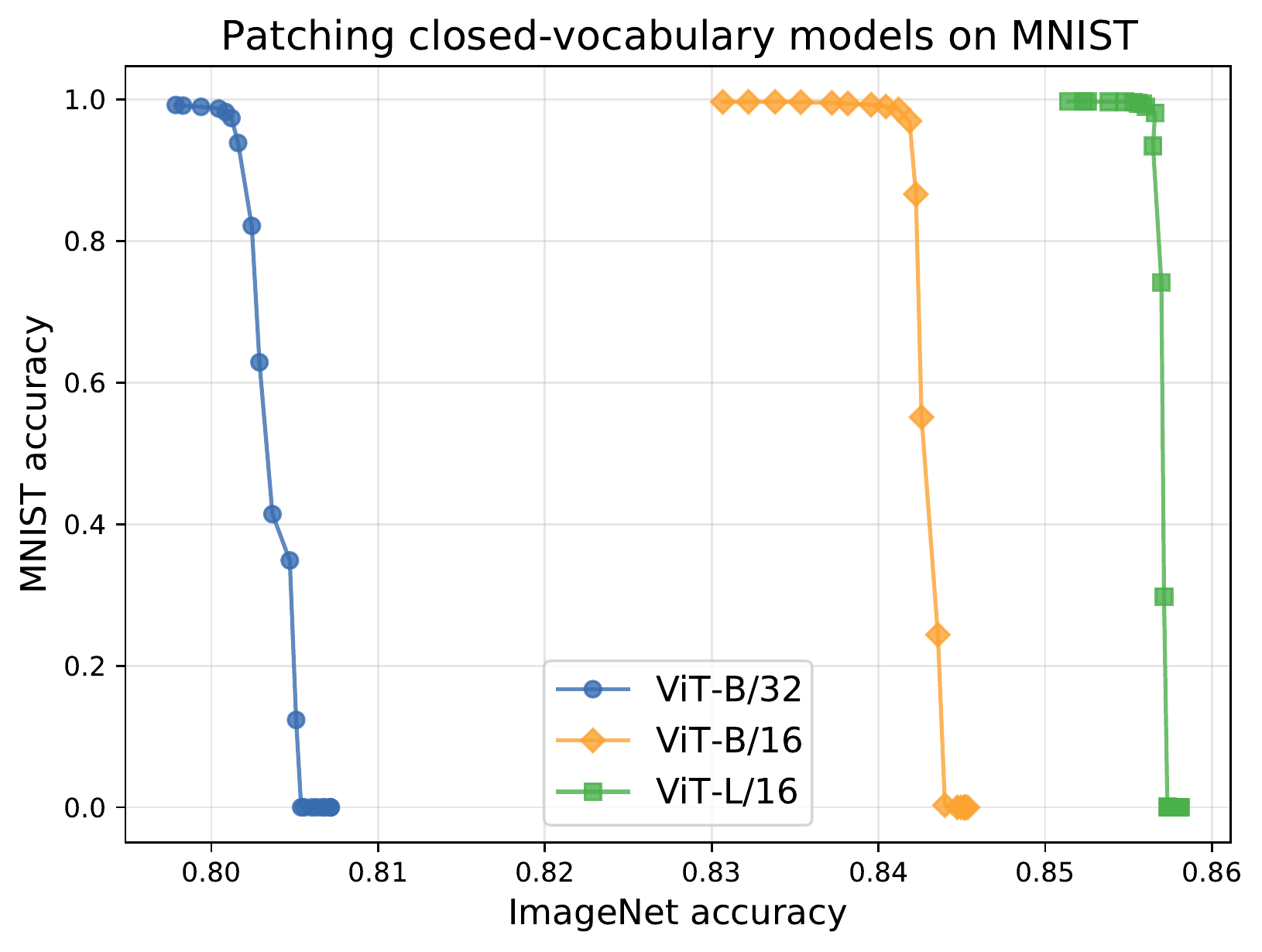}
    \caption{\textbf{\ourmethod is also effective for closed-vocabulary models} For ViT models trained from scratch on ImageNet, \ourmethod improves accuracy on MNIST to over 99\%, while accuracy on ImageNet decreases by less than one percentage point.}%
    \label{fig:closed-vocab}%
\end{figure}

Figure \ref{fig:closed-vocab} shows results for patching closed-vocabulary ViT-B/32, ViT-B/16 and ViT-L/16 models that are trained from scratch on ImageNet. For all models, accuracy on MNIST improves to over 99\%, while accuracy on ImageNet decreases by less than one percentage point. These experiments show that patching is effective beyond open-vocabulary models.

\FloatBarrier

\section{Broad Transfer}
\label{sec:appendix_more_transfer}

This section provides additional results for the broad transfer experiments in Section~\ref{sec:transfer}, as well as new experiments where we fine-tune on ImageNet.

\begin{figure}
    \centering
    \includegraphics[width=\textwidth]{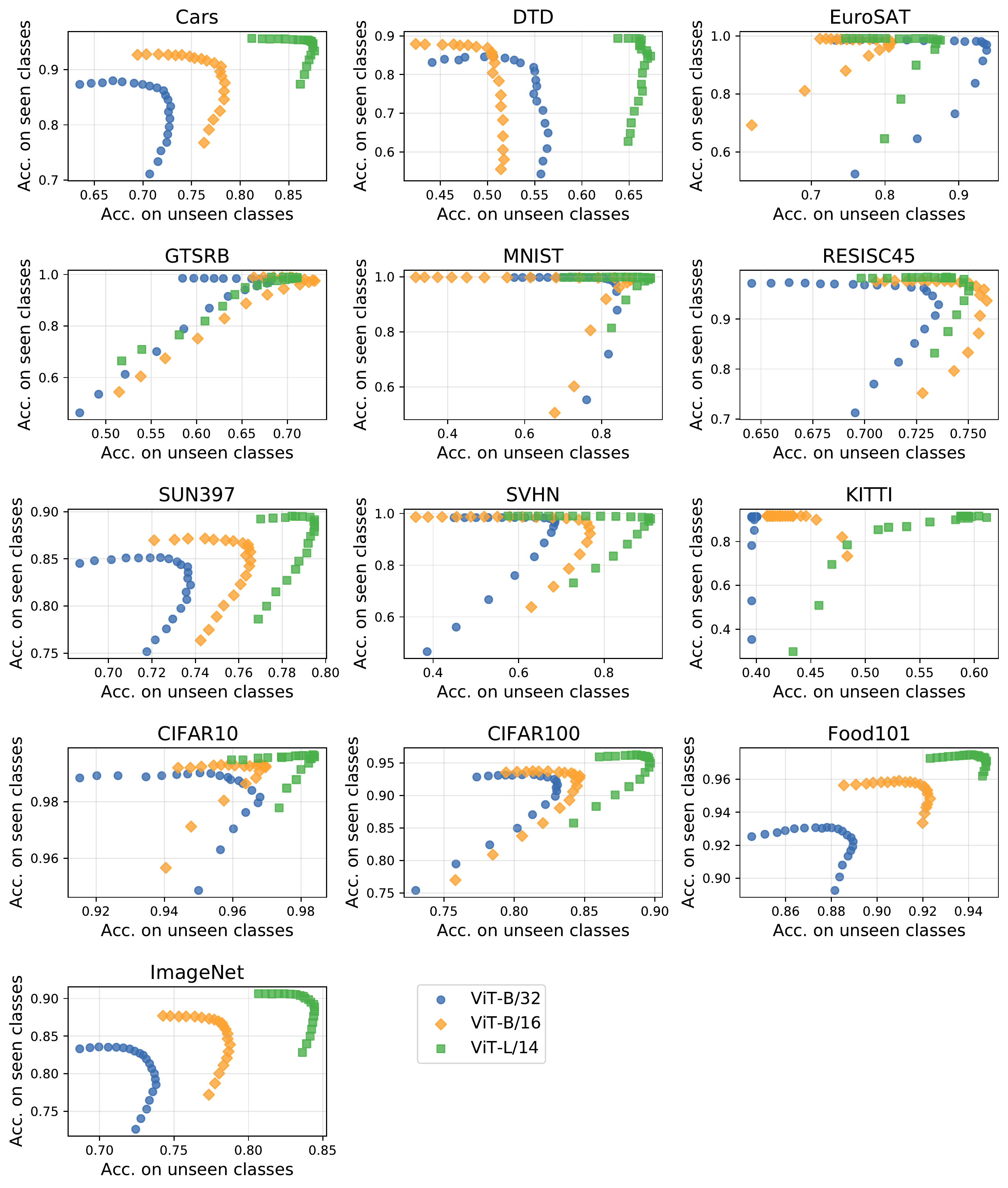}
    \caption{\textbf{The effect of patching on unseen classes.} We randomly partition each dataset into tasks $A$ and $B$ with disjoint class spaces of roughly equal size.
In this Figure we fine-tune on task $A$ then interpolate the fine-tuned model with the unpatched model.
We measure accuracy on task $A$ ($y$-axis) and task $B$ ($x$-axis). This Figure supplements Table~\ref{tab:unseen_gen}.
}
    \label{fig:classgen}
\end{figure}

\begin{figure}
    \centering
    \includegraphics[width=.8\textwidth]{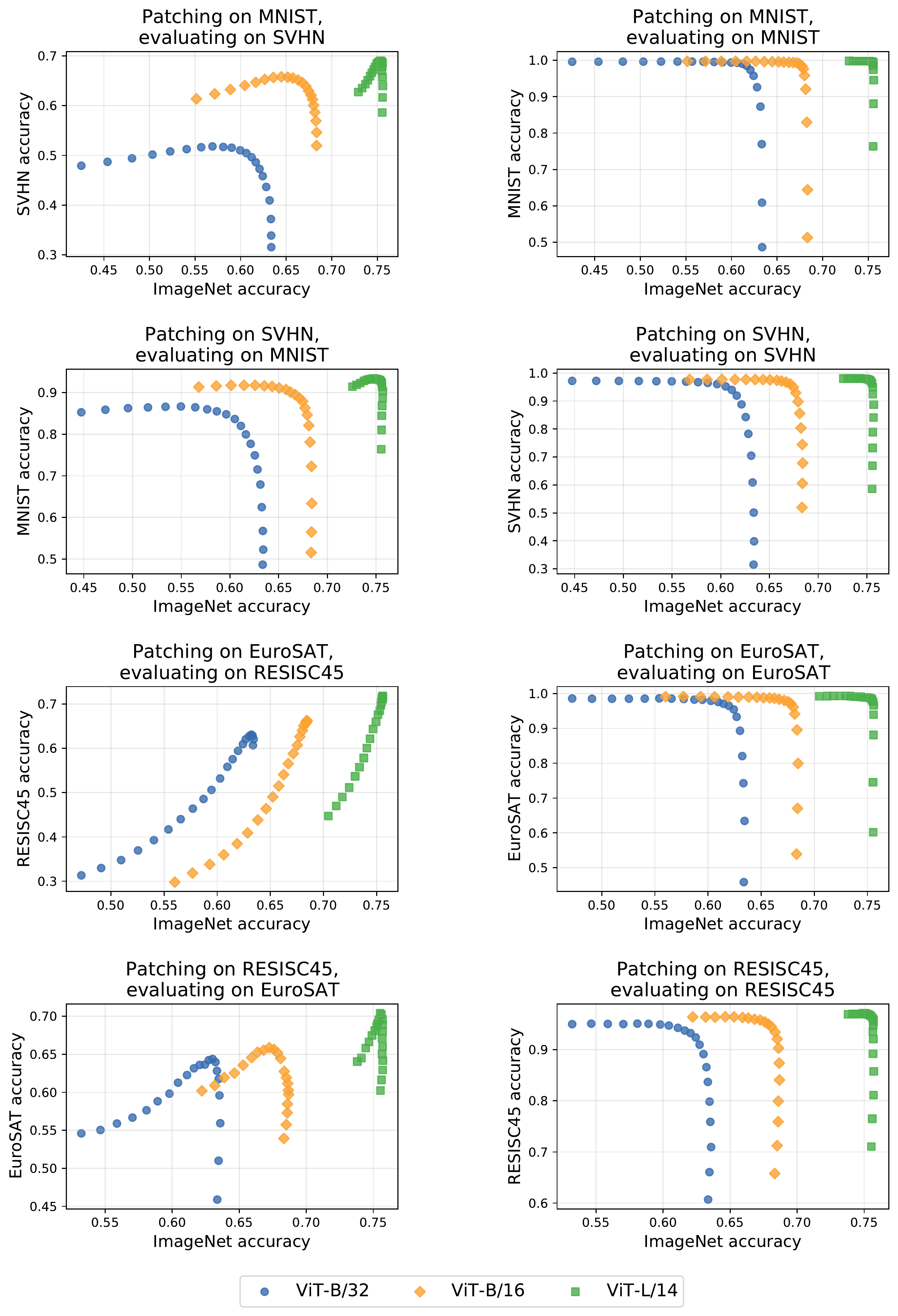}
    \caption{\textbf{The effect of patching on related tasks.}
    In this figure we fine-tune on task $A$ then interpolate the weights of the fine-tuned and unpatched model.
    In addition to measuring accuracy of task $A$ and a supported task ImageNet, we also measure accuracy on a different task $B$.
    This figure along with Figure~\ref{fig:pair2} supplement Table~\ref{tab:transfer_pairs}.
    }
    \label{fig:pair1}
\end{figure}

\begin{figure}
    \centering
    \includegraphics[width=.8\textwidth]{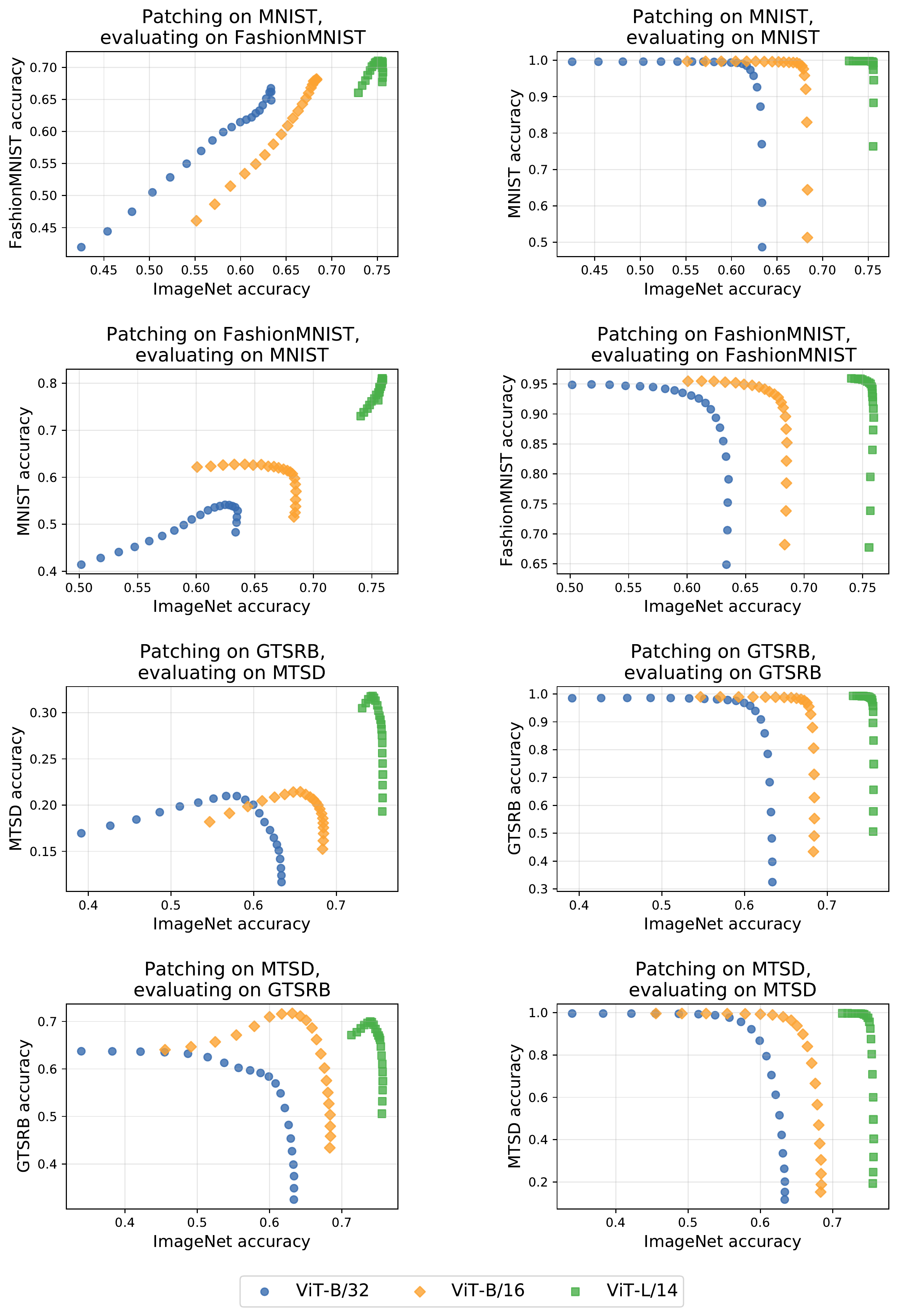}
    \caption{\textbf{The effect of patching on related tasks.}    In this figure we fine-tune on task $A$ then interpolate the weights of the fine-tuned and unpatched model.
    In addition to measuring accuracy of task $A$ and a supported task ImageNet, we also measure accuracy on a different task $B$.
    This figure along with Figure~\ref{fig:pair1} supplements Table~\ref{tab:transfer_pairs}.}
    \label{fig:pair2}
\end{figure}

In Section \ref{sec:transfer}, Tables \ref{tab:unseen_gen} and \ref{tab:transfer_pairs} only show results for the single mixing coefficient that out procedure chooses.
To supplement these tables, we show how accuracy changes for various mixing coefficients $\alpha \in \{0, 0.05, 0.1, ..., 1\}$.
For Table~\ref{tab:unseen_gen} the corresponding figure with all mixing coefficient information is Figure~\ref{fig:classgen}.
Similarly, Table~\ref{tab:transfer_pairs} is expanded in Figures~\ref{fig:pair1} and \ref{fig:pair2}.

\begin{figure}
    \centering
    \includegraphics[width=\textwidth]{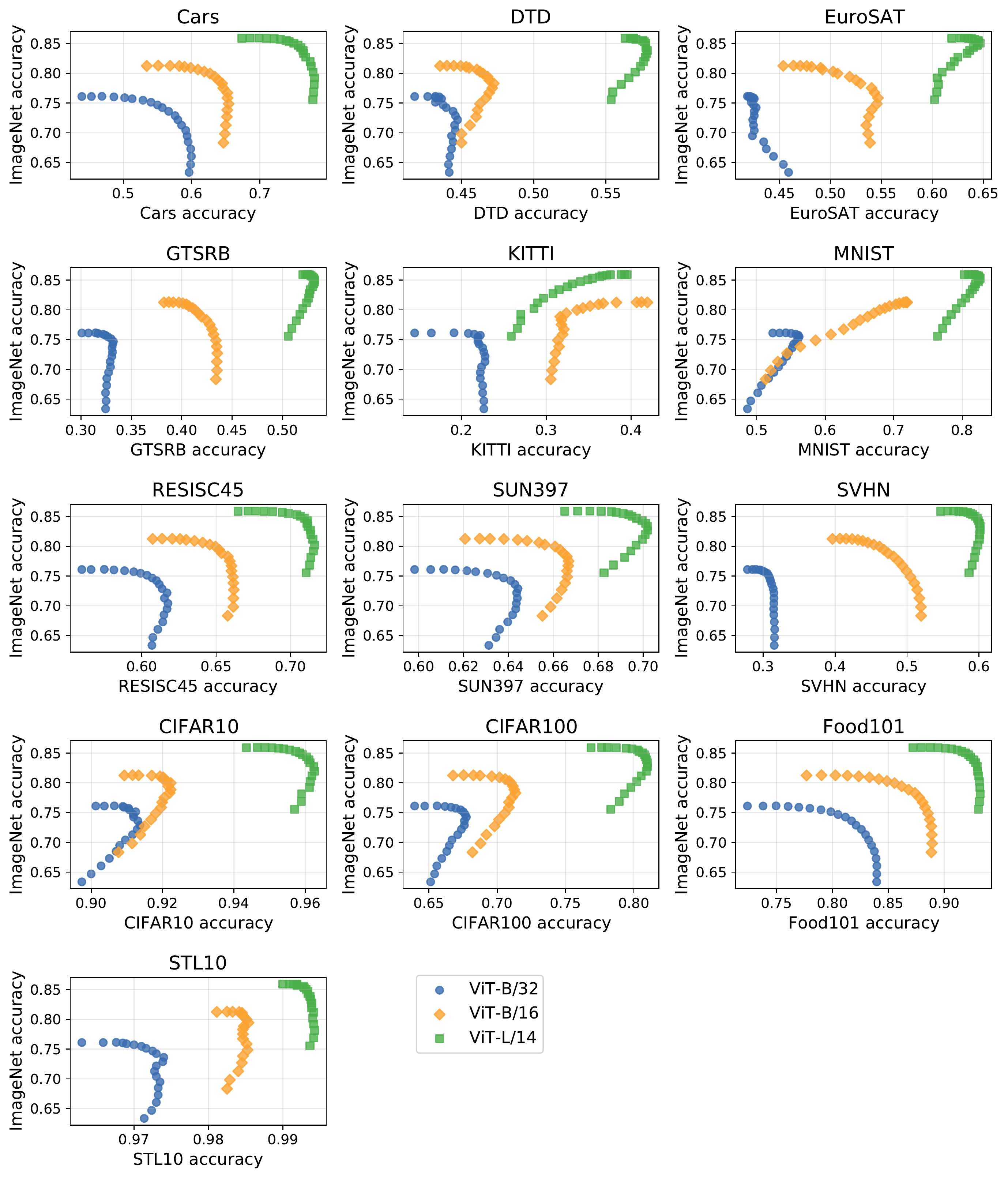}
    \caption{\textbf{Measuring broad transfer to 13 image classification datasets when fine-tuning on ImageNet.}
    We fine-tune on ImageNet then interpolate with the unpatched model with mixing coefficients $\alpha \in \{0, 0.05, ..., 1.0\}$. In addition to measuring ImageNet accuracy, we also measure open-vocabulary accuracy on 13 tasks.
    }
    \label{fig:imtransfer}
\end{figure}

Finally, we measure broad transfer on 13 datasets when patching on ImageNet.
In Figure~\ref{fig:imtransfer} we fine-tune on ImageNet then interpolate with the unpatched model.
Surprisingly, for the ViT-B/16 model, fine-tuning on ImageNet improves accuracy on KITTI accuracy by more than 10 percentage points, and MNIST accuracy by more than 20 percentage points, even without patching.
More investigation into broad transfer is required to understand when and how it applies.

\FloatBarrier

\section{Patching models on multiple tasks}
\label{sec:appendix_multi}

\begin{figure}
    \centering
    \includegraphics[width=\textwidth]{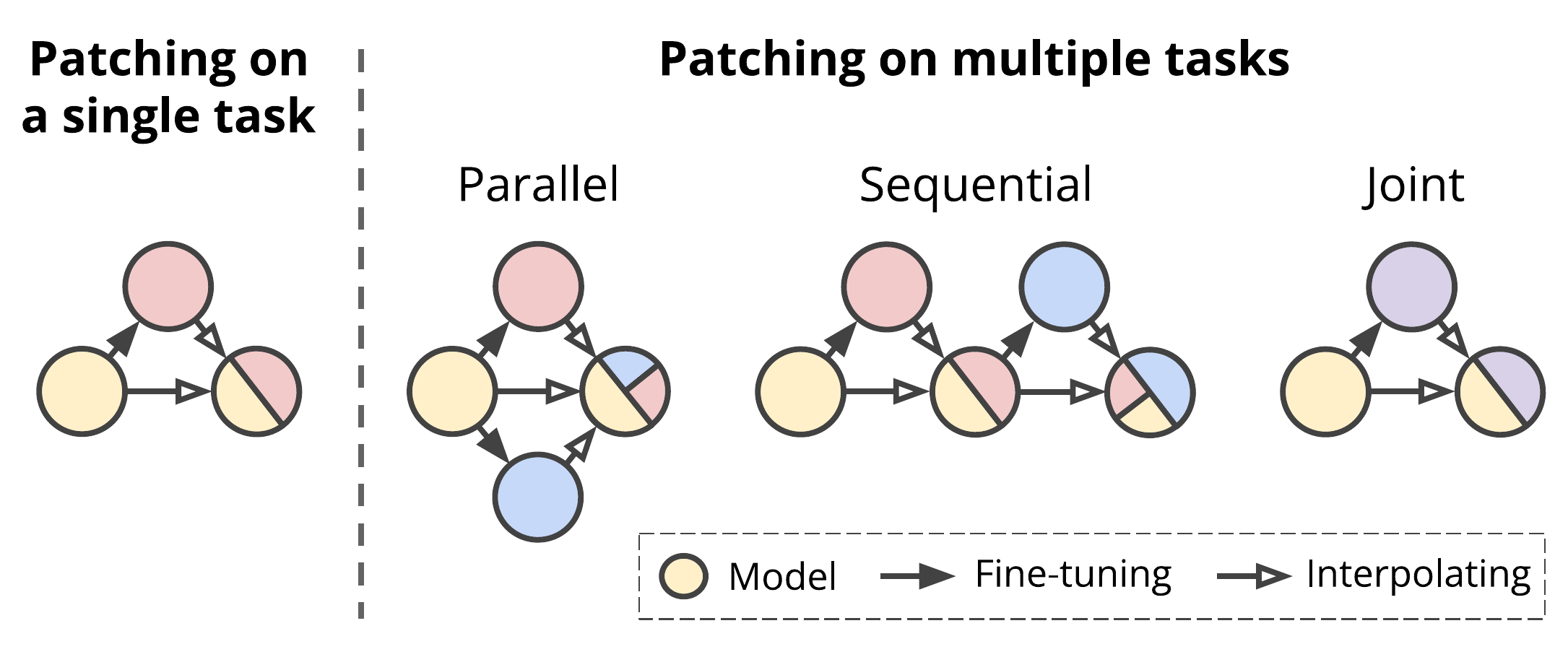}
    \caption{\textbf{An illustration of different patching strategies.} The area proportions on the circles that represent patched models are merely illustrative. In practice, the mixing coefficients are determined based on held-out validation sets. These diagrams are inspired by \citet{matena2021merging}.}
    \label{fig:multi_diagram}
\end{figure}

This section expands the experiments and results on patching models on multiple tasks from Section \ref{sec:multi_patches}.
Figure \ref{fig:multi_diagram} illustrates the different patching strategies.

We first show an exhaustive search over the mixing coefficients for parallel patching on pairs of datasets.
Next, we provide additional results for sequential patching, including additional orders in which the datasets are seen and how the number of datasets affects the quality of patching.
Finally, we present results on SplitCIFAR \cite{rebuffi2017icarl}.

\subsection{Exhaustive parallel search}

\begin{figure}%
    \centering
    \includegraphics[width=\linewidth]{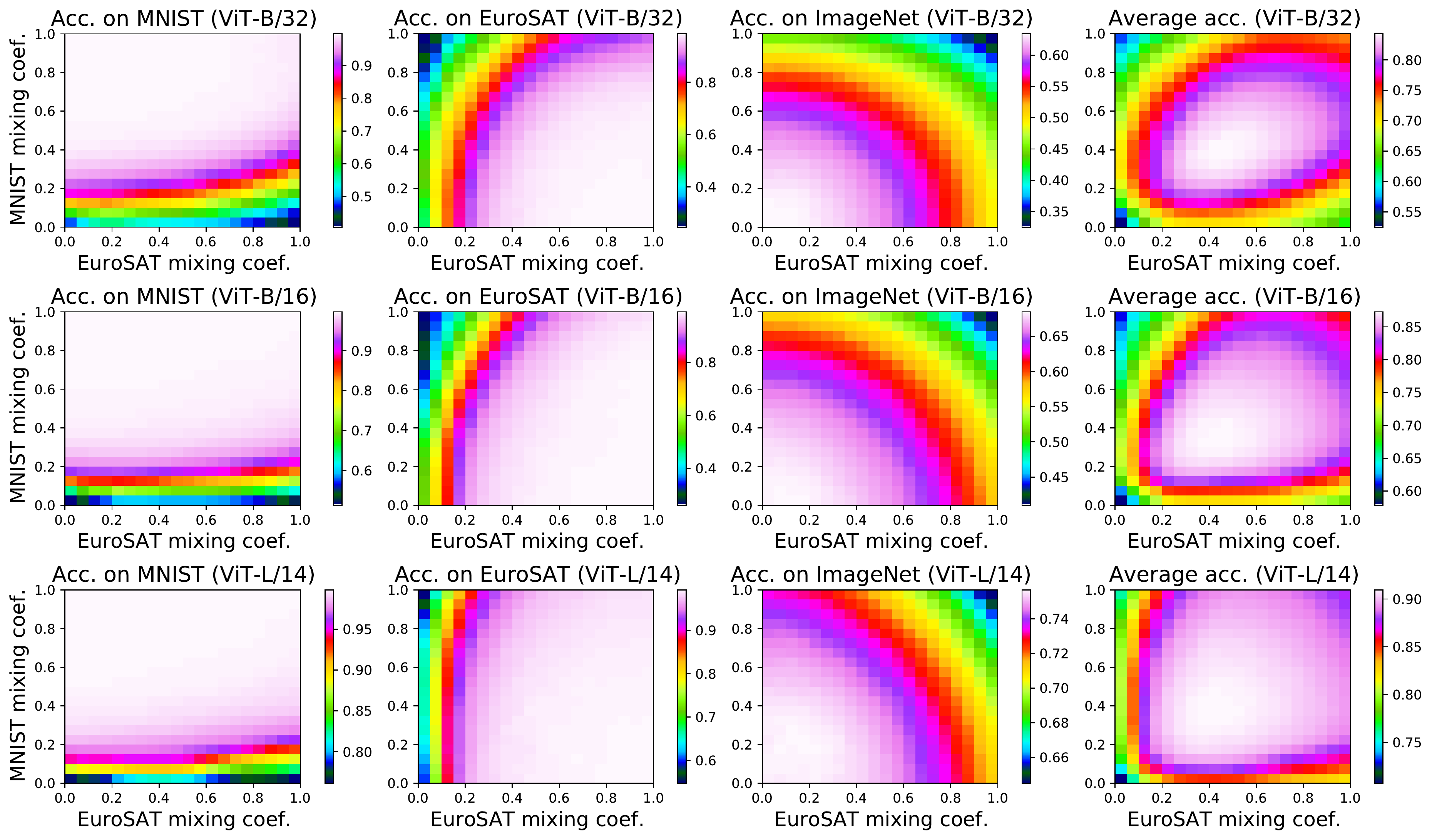}
    \caption{Exhaustive search for parallel patching on MNIST and EuroSAT.}
    \label{fig:grid_mnist_eurosat}
\end{figure}

\begin{figure}%
    \centering
    \includegraphics[width=\linewidth]{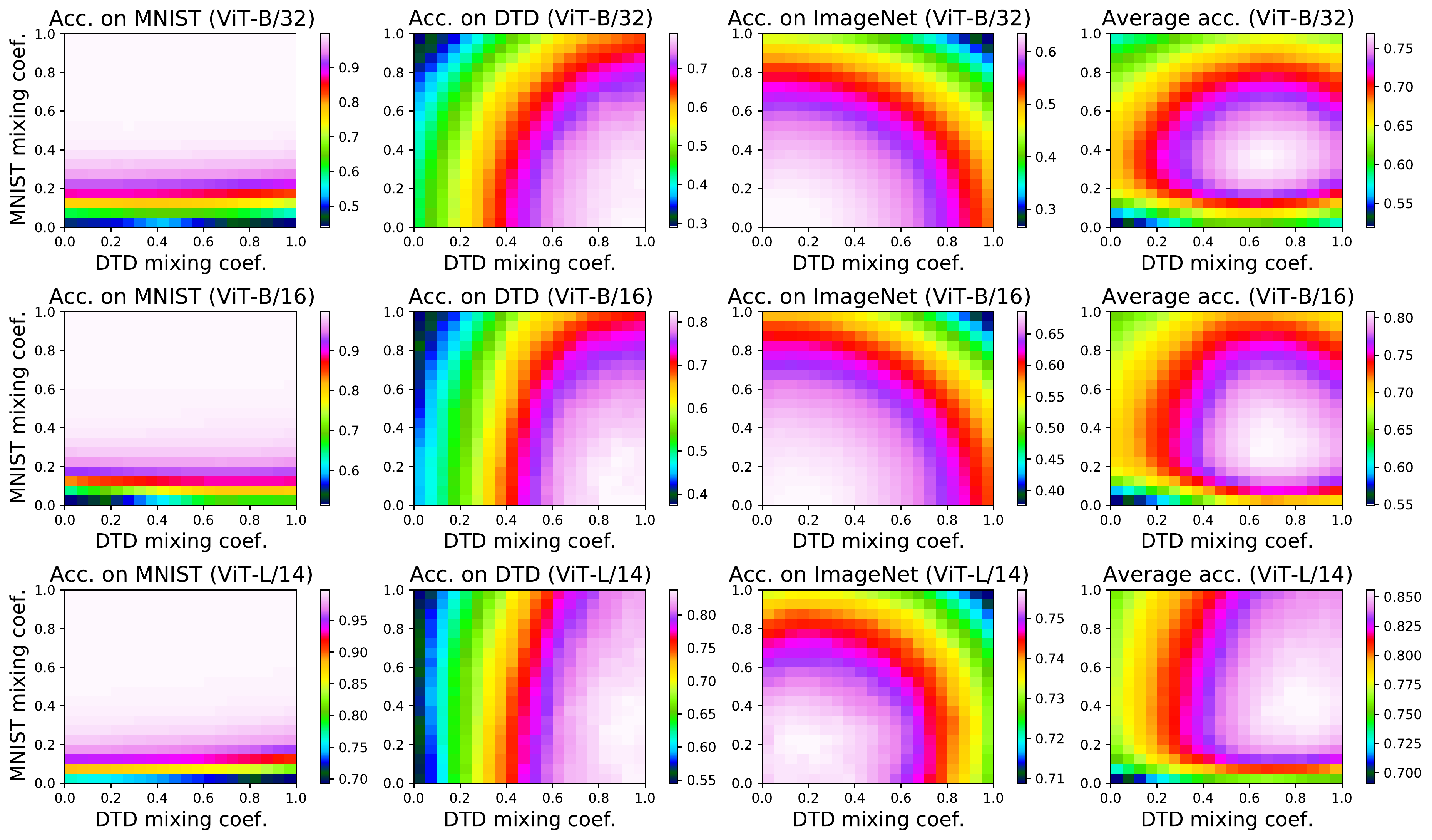}
    \caption{Exhaustive search for parallel patching on MNIST and DTD.}
    \label{fig:grid_mnist_dtd}
\end{figure}

\begin{figure}%
    \centering
    \includegraphics[width=\linewidth]{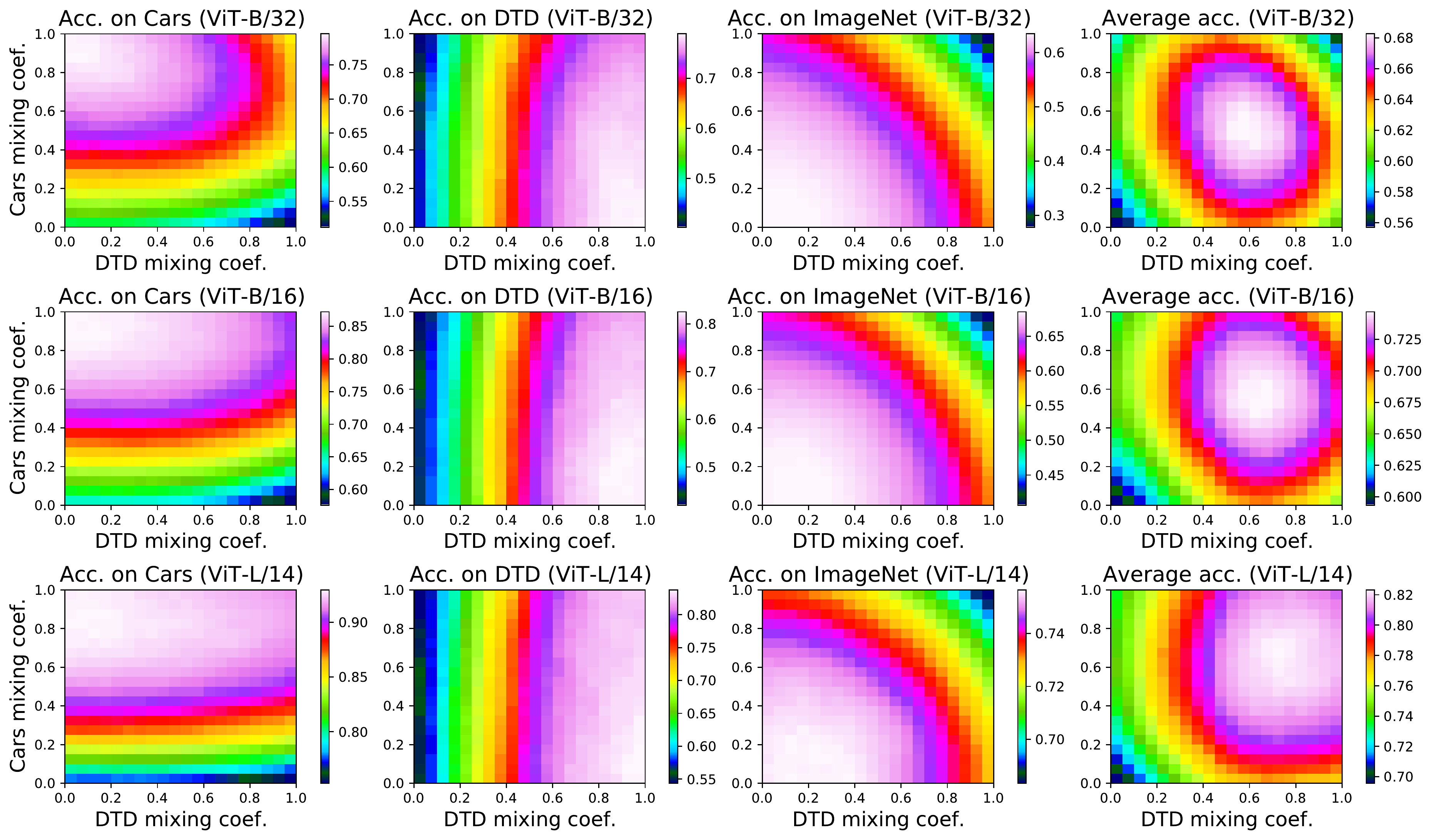}
    \caption{Exhaustive search for parallel patching on Cars and DTD.}
    \label{fig:grid_cars_dtd}
\end{figure}

Recall from Section \ref{sec:multi_patches} that exhaustively searching the mixing coefficients is prohibitively expensive when many \newtasks are used.
Here, we study patching on two \newtasks, for which an exhaustive search is still feasible, and contrast it with other search strategies.
More specifically, we examine three pairs of datasets: i) MNIST and EuroSAT; ii) MNIST and DTD; and iii) Cars and DTD.
The results are shown in Figures \ref{fig:grid_mnist_eurosat} to \ref{fig:grid_cars_dtd}.
For two \newtasks $\mathcal{D}_1$ and $\mathcal{D}_2$, let $\theta_\textrm{ft}^{(1)}$ and $\theta_\textrm{ft}^{(2)}$ be the models fine-tuned on them.
We then measure accuracy of models $\theta = (1-\alpha_1-\alpha_2)\cdot\theta_\textrm{zs} + \alpha_1\cdot\theta_\textrm{ft}^{(1)} + \alpha_2\cdot\theta_\textrm{ft}^{(2)}$ for $\alpha_1, \alpha_2 \in [0, 1]$.
In most cases, there exists some values of the mixing coefficients such that accuracy is high on all the three tasks. For instance, when patching a ViT-L/14 on MNIST and EuroSAT, when $\alpha_1=0.35$ and $\alpha_2=0.45$, accuracy on MNIST and EuroSAT is 39 and 23 percentage points higher compared to the unpatched model, while accuracy on ImageNet decreases by less than 1 percentage point.
Note that the area of high average accuracy typically increases with scale, supporting findings of Section \ref{sec:scale}.

In Table \ref{tab:app_grid_pairs} we contrast the average accuracy obtained via exhaustive search with using other search strategies, \textit{uniform search} and \textit{black-box optimization}.
Recall that these methods optimize for average accuracy on the validation sets.
The uniform search strategy, also described in Section \ref{sec:multi_patches}, consists of searching over a single scalar $\beta \in [0, 1]$, inspecting the models $\theta = (1-\beta)\cdot \theta_\textrm{zs} + \alpha/2 \cdot \theta_\textrm{ft}^{(1)} + \alpha/2 \cdot \theta_\textrm{ft}^{(2)}$.
For black-box optimization, we explore an adaptive black box optimization algorithm, Nevergrad \cite{nevergrad}\footnote{\url{https://facebookresearch.github.io/nevergrad/}} over coefficients $\alpha^{(1)}, \alpha^{(2)} \in [0, 1]$ for the model  $\theta = (1-\alpha_1-\alpha_2)\cdot\theta_\textrm{zs} + \alpha_1\cdot\theta_\textrm{ft}^{(1)} + \alpha_2\cdot\theta_\textrm{ft}^{(2)}$.
We initialize all $\alpha^{(1)}$ and $\alpha^{(2)}$ with $1/2$, and run for 50 optimization steps.
As shown in Table \ref{tab:app_grid_pairs}, uniform search and black-box optimization are comparable in performance, and substantially outperform the unpatched model.
However, both strategies still lag behind exhaustive search, indicating headroom for more sophisticated search strategies. 


\begin{table*}
\setlength\tabcolsep{5.0pt}
\begin{center}
\begin{tabular}{lcccc|cccc|cccc}
 & \multicolumn{4}{c}{MNIST/EuroSAT} & \multicolumn{4}{c}{MNIST/DTD} & \multicolumn{4}{c}{Cars /DTD} \\
 & ZS & U & BB & E & ZS & U & BB & E & ZS & U & BB & E \\
\midrule
ViT-B/32 & 52.5 & 78.1 & 78.1 & 84.3 & 51.9 & 70.7 & 72.0 & 76.8 & 55.7 & 65.6 & 65.7 & 68.3 \\
ViT-B/16 & 57.8 & 81.8 & 81.9 & 87.3 & 54.9 & 75.1 & 76.5 & 80.8 & 59.3 & 72.0 & 71.9 & 74.7 \\
ViT-L/14 & 70.7 & 86.9 & 87.0 & 90.9 & 69.1 & 81.8 & 82.6 & 85.6 & 69.6 & 79.6 & 79.8 & 82.3 \\
\end{tabular}
\caption{\textbf{Contrasting multiple search strategies for parallel patching.} \textit{ZS}: zero-shot; \textit{U}: uniform search; \textit{BB}: black-box optimization; \textit{E}: exhaustive search.}
\label{tab:app_grid_pairs}
\end{center}
\end{table*}

\subsection{Sequential patching}

\begin{figure}
    \centering
    \includegraphics[width=\textwidth]{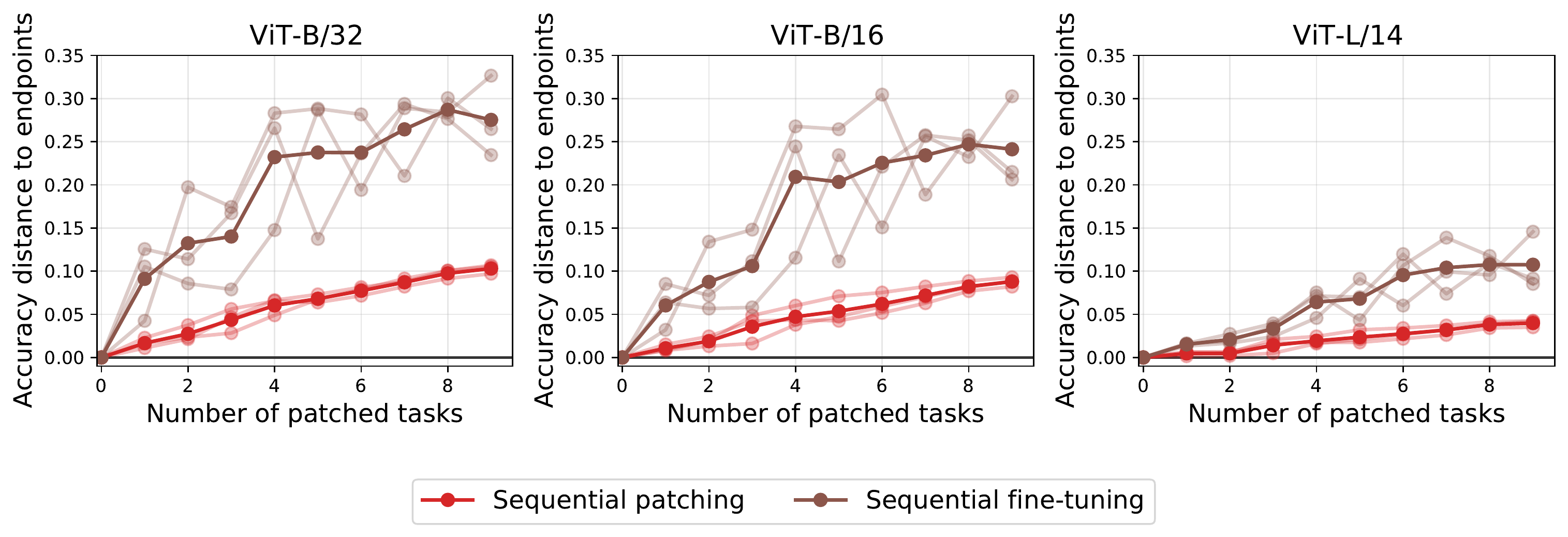}
    \caption{\textbf{The evolution of sequential patching as more tasks are added.} We show how a single model patched with different strategies compares to using one specialized model per task, as more patching tasks are added---the difference in average accuracy is shown in the $y$-axis, and referred to as accuracy distance to endpoints (Section \ref{sec:appendix_scaling_details}). When sequentially patching, the distance in accuracy to using specialized models increases when more tasks are added, although substantially less than when using sequential fine-tuning (without weight interpolation). Results are shown for nine patching tasks, for three different random seeds that control the order in which datasets are seen. The average across random seeds is highlighted.}
    \label{fig:sequential_breakdown}
\end{figure}

In Figure \ref{fig:sequential_breakdown}, we show the evolution of sequential patching as more tasks are added.
The accuracy distance of using a single, patched model to using multiple specialized models increases with with the number of patched tasks, leaving headroom for future work on more sophisticated sequential strategies for patching.
Interestingly, sequential patching outperforms sequential fine-tuning (where no interpolation is used) by a large margin.

\subsection{SplitCIFAR}

\begin{figure}
    \centering
    \includegraphics[width=\textwidth]{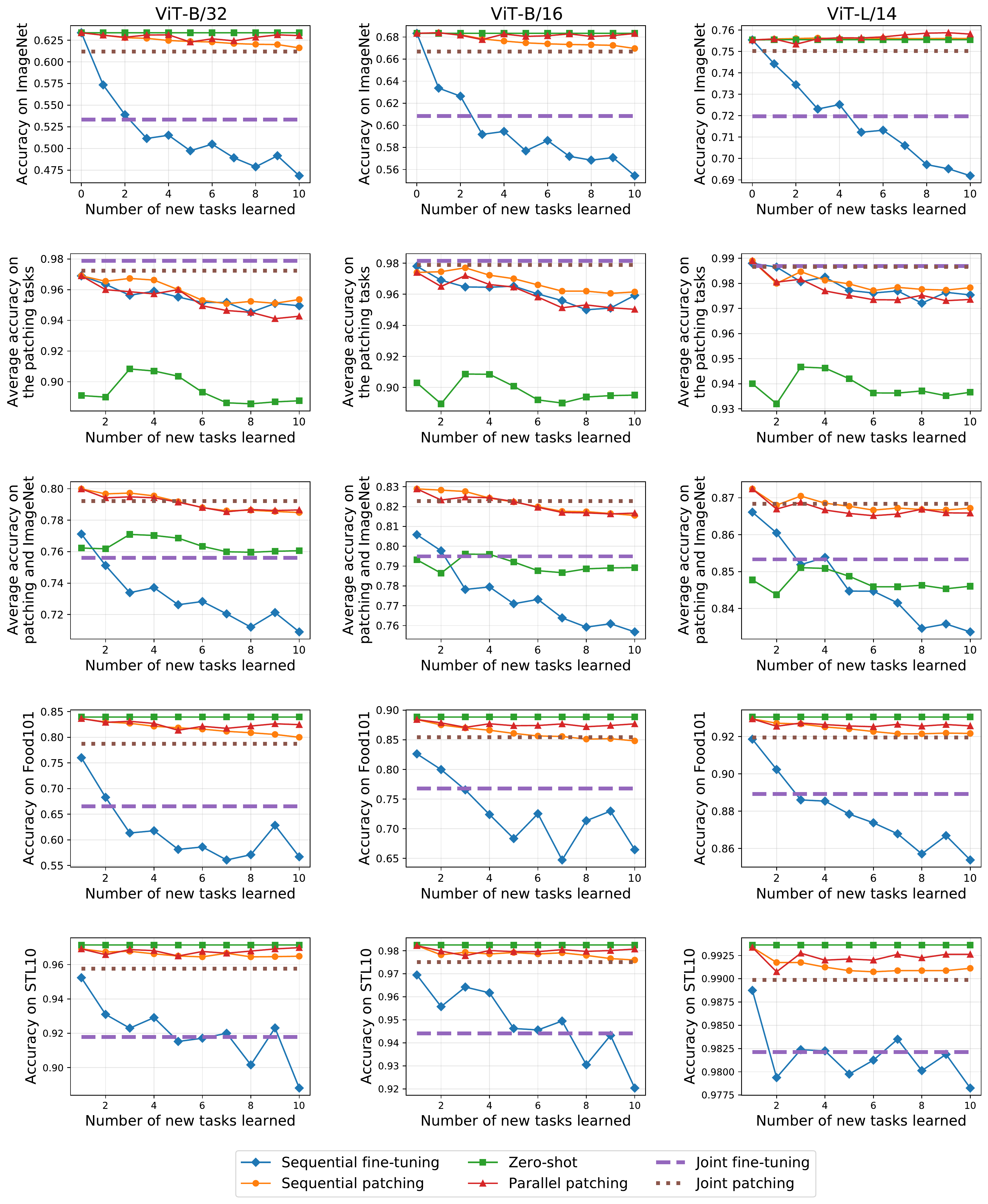}
    \caption{Comparing the methods from Section~\ref{sec:method} when patching on ten tasks from SplitCIFAR~\cite{rebuffi2017icarl}.
    When choosing the mixing coefficient we optimize for average accuracy on the held-out validation sets of the supported task ImageNet and the new patching tasks.
    We also show performance on additional supported tasks Food101 and STL10 to ensure \ourmethod does not overfit to the
    supported task used for choosing $\alpha$.
    \ourmethod show less catastrophic forgetting than the alternative approaches on ImageNet, Food101, and STL10.
    }
    \label{fig:cifar}
\end{figure}

Figure~\ref{fig:cifar} compares the patching methods described in Section \ref{sec:method} when patching on ten tasks from SplitCIFAR100~\cite{rebuffi2017icarl}.
We split CIFAR100 randomly into ten different 10-way classification problems which are either learned jointly (as in joint fine-tuning or joint patching) or independently.
In Figure~\ref{fig:cifar} we show how the number of tasks learned affects accuracy on i) ImageNet, the supported task used for this experiment (first row), 
ii) the patching tasks (second row) and 
iii) average accuracy on the patching tasks averaged with ImageNet accuracy (third row).
We also display accuracy for two additional tasks, Food101 and STL10, in rows four and five, respectively.

While this experiment explores \ourmethod in a more conventional continual learning setting, there is a key difference:
we also examine model accuracy on other tasks like ImageNet and Food101.
To choose the mixing coefficients we optimize for performance on the held-out validation set of all patching tasks seen so far (i.e, the SplitCIFAR100 tasks) and the supported task ImageNet with equal weight.
This is the reason for also examining accuracy on Food101 and STL10---we want to make sure that we are not overfitting to the
representative supported task, ImageNet.

As intended, \ourmethod show less catastrophic forgetting than the alternative approaches on ImageNet, Food101, and STL10.
All methods are fine-tuned with a total of 2,000 iterations. As such, we only use 200 iterations per-task when fine-tuning independently. 

\FloatBarrier

\section{Typographic attacks}
\label{sec:appendix_typo}
In this section we present more details on our typographic attacks experiments presented in Section~\ref{sec:case_studies}.
We first discuss our procedure for creating synthetic typographic attacks.
We then outline our real world data collection scheme.
Finally we present additional experimental details.

\paragraph{Synthetic dataset details.}
\begin{figure}
    \centering
    \includegraphics[width=\linewidth]{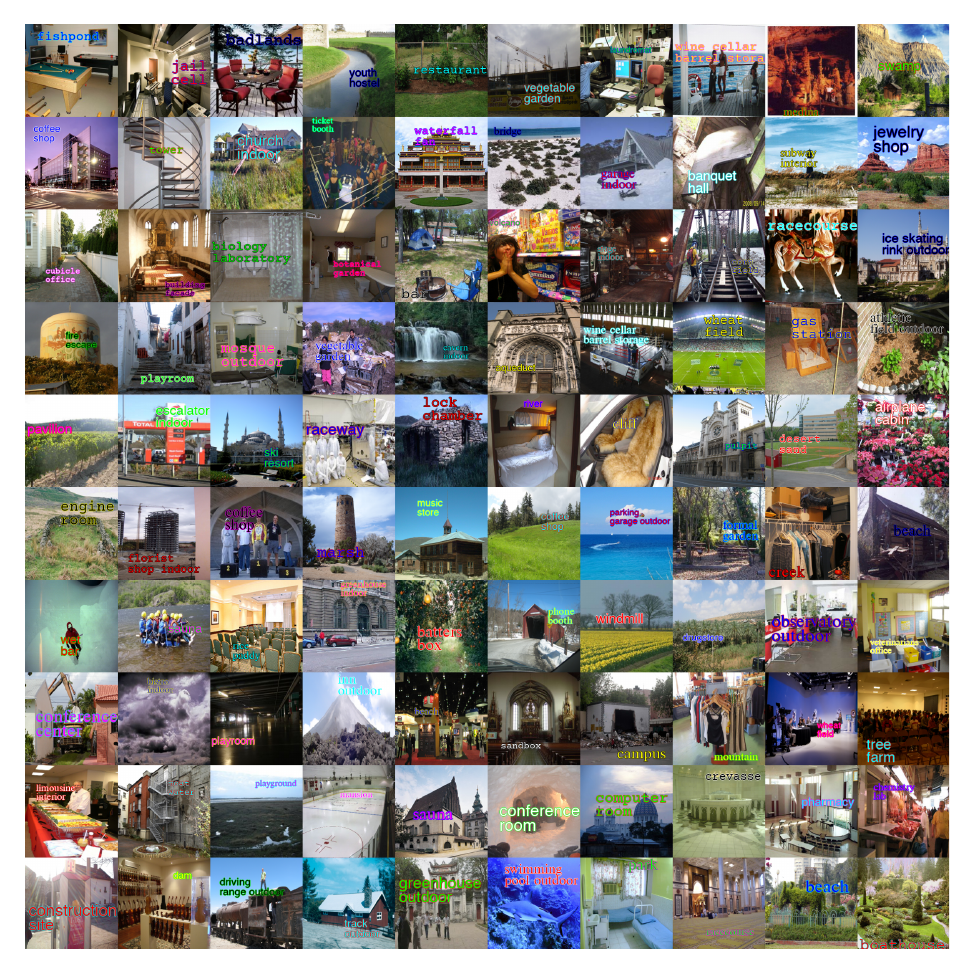}
    \caption{\textbf{Synthetically generated typographic attacks used for patching.}
    Starting with the SUN397 dataset, we procedurally modify images by adding text specifying an incorrect class label for the image.
    By patching on this data, we observe transfer to real-world classification of images visualized in Figure~\ref{fig:real_data}---even with no class-space overlap.
    }
    \label{fig:sun_attack_data}
\end{figure}

Starting with the SUN397~\cite{sun397} dataset, we procedurally add text to images as seen in Figure~\ref{fig:sun_attack_data}.
We resize the shorter dimension to 224 pixels using bicubic interpolation and take a 224 pixel by 224 pixel center crop, which is the standard CLIP resize and crop augmentation~\cite{radford2021learning}.
We randomize over three fonts: Courier, Helvetica and Times.
For font size, we randomly sample over the range 20 to 40 points.
We consider eight colors, and sample uniformly from them: red, green, blue, cyan, magenta, yellow, white, and black.
To make sure font is visible, we outline text with a 1 point shadow that is a different color than the main font color.
The text is randomly placed in the image with checks to ensure that the text fits completely on the image.
The text specifies an incorrect class label, chosen at random.
For instance, if the ground truth label from an image is ``raceway'', the text content will be uniformly sampled from the $397-1=396$ other options.
We apply this typographic attack procedure on the originally SUN397 train and test sets to create attacked train and test sets.

\paragraph{Real dataset details.}
\begin{figure}
    \centering
    \includegraphics[width=\linewidth]{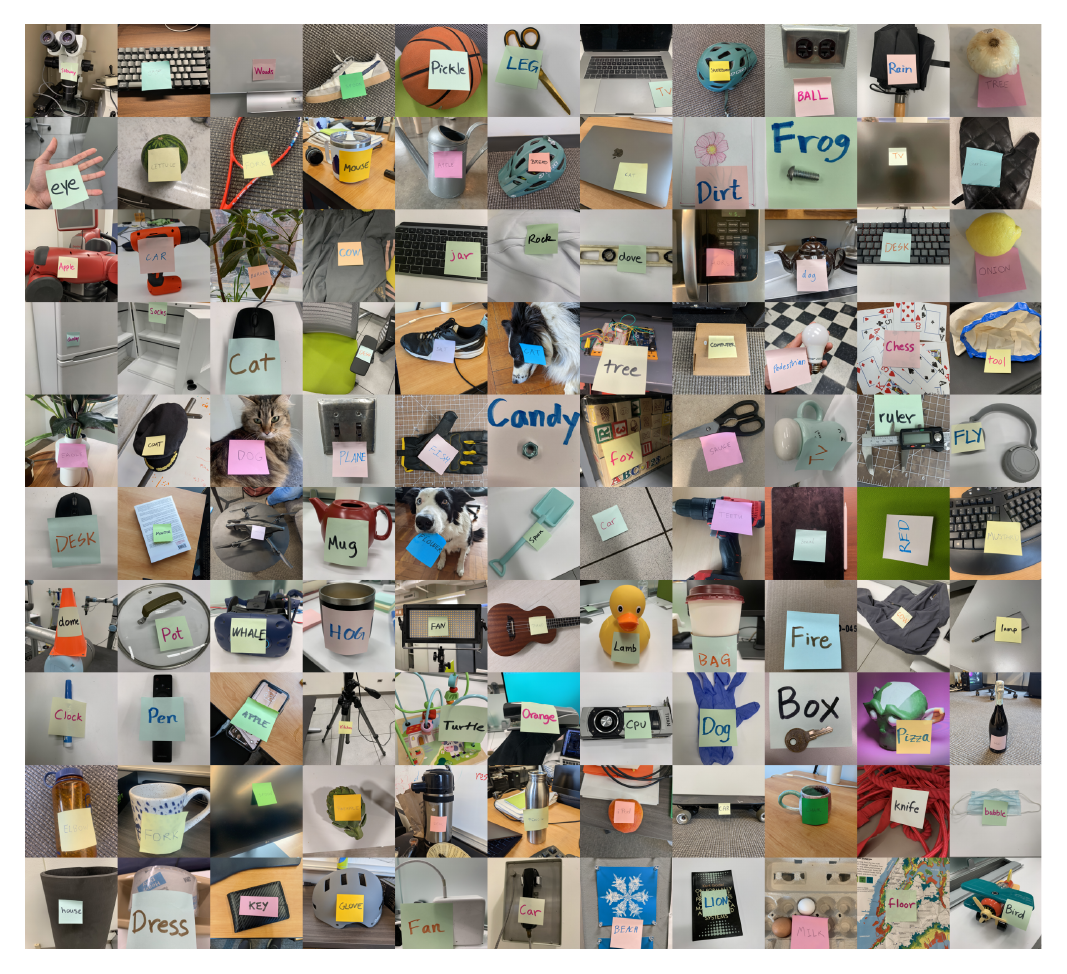}
    \caption{\textbf{Real typographic attacks used for testing broad transfer.}
    Here we present all 110 images from our real-world typographic attacks dataset.
    Annotators were asked to choose both the target objects and the textual corruptions.
    Hence, during data collection there were no restrictions on the true object semantic category or the typographic attack semantic category.
    Note that text is visually dissimilar for the computer generated fonts in Figure~\ref{fig:sun_attack_data}.
    Additionally, images in this dataset are \emph{object-centric}, whereas images in Figure~\ref{fig:sun_attack_data} capture a distribution of scenes.
    }
    \label{fig:real_data}
\end{figure}

To test if patching on synthetic typographic attacks transfers to real-world use-cases, we generate a real-world test set of typographic attacks.
We provide the following instructions to in-house annotators: 1) Write the name of an object on a sticky note, then place the sticky note on an object that is not the one that you have written. For instance, you can write ``desk'' on the sticky note then put it on a mug. The objects should be reasonable categories (e.g., common objects, animals, etc.).
2) Take a picture---make sure that the underlying object is centered and the text is visible.
We acquire consent from each annotator to use their images for research purposes and to release data publicly.
No faces or other identifying human characteristics are present in the dataset.

We show images from our dataset in Figure~\ref{fig:real_data}, highlighting the category diversity.
As highlights: calipers are attacked with text saying ``ruler,'' an artichoke is attacked with text saying ``pineapple,'' and a GPU is attacked with text saying "CPU."
Critically, none of the object categories---either in the image or in the text attack---overlap with those in SUN397.
This allows us to study typographic attacks in the broad transfer setting (see Section~\ref{sec:transfer} for more details).

\paragraph{Experimental details.}
We first fine-tune on the attacked version of SUN397, using our standard fine-tuning setup discussed in Section~\ref{sec:method}.
We then evaluate interpolated models on attacked SUN397 and real-world test sets.
For SUN397, this amounts to standard multi-class classification as the class space is fixed.
For real typographic attacks, the task is binary classification between the image category and the text category.
The classification head for each image is generated using the frozen CLIP text encoder.

\FloatBarrier

\section{Counting}
\label{sec:appendix_counting}

\begin{figure}
    \centering
    \includegraphics[width=\linewidth]{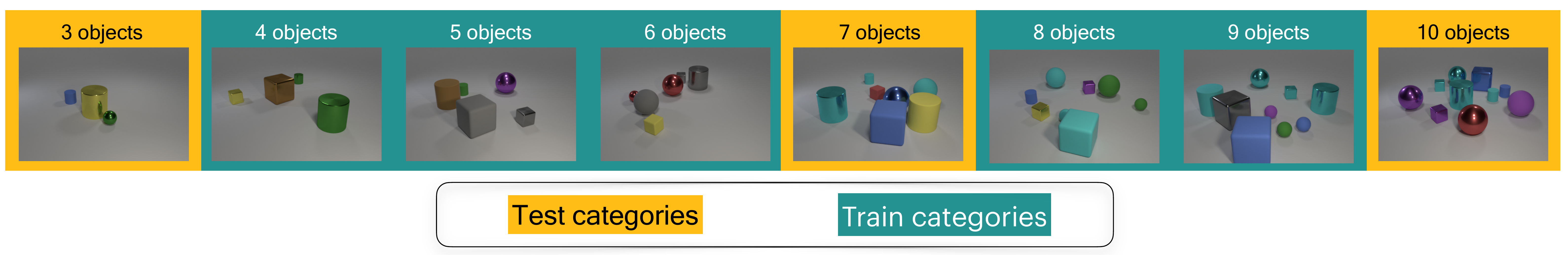}
    \caption{\textbf{CLEVR Counting images and splits.}
    Each image has a number of objects between 3 and 10. For a patching (train) task, models are fine-tuned on images with 4, 5, 6, 8 or 9 objects. We test on images with 3, 7 or 10 objects to evaluate the model's interpolation and extrapolation of the class space.
    }
    \label{fig:counting_task}
\end{figure}

\begin{figure}
    \centering
    \includegraphics[width=0.65\linewidth]{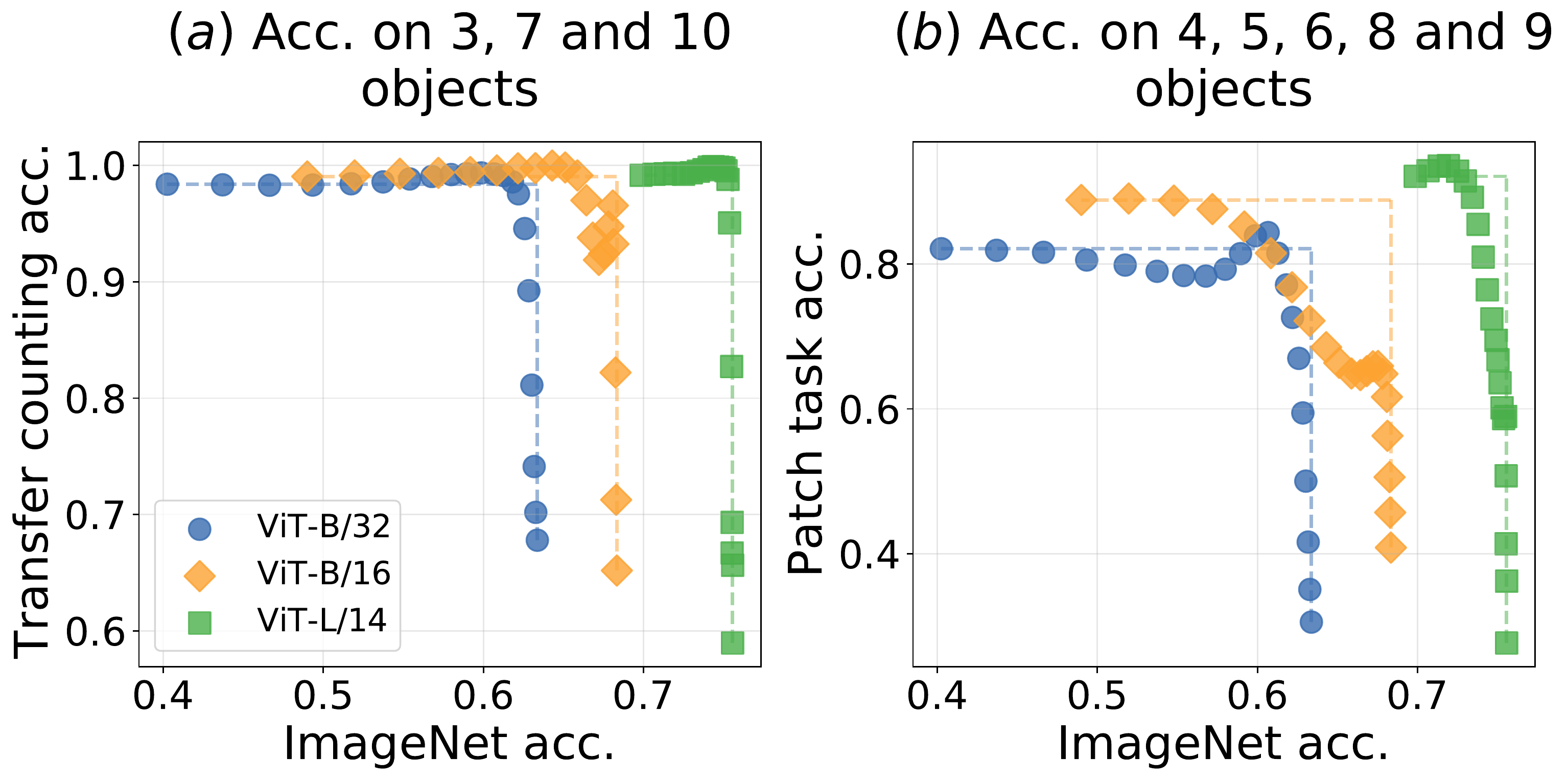}
    \caption{\textbf{CLEVR Counting broad transfer interpolation plots.}
    Curves for different mixing coefficients for (a) the broad transfer task and (b) the patching task.
    The curves suggest that training on the patching task transfer to the training task, even though the class spaces are different.
    }
    \label{fig:counting_transfer}
\end{figure}

We discuss specifics about the dataset and experimental setup for evaluating broad transfer for object counting.

\paragraph{Dataset details.}
We consider the CLEVR dataset~\cite{clevr} with annotations for the number of objects per image.
We split the dataset based on its classes into train and test sets as seen in Figure~\ref{fig:counting_task}.
By training on one split and testing on the other with unseen classes, we are able to evaluate broad transfer for this counting task.

\paragraph{Experimental details.}
The patching task is a 5-way classification task.
The evaluation task is a 3-way classification task, with a different set of classes.
Again the head for the test-time task is created zero-shot using the CLIP text encoder.
Full interpolation curves for transfer and the original task are presented in Figure~\ref{fig:counting_transfer}.

\section{Visual Question Answering}
\label{sec:appendix_vqa}

\begin{figure}
    \centering
    \includegraphics[width=0.7\linewidth]{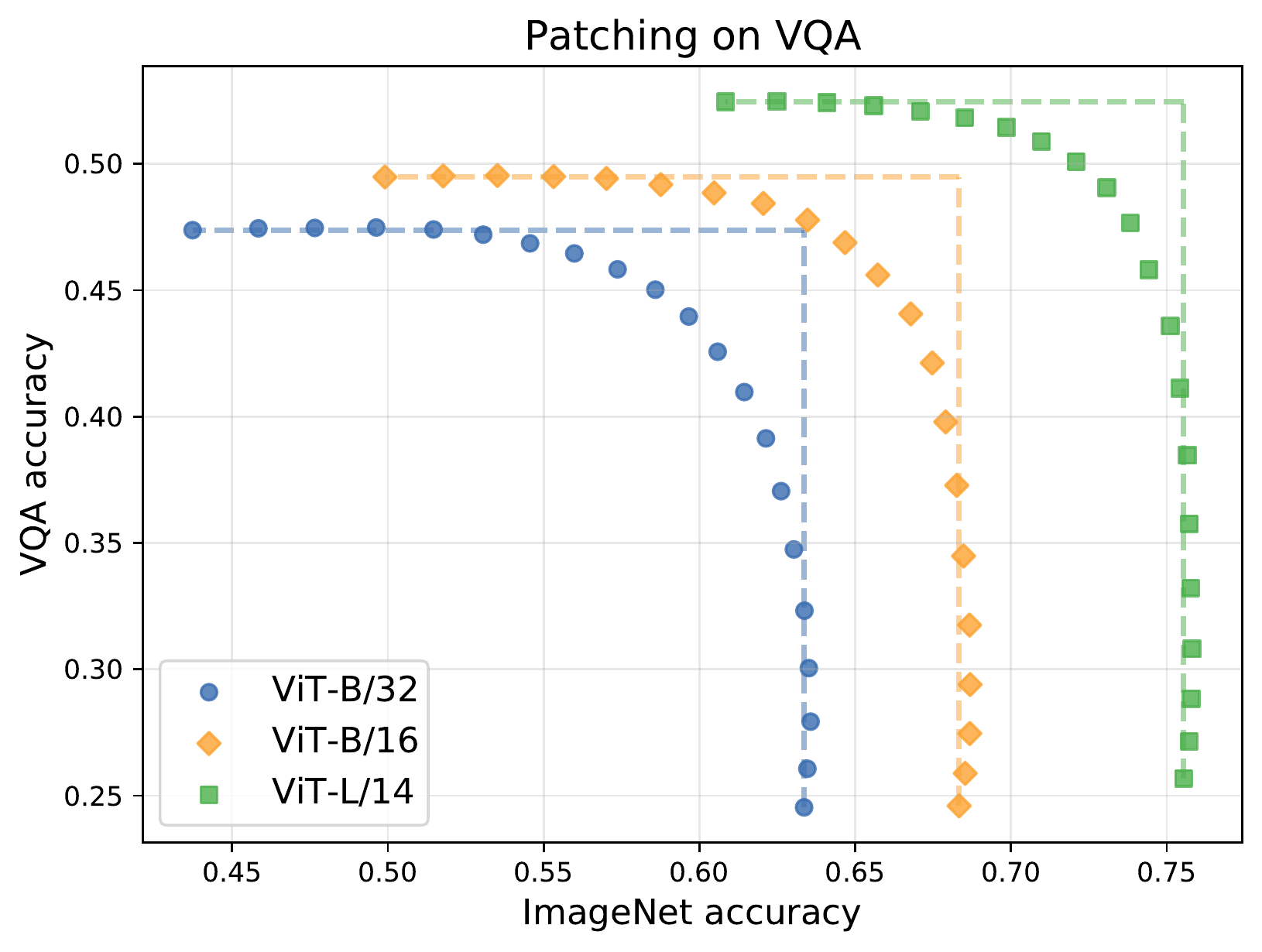}
    \caption{\textbf{Patching for visual question answering.}
    For all models, patching can improve the performance on multiple choice VQA v1 \cite{vqa} with little loss in ImageNet accuracy.}
    \label{fig:vqa_app}
\end{figure}

For multiple-choice visual question answering, we use CLIP to contrast images with each candidate answer. For each candidate answer, we construct a prompt that also includes the question following the template ``Question: [question text] Answer: [answer text]''.
This text is fed to CLIP's text encoder, which remains frozen during the patching process.\footnote{We found that unfreezing the text encoder improved accuracy, but we present results with an frozen text encoder for experimental consistency.}
During fine-tuning, the weights of the vision encoder are updated using a contrastive loss: the feature similarity with the text features of the correct prompt are maximized with respect to the other candidates.
We patch and evaluate on multiple-choice VQA v1 \cite{vqa}, where each question is associated with 18 candidate answers.
Interestingly, 87\% of the prompts used for evaluation are not present in the training data.
The results are shown in Figure \ref{fig:vqa_app}.
Patching improves VQA performance by 13 to 18 percentage points, while reducing ImageNet accuracy by less than one percentage point.

\section{Computational resources}

For all of our experiments, we used NVIDIA A-40 GPUs with 46GB of RAM from an internal cluster.
We estimate the total amount of compute used for the experiments in this paper is around 10 thousand GPU hours.

\FloatBarrier
\section{Tabular results}
\label{sec:appendix_tables}

We now present numerical results for the main experiments presented in this paper.
Tables \ref{tab:single_imagenet} to \ref{tab:single_stl10} present results for patching on a single task, for various supported tasks.
Table \ref{tab:app_multitask} details results for various strategies for patching on multiple tasks.

\begin{table*}
\setlength\tabcolsep{1.3pt}
\renewcommand{\arraystretch}{0.9}
\footnotesize
\begin{center}
\begin{tabular}{cc|cc|cc|cc|cc|cc|cc|cc|cc|cc|c}
\toprule
& & \multicolumn{2}{c|}{Cars} & \multicolumn{2}{c|}{DTD} & \multicolumn{2}{c|}{EuroSAT} & \multicolumn{2}{c|}{GTSRB} & \multicolumn{2}{c|}{KITTI} & \multicolumn{2}{c|}{MNIST} & \multicolumn{2}{c|}{RESISC45} & \multicolumn{2}{c|}{SUN397} & \multicolumn{2}{c|}{SVHN} & \\
& $\alpha$ & S & P & S & P & S & P & S & P & S & P & S & P & S & P & S & P & S & P & Avg\\\midrule
\multirow{21}{*}{B/32}& 0.00 & 63.4 & 59.6 & 63.4 & 44.1 & 63.4 & 45.9 & 63.4 & 32.4 & 63.4 & 22.6 & 63.4 & 48.3 & 63.4 & 60.7 & 63.4 & 63.1 & 63.4 & 31.5& 54.4\\
& 0.05 & 63.4 & 61.9 & 63.4 & 47.0 & 63.4 & 63.4 & 63.4 & 39.7 & 63.4 & 25.5 & 63.4 & 60.9 & 63.5 & 65.9 & 63.5 & 64.4 & 63.4 & 39.9& 57.7\\                                       
& 0.10 & 63.3 & 63.8 & 63.4 & 50.1 & 63.3 & 74.3 & 63.3 & 48.1 & 63.4 & 35.7 & 63.3 & 77.0 & 63.6 & 71.0 & 63.5 & 65.9 & 63.4 & 50.1& 61.5\\                                       
& 0.15 & 63.3 & 65.5 & 63.4 & 53.7 & 63.2 & 82.0 & 63.2 & 57.6 & 63.5 & 47.5 & 63.2 & 87.3 & 63.5 & 75.9 & 63.5 & 67.2 & 63.3 & 60.9& 64.9\\
& 0.20 & 63.2 & 67.4 & 63.3 & 56.5 & 63.0 & 89.3 & 63.0 & 68.3 & 63.4 & 55.6 & 62.8 & 92.6 & 63.5 & 79.9 & 63.4 & 68.5 & 63.1 & 70.4& 67.6\\                                     
& 0.25 & 63.1 & 69.1 & 63.1 & 59.7 & 62.7 & 93.3 & 62.8 & 78.5 & 63.3 & 60.2 & 62.4 & 95.7 & 63.3 & 83.7 & 63.3 & 69.5 & 62.8 & 78.3& 69.7\\
\bottomrule
\end{tabular}
\end{center}
\caption{\textbf{Patching on a single new task where ImageNet is the supported task.} \textit{S}: accuracy on the supported task (ImageNet); \textit{P}: accuracy on the patching task; \textit{Avg:} average accuracy on patching and supported tasks.}
\label{tab:single_imagenet}
\end{table*}

\begin{table*}
\setlength\tabcolsep{1.3pt}
\renewcommand{\arraystretch}{0.9}
\footnotesize
\begin{center}
\begin{tabular}{cc|cc|cc|cc|cc|cc|cc|cc|cc|cc|c}
\toprule
& & \multicolumn{2}{c|}{Cars} & \multicolumn{2}{c|}{DTD} & \multicolumn{2}{c|}{EuroSAT} & \multicolumn{2}{c|}{GTSRB} & \multicolumn{2}{c|}{KITTI} & \multicolumn{2}{c|}{MNIST} & \multicolumn{2}{c|}{RESISC45} & \multicolumn{2}{c|}{SUN397} & \multicolumn{2}{c|}{SVHN} & \\
& $\alpha$ & S & P & S & P & S & P & S & P & S & P & S & P & S & P & S & P & S & P & Avg\\\midrule
\multirow{21}{*}{B/32}& 0.00 & 89.7 & 59.6 & 89.8 & 44.1 & 89.7 & 45.9 & 89.7 & 32.4 & 89.7 & 22.6 & 89.7 & 48.3 & 89.8 & 60.7 & 89.7 & 63.1 & 89.7 & 31.5& 67.6\\
& 0.05 & 89.6 & 61.9 & 89.7 & 47.0 & 90.1 & 63.4 & 90.1 & 39.7 & 89.8 & 25.5 & 90.2 & 60.9 & 89.8 & 65.9 & 89.5 & 64.4 & 90.0 & 39.9& 71.0\\
& 0.10 & 89.5 & 63.8 & 89.6 & 50.1 & 90.6 & 74.3 & 90.2 & 48.1 & 89.8 & 35.7 & 90.4 & 77.0 & 89.7 & 71.0 & 89.4 & 65.9 & 90.3 & 50.1& 74.7\\
& 0.15 & 89.3 & 65.5 & 89.5 & 53.7 & 90.7 & 82.0 & 90.3 & 57.6 & 89.8 & 47.5 & 90.2 & 87.3 & 89.6 & 75.9 & 89.4 & 67.2 & 90.3 & 60.9& 78.1\\
& 0.20 & 89.3 & 67.4 & 89.4 & 56.5 & 90.8 & 89.3 & 90.1 & 68.3 & 89.7 & 55.6 & 90.2 & 92.6 & 89.6 & 79.9 & 89.2 & 68.5 & 90.0 & 70.4& 80.9\\
& 0.25 & 89.2 & 69.1 & 89.2 & 59.7 & 90.6 & 93.3 & 89.8 & 78.5 & 89.4 & 60.2 & 89.7 & 95.7 & 89.4 & 83.7 & 89.2 & 69.5 & 89.6 & 78.3& 83.0\\
\bottomrule
\end{tabular}
\end{center}
\caption{\textbf{Patching on a single new task where CIFAR10 is the supported task.} \textit{S}: accuracy on the supported task (CIFAR10); \textit{P}: accuracy on the patching task; \textit{Avg:} average accuracy on patching and supported tasks.}
\label{tab:single_cifar10}
\end{table*}

\begin{table*}
\setlength\tabcolsep{1.3pt}
\renewcommand{\arraystretch}{0.9}
\footnotesize
\begin{center}
\begin{tabular}{cc|cc|cc|cc|cc|cc|cc|cc|cc|cc|c}
\toprule
& & \multicolumn{2}{c|}{Cars} & \multicolumn{2}{c|}{DTD} & \multicolumn{2}{c|}{EuroSAT} & \multicolumn{2}{c|}{GTSRB} & \multicolumn{2}{c|}{KITTI} & \multicolumn{2}{c|}{MNIST} & \multicolumn{2}{c|}{RESISC45} & \multicolumn{2}{c|}{SUN397} & \multicolumn{2}{c|}{SVHN} & \\
& $\alpha$ & S & P & S & P & S & P & S & P & S & P & S & P & S & P & S & P & S & P & Avg\\\midrule
\multirow{21}{*}{B/32} & 0.00 & 65.1 & 59.6 & 65.1 & 44.1 & 65.1 & 45.9 & 65.1 & 32.4 & 65.1 & 22.6 & 65.1 & 48.3 & 65.1 & 60.7 & 65.1 & 63.1 & 65.1 & 31.5& 55.2\\
& 0.05 & 64.6 & 61.9 & 64.9 & 47.0 & 66.5 & 63.4 & 65.8 & 39.7 & 65.1 & 25.5 & 66.1 & 60.9 & 65.2 & 65.9 & 65.1 & 64.4 & 65.7 & 39.9& 58.8\\
& 0.10 & 64.3 & 63.8 & 64.9 & 50.1 & 67.7 & 74.3 & 66.5 & 48.1 & 65.5 & 35.7 & 67.1 & 77.0 & 65.7 & 71.0 & 64.9 & 65.9 & 66.0 & 50.1& 62.7\\
& 0.15 & 64.2 & 65.5 & 65.1 & 53.7 & 68.2 & 82.0 & 66.7 & 57.6 & 66.0 & 47.5 & 67.2 & 87.3 & 66.2 & 75.9 & 64.7 & 67.2 & 66.2 & 60.9& 66.2\\
& 0.20 & 64.2 & 67.4 & 65.2 & 56.5 & 68.3 & 89.3 & 66.1 & 68.3 & 66.5 & 55.6 & 66.9 & 92.6 & 66.8 & 79.9 & 64.8 & 68.5 & 66.0 & 70.4& 69.1\\
& 0.25 & 64.2 & 69.1 & 65.4 & 59.7 & 68.3 & 93.3 & 65.3 & 78.5 & 66.6 & 60.2 & 66.1 & 95.7 & 66.7 & 83.7 & 64.9 & 69.5 & 65.4 & 78.3& 71.2\\
\bottomrule
\end{tabular}
\end{center}
\caption{\textbf{Patching on a single new task where CIFAR100 is the supported task.} \textit{S}: accuracy on the supported task (CIFAR100); \textit{P}: accuracy on the patching task; \textit{Avg:} average accuracy on patching and supported tasks.}
\label{tab:single_cifar100}
\end{table*}

\begin{table*}
\setlength\tabcolsep{1.3pt}
\renewcommand{\arraystretch}{0.9}
\footnotesize
\begin{center}
\begin{tabular}{cc|cc|cc|cc|cc|cc|cc|cc|cc|cc|c}
\toprule
& & \multicolumn{2}{c|}{Cars} & \multicolumn{2}{c|}{DTD} & \multicolumn{2}{c|}{EuroSAT} & \multicolumn{2}{c|}{GTSRB} & \multicolumn{2}{c|}{KITTI} & \multicolumn{2}{c|}{MNIST} & \multicolumn{2}{c|}{RESISC45} & \multicolumn{2}{c|}{SUN397} & \multicolumn{2}{c|}{SVHN} & \\
& $\alpha$ & S & P & S & P & S & P & S & P & S & P & S & P & S & P & S & P & S & P & Avg\\\midrule
\multirow{21}{*}{B/32}& 0.00 & 84.0 & 59.6 & 84.0 & 44.1 & 84.0 & 45.9 & 84.0 & 32.4 & 84.0 & 22.6 & 84.0 & 48.3 & 84.0 & 60.7 & 84.0 & 63.1 & 84.0 & 31.5& 64.7\\                                    
& 0.05 & 84.0 & 61.9 & 84.0 & 47.0 & 83.9 & 63.4 & 84.0 & 39.7 & 83.9 & 25.5 & 84.0 & 60.9 & 84.0 & 65.9 & 84.0 & 64.4 & 83.9 & 39.9& 68.0\\                                    
& 0.10 & 83.9 & 63.8 & 83.9 & 50.1 & 83.9 & 74.3 & 83.9 & 48.1 & 83.9 & 35.7 & 83.9 & 77.0 & 83.9 & 71.0 & 84.0 & 65.9 & 83.9 & 50.1& 71.7\\                                    
& 0.15 & 83.9 & 65.5 & 83.7 & 53.7 & 83.6 & 82.0 & 83.8 & 57.6 & 83.8 & 47.5 & 83.7 & 87.3 & 83.9 & 75.9 & 84.0 & 67.2 & 83.8 & 60.9& 75.1\\                                    
& 0.20 & 83.9 & 67.4 & 83.5 & 56.5 & 83.3 & 89.3 & 83.6 & 68.3 & 83.7 & 55.6 & 83.3 & 92.6 & 83.7 & 79.9 & 84.0 & 68.5 & 83.6 & 70.4& 77.8\\                                    
& 0.25 & 83.9 & 69.1 & 83.3 & 59.7 & 82.7 & 93.3 & 83.3 & 78.5 & 83.6 & 60.2 & 82.9 & 95.7 & 83.5 & 83.7 & 84.0 & 69.5 & 83.1 & 78.3& 79.9\\
\bottomrule
\end{tabular}
\end{center}
\caption{\textbf{Patching on a single new task where Food101 is the supported task.} \textit{S}: accuracy on the supported task (Food101); \textit{P}: accuracy on the patching task; \textit{Avg:} average accuracy on patching and supported tasks.}
\label{tab:single_food101}
\end{table*}

\begin{table*}
\setlength\tabcolsep{1.3pt}
\renewcommand{\arraystretch}{0.9}
\footnotesize
\begin{center}
\begin{tabular}{cc|cc|cc|cc|cc|cc|cc|cc|cc|cc|c}
\toprule
& & \multicolumn{2}{c|}{Cars} & \multicolumn{2}{c|}{DTD} & \multicolumn{2}{c|}{EuroSAT} & \multicolumn{2}{c|}{GTSRB} & \multicolumn{2}{c|}{KITTI} & \multicolumn{2}{c|}{MNIST} & \multicolumn{2}{c|}{RESISC45} & \multicolumn{2}{c|}{SUN397} & \multicolumn{2}{c|}{SVHN} & \\
& $\alpha$ & S & P & S & P & S & P & S & P & S & P & S & P & S & P & S & P & S & P & Avg\\\midrule
\multirow{21}{*}{B/32}& 0.00 & 97.1 & 59.6 & 97.1 & 44.1 & 97.1 & 45.9 & 97.1 & 32.4 & 97.1 & 22.6 & 97.1 & 48.3 & 97.1 & 60.7 & 97.1 & 63.1 & 97.1 & 31.5& 71.3\\
& 0.05 & 97.2 & 61.9 & 97.2 & 47.0 & 97.0 & 63.4 & 97.2 & 39.7 & 97.1 & 25.5 & 97.2 & 60.9 & 97.2 & 65.9 & 97.2 & 64.4 & 97.2 & 39.9& 74.6\\
& 0.10 & 97.1 & 63.8 & 97.2 & 50.1 & 97.0 & 74.3 & 97.2 & 48.1 & 97.1 & 35.7 & 97.2 & 77.0 & 97.2 & 71.0 & 97.2 & 65.9 & 97.1 & 50.1& 78.3\\
& 0.15 & 97.1 & 65.5 & 97.2 & 53.7 & 96.9 & 82.0 & 97.2 & 57.6 & 97.1 & 47.5 & 97.1 & 87.3 & 97.1 & 75.9 & 97.2 & 67.2 & 97.1 & 60.9& 81.8\\
& 0.20 & 97.1 & 67.4 & 97.1 & 56.5 & 96.5 & 89.3 & 97.1 & 68.3 & 97.0 & 55.6 & 96.9 & 92.6 & 97.1 & 79.9 & 97.1 & 68.5 & 97.1 & 70.4& 84.5\\
& 0.25 & 97.1 & 69.1 & 97.1 & 59.7 & 96.4 & 93.3 & 97.0 & 78.5 & 97.0 & 60.2 & 96.8 & 95.7 & 97.1 & 83.7 & 97.1 & 69.5 & 96.9 & 78.3& 86.7\\
\bottomrule
\end{tabular}
\end{center}
\caption{\textbf{Patching on a single new task where STL10 is the supported task.} \textit{S}: accuracy on the supported task (STL10); \textit{P}: accuracy on the patching task; \textit{Avg:} average accuracy on patching and supported tasks.}
\label{tab:single_stl10}
\end{table*}

\begin{table*}
\setlength\tabcolsep{4.6pt}
\footnotesize
\begin{center}
\begin{tabular}{llccccccccccc}
Model &  Method & \rot{ImageNet} & \rot{Cars} & \rot{DTD} & \rot{EuroSAT} & \rot{GTSRB} & \rot{KITTI} & \rot{MNIST} & \rot{RESISC45} & \rot{SUN397} & \rot{SVHN} & \rot{Average} \\\midrule
\multirow{5}{*}{B/32} & Unpatched & 63.4 & 59.6 & 44.1 & 45.9 & 32.4 & 22.6 & 48.7 & 60.7 & 63.1 & 31.5 & 54.4\\
& Parallel patching & 60.1 & 63.4 & 50.2 & 75.6 & 58.3 & 42.1 & 90.8 & 71.7 & 64.8 & 66.7 & 62.5\\
& Seq. patching (seed 0) & 55.0 & 67.4 & 65.0 & 86.3 & 83.5 & 63.3 & 96.6 & 77.1 & 68.0 & 75.7 & 65.4\\
& Seq. patching (seed 1) & 55.0 & 60.1 & 60.9 & 86.5 & 82.5 & 58.1 & 98.0 & 85.9 & 67.2 & 87.0 & 65.6\\
& Seq. patching (seed 2) & 56.3 & 61.1 & 58.8 & 85.4 & 87.1 & 57.0 & 96.9 & 84.7 & 69.5 & 88.5 & 66.4\\
& Joint patching & 61.5 & 75.2 & 76.0 & 98.0 & 96.9 & 76.2 & 99.2 & 92.6 & 72.6 & 94.2 & 74.1\\
& Multiple models & 63.4 & 79.2 & 78.7 & 98.6 & 98.5 & 77.9 & 99.6 & 95.0 & 75.1 & 97.2 & 76.1\\
\bottomrule
\end{tabular}
\end{center}
\caption{\textbf{Contrasting strategies for patching on multiple tasks.} ImageNet is used as the supported task while the other nine datasets are used for patching.}
\label{tab:app_multitask}
\end{table*}

\clearpage

\end{document}